\documentclass[mnsc,nonblindrev]{informs3_hide}
\OneAndAHalfSpacedXI 

\usepackage[english]{babel}
\usepackage[autostyle, english = american]{csquotes}
\MakeOuterQuote{"}
\usepackage[colorlinks,linkcolor=blue,citecolor=blue,hypertexnames=false]{hyperref}
\usepackage[normalem]{ulem}

\usepackage{cleveref}
\crefname{subsection}{section}{sections}
\usepackage{eqnarray}
\usepackage{algorithm,algpseudocode}
\usepackage{amsmath, bbm, enumitem, xparse, multicol, bm, lipsum}
\usepackage{amssymb}

\usepackage{graphicx}
\usepackage{subcaption}

\newcommand{\eps}{\varepsilon}
\newcommand{\bI}{\mathbbm{1}}
\newcommand{\bE}{\mathbb{E}}
\newcommand{\bR}{\mathbb{R}}

\newcommand{\ALG}{\mathsf{ALG}}

\newcommand{\bV}{\bar{V}}
\newcommand{\bVL}{\bar{V}^{\mathrm{Off}}}
\newcommand{\bVH}{\bar{V}^{\mathrm{Semi}}}
\newcommand{\bVF}{\bar{V}^{\mathrm{Fld}}}
\newcommand{\LL}{L^{\mathrm{Off}}}
\newcommand{\LF}{L^{\mathrm{Fld}}}

\newcommand{\hV}{\hat{V}}
\newcommand{\hM}{\hat{M}}
\newcommand{\Var}{\mathrm{Var}}

\newcommand{\tr}{\tilde{r}}
\newcommand{\ta}{\tilde{a}}
\newcommand{\tj}{\tilde{\jmath}}
\newcommand{\tv}{\tilde{v}}
\newcommand{\tx}{\tilde{x}}
\newcommand{\td}{\tilde{d}}
\newcommand{\hK}{K}
\newcommand{\bc}{\bm{c}}
\newcommand{\by}{\bm{y}}
\newcommand{\tc}{\tilde{c}}
\newcommand{\tbc}{\tilde{\bm{c}}}
\newcommand{\bd}{\bm{d}}
\newcommand{\tba}{\tilde{\bm{a}}}
\newcommand{\ba}{\bm{a}}
\newcommand{\tmu}{\tilde{\mu}}
\newcommand{\tbmu}{\tilde{\bm{\mu}}}
\newcommand{\hbmu}{\hat{\bm{\mu}}}
\newcommand{\bmu}{\bm{\mu}}

\newcommand{\Ftat}{F(\cdot|\tilde{\bm{a}}_t)}
\newcommand{\tdelta}{\tilde{\delta}}
\newcommand{\bx}{\bm{x}}
\newcommand{\tbxi}{\tilde{\bm{\xi}}}
\newcommand{\bxi}{\bm{\xi}}

\usepackage[dvipsnames]{xcolor}

\usepackage{xparse}

\NewDocumentEnvironment{myproof}{o}
{\IfNoValueTF{#1}{\paragraph{{Proof.} }} {\paragraph{{#1.} }} }
{\hfill$\Halmos$}

\usepackage{natbib,bbm,multirow,multicol}
 \bibpunct[, ]{(}{)}{,}{a}{}{,}%
 \def\bibfont{\small}%
\TheoremsNumberedThrough     
\ECRepeatTheorems

\EquationsNumberedThrough    

\begin{document}




\TITLE{
Degeneracy is OK: Logarithmic Regret for Network Revenue Management with Indiscrete Distributions
}

\ARTICLEAUTHORS{%
\AUTHOR{$\text{Jiashuo Jiang}^\dagger$ \quad $\text{Will Ma}^\ddagger$ \quad $\text{Jiawei Zhang}^\S$}

\AFF{\  \\
$\dagger~$Department of Industrial Engineering \& Decision Analytics, Hong Kong University of Science and Technology\\
$\ddagger~$Decision, Risk, and Operations Division, Graduate School of Business, Columbia University\\
$\S$~Department of Technology, Operations \& Statistics, Stern School of Business, New York University\\
}
}

\ABSTRACT{

We study the classical Network Revenue Management (NRM) problem with accept/reject decisions and $T$ IID arrivals.
We consider a distributional form where each arrival must fall under a finite number of possible categories, each with a deterministic resource consumption vector, but a random value distributed continuously over an interval.
We develop an online algorithm that achieves  $O(\log^2 T)$ regret under this model, with the only (necessary) assumption being that the probability densities are bounded away from 0.
We derive a second result that achieves $O(\log T)$ regret under an additional assumption of second-order growth.
To our knowledge, these are the first results achieving logarithmic-level regret in an NRM model with continuous values that do not require any kind of ``non-degeneracy'' assumptions.
Our results are achieved via new techniques including a new method of bounding myopic regret, a ``semi-fluid'' relaxation of the offline allocation, and an improved bound on the ``dual convergence''.

}



\maketitle

\section{Introduction}

Network Revenue Management (NRM) is a {capacity control problem} in which limited resources are to be allocated over a finite time horizon of length $T$.
During each time step $t=1,\ldots,T$, a query $t$ arrives, demanding a vector $\tba_t$ of resources and providing a reward $\tr_t$.
An irrevocable decision must then be made about whether to serve query $t$, in which case $\tba_t$ would be subtracted from the resources and $\tr_t$ would be collected.
Query $t$ is only feasible to serve if the remaining resources exceed $\tba_t$ component-wise, and a feasible query $t$ can still be judiciously rejected, e.g.\ if $\tr_t$ is low compared to the resources consumed in $\tba_t$.
The goal is to maximize the total reward collected from serving queries using the initial resource capacities, when the values $(\tr_t,\tba_t)$ for each query $t$ are unknown before it arrives but known to be drawn IID across time.

Due to the curse of dimensionality in this problem, a mathematically rich literature has evolved out of developing heuristics and obtaining guarantees on their performance.  We consider the stream of literature that analyzes \textit{regret}, which is the additive loss of an online allocation algorithm compared to an optimal offline allocation that knows all values of $(\tr_t,\tba_t)$ beforehand, taken in expectation over the IID query draws (and any further randomness in the algorithm).
Generally, regret is larger for longer time horizons $T$, and this literature is concerned with how the regret grows as a function of $T$ when all other system parameters stay fixed (but the initial resource capacities are also allowed to scale arbitrarily with $T$).

Algorithms with $O(\sqrt{T})$ regret have been known for several decades, as we discuss in \Cref{sec:furtherRelated}.
Since then, many papers have developed algorithms with $\tilde{O}(1)$ regret, which guarantees the regret to grow logarithmically or slower in $T$, under either of the following two
kinds of assumptions.
The first involves having a \textit{small number of possible realizations} for the vector $(\tr_t,\tba_t)$, described by a discrete distribution on $N$ points.
As $T$ grows, $N$ stays fixed and is treated as a constant in the analysis.
However, such an assumption abandons some natural models, e.g.\ that of $\tr_t$ being drawn uniformly from [0,1].
On the other hand, papers that can capture these continuous distributions require a different set of assumptions, which we will call \textit{non-degeneracy}.
At a high level, these papers assume that the mathematical program being re-solved by the online algorithm over time to make its decisions is always well-behaved, and has a unique optimal solution.
However, such assumptions appear to be motivated primarily by the analysis, and are difficult to intuit or verify, {or even state.  We defer their extended statements to \Cref{sec:conclusion}, where we also explain why perturbation attempts to overcome degeneracy would incur a regret of $\Omega(\sqrt{T})$.}

\

\noindent\fbox{
\parbox{\textwidth}{
\OneAndAHalfSpacedXI
\textbf{Our contribution.} We establish logarithmic regret in a natural model of Network Revenue Management {that makes} neither the small-$N$ nor non-degeneracy assumptions.
}}

\

To our knowledge, such a result has not previously existed in the literature, which we now review.
Our result builds upon the existing literature that establishes
logarithmic-or-better regret in either the small-$N$ or non-degenerate settings, or in the multi-secretary special case.

\textbf{NRM with small-$N$.}
\citet{jasin2012re} initially establish a constant $O(1)$ regret under \textit{both} the small-$N$ and non-degeneracy assumptions.
\citet{bumpensanti2020re} and \citet{vera2021bayesian} were the first to establish $O(1)$ regret for a general NRM problem without any non-degeneracy assumptions.  We note that \citet{arlotto2019uniformly} first established $O(1)$ regret in the \textit{multi-secretary}
special case, where all queries demand one unit of a single resource (i.e. $\tba_t=(1)$ w.p.~1).
These are all surprising results, in that given a fixed discrete distribution for $(\tr_t,\tba_t)$, the regret is upper-bounded by an absolute numerical constant, regardless of how long a time horizon $T$ (and correspondingly large resource capacities) over which regret can be incurred.

\textbf{Multi-secretary with general distributions.}
A caveat to the aforementioned analysis is that the $O(1)$ constant depends on $N$, and $N$ can be $\infty$,
e.g.\ for continuous distributions.
To understand continuous distributions better, \citet{bray2019does} studies the example of $\tr_t$ being drawn uniformly from [0,1] in the multi-secretary special case. Both \citet{lueker1998average} and \citet{bray2019does} establish an upper bound on regret that grows logarithmically with $T$, and importantly, show this regret rate of $\Theta(\log T)$ to be \textit{tight}---that is, a constant regret is no longer possible.
Recently, \citet{besbes2022multisecretary} make further progress on the multi-secretary problem by establishing a notion of complexity for general distributions, which affects regret.

\textbf{NRM with general distributions and non-degeneracy.}
Although the preceding papers consider continuous and general reward distributions, it is unclear how they extend beyond the multi-secretary special case.
Meanwhile, several papers \citep{li2021online, balseiro2021survey, bray2022logarithmic} have studied general distributions for NRM, under a non-degeneracy condition that guarantees the binding constraints of the fluid relaxation to remain unchanged when the right-hand-side constraint is being replaced by the real-time per period remaining capacities {(see \Cref{sec:conclusion})}.  However, this requires that the real-time per period remaining capacities stay in a neighborhood of the initial per period remaining capacities. In order to guarantee this, the previous literature needs to assume that the initial capacities scale linearly with $T$. This assumption would require: i) the resource with a non-binding constraint in the fluid relaxation to have a buffer capacity that equals $\delta T$ for some fixed constant $\delta>0$, and; ii) the resource with a binding constraint in the fluid relaxation to have a unique optimal dual variable lower bounded by $\delta$, where the optimal dual variable remains fixed regardless of how large $T$ is since initial capacities scale linearly in $T$ (the regret would scale polynomially in $1/\delta$). However, the square-root law of inventory in practice suggests that the buffer capacity should generally scale as $O(\sqrt{T})$, violating i); and if all resource constraints are binding, then the optimal dual variables can be non-unique or arbitrarily close to $0$.

All in all, as highlighted in \citet{bumpensanti2020re}, degeneracy is likely to occur in practice, and it has remained unknown whether a logarithmic level regret can be achieved in NRM with indiscrete distributions and without non-degeneracy.

\subsection{Logarithmic Regret for NRM with Discrete Demands and Continuous Rewards}

We focus on the following structural form for the distribution of $(\tr_t,\tba_t)$.
First, the demand vector $\tba_t$ is drawn from a discrete distribution supported on finitely many possibilities $\ba_1,\ldots,\ba_n$.
Then, conditional on $\tba_t=\ba_j$ for any $j=1,\ldots,n$, the reward $\tr_t$ is drawn from a continuous distribution $F_j$ supposed on an interval $[l_j,u_j]$ whose density is lower bounded by a constant $\alpha>0$.

Our main result (\textbf{\Cref{sec:knowndistribution}}) is to develop an algorithm with $O(\log^2 T)$ regret, assuming only this structural form for the distribution of $(\tr_t,\tba_t)$.
We note that finiteness of $n$ and the density lower bound $\alpha$ are {prevalent conditions} to get a regret that is sub-polynomial in $T$ \citep[see][]{besbes2022multisecretary}.
And although $n$ is finite, the continuous densities still induce an infinite support for the realization of vector $(\tr_t,\tba_t)$---therefore, we have successfully established logarithmic level regret in a model of NRM that allows for infinitely many possible realizations, without assuming non-degeneracy.

Our model has a natural practical interpretation---there is a discrete list of $n$ flight itineraries, and a continuous range of prices $[l_j,u_j]$ that customers are willing to pay for each itinerary $j$.
And although we have been describing accept/reject formulations of NRM,
we can also analyze pricing formulations---where a price must be posted before seeing the customer willingness-to-pay $\tilde{r}_t$, and that price is collected as reward if and only if it is no greater than $\tilde{r}_t$.  Our results will translate to the pricing formulation as long as demands are "independent"---that is, each customer is interested in a specific itinerary $j$ and whether they purchase it does not depend on the prices of itineraries $j'\neq j$.
This translation is achieved through the standard reduction of virtual valuations, as we detail in \Cref{sec:pricebased}.
We should note that pricing was the original formulation of NRM \citep{gallego1997multiproduct}, in which logarithmic \citep{jasin2014reoptimization} and constant \citep{wang2022constant} regret is known even without the independent demand assumption; however, these papers do need the non-degeneracy assumption.

The $O(\log^2 T)$ regret established in our main result can also be seen as an extension of a corollary of \citet{besbes2022multisecretary}.
To elaborate, in our model two different indices $j,j'$ can have $\ba_j=\ba_{j'}$; moreover, they can have \textit{non-overlapping} reward intervals with $l_j<u_j<l_{j'}<u_{j'}$.  Thus, with a single resource and $\ba_j=(1)$ for all $j=1,\ldots,n$, we can capture the multi-secretary reward distribution with density lower-bounded over disjoint intervals, for which \citet{besbes2022multisecretary} already established a $O(\log^2 T)$ regret.

Our second result (\textbf{\Cref{sec:Logsection}}) is to derive an $O(\log T)$ regret bound with additional assumptions that require the density to be upper bounded and guarantee the strong convexity of the Lagrangian dual function of the fluid approximation. We also show that the constant term in this $O(\log T)$ regret bound depends polynomially on all of the problem parameters, whereas in contrast, the constant term in our $O(\log^2 T)$ regret bound depended exponentially on $n$.

Aside from achieving improved regret bounds, we outline how our algorithms differ from the literature.
For our algorithm in \Cref{sec:knowndistribution}, the relaxation that it is re-solving over time is different---it is re-solving a new relaxation that we call the "semi-fluid" relaxation of the offline optimum. We then round the solution to this relaxation, obtaining a brand new algorithm.
By contrast, our algorithm in \Cref{sec:Logsection} is the same as the classical "certainty-equivalent policy" (e.g. \cite{balseiro2021survey, li2021online, bray2022logarithmic}) that re-solves the fluid upper bound and uses its optimal (dual) solution to guide our decision. Our innovation is in the theoretical analysis, where we show that the certainty-equivalent policy can obtain the logarithmic regret with additional conditions, but without requiring strict complementary slackness of the fluid approximation.

\subsection{Further Related Work} \label{sec:furtherRelated}

The network revenue management (NRM) problem has been extensively studied in the literature and one main topic is to develop near-optimal policies with strong theoretical guarantees. One common way is to derive the policy from the optimal solution of the ex-ante relaxation. To be specific, \cite{talluri1998analysis} propose a static bid-price policy based on the optimal dual variable of the ex-ante relaxation and proves that the regret bound is $O(\sqrt{T})$. Then, a dynamic update of the bid-price is considered in the literature. Subsequently,
\cite{reiman2008asymptotically} shows that by re-solving the ex-ante relaxation once to update the bid-price, one can obtain an improved regret bound $o(\sqrt{T})$. Then, \cite{jasin2012re} shows that under a non-degeneracy condition for the ex-ante relaxation, a policy which re-solves the ex-ante relaxation at each time period will lead to an $O(1)$ regret. The relationship between the performances of the control policies and the number of times of re-solving the ex-ante relaxation is further discussed in their later paper \citep{jasin2013analysis}. More recently, \cite{bumpensanti2020re} proposes an infrequent re-solving policy and shows a regret bound of $O(1)$ without the ``non-degeneracy'' assumption. This has been extended by \citet{balseiro2022uniformly} to fair allocation problems. With a different approach, \cite{vera2021bayesian} proves the same $O(1)$ upper bound for the NRM problem and their approach is further generalized in series of papers (e.g. \cite{freund2019good,vera2021online,freund2021overbooking}). Recent studies on the NRM problem includes variants such as the reusable resource setting \citep{baek2022bifurcating}, unknown distribution setting \citep{li2020simple, balseiro2022best} and imperfect distribution knowledge setting under a non-stationary environment \citep{jiang2020online}.

Another problem that is closely related to the NRM problem is called the online packing problem, where a more general formulation is studied and less distribution knowledge is assumed. The packing problem covers a wide range of applications, including secretary problem \citep{ferguson1989solved, arlotto2019uniformly}, online knapsack problem \citep{arlotto2020logarithmic, Jiang2020OnlineRA}, resource allocation problem \citep{asadpour2020online}, network routing problem \citep{buchbinder2009online}, matching problem \citep{mehta2007adwords}, etc. The problem is usually studied under either a stochastic model where the reward and size of each query are drawn independently from an unknown distribution $\mathcal{P}$, or a more general random permutation model where the queries arrive in a random order  \citep{molinaro2014geometry, agrawal2014dynamic, kesselheim2014primal, gupta2014experts}.

{Our work comes subsequent to \cite{besbes2022multisecretary}, but has the following relation with an extended version of their work \citep{besbes2023dynamic}. In the extended version, they develop a general and practical framework, Repeatedly Act using Multiple Simulations (RAMS), that simulates future demand scenarios to guide online decisions. Their framework presents a new way to derive our algorithms and bounds.}

\section{Problem Formulation and Our Approach} \label{sec:probDef}
We consider an online resource allocation problem, where there are $m$ resources and each resource $i\in[m]$ has an initial fractional capacity $C_i\in\mathbb{R}_{\ge0}$. There are $T$ discrete time periods and at each period $t\in[T]$, one query arrives, denoted by query $t$. Each query $t$ has a \textit{random} size $\tba_t=(\ta_{t,1},\dots,\ta_{t,m})\in\mathbb{R}^m_{\ge0}$, where $\ta_{t,i}$ denotes how much resource $i$ will be consumed if query $t$ is served, for all $i\in[m]$, and a \textit{random} reward $\tr_t\in\mathbb{R}_{\ge0}$ that denotes how much reward can be collected by serving query $t$.  We assume that the value of $(\tr_t, \tba_t)$ for each $t\in[T]$ is drawn \textit{independently} from an identical distribution denoted by $F(\cdot)$. We suppose that the queries are of finite types\footnote{Here we are using a different meaning of "type" than \citet{besbes2022multisecretary}.  According to their meaning, because we have a continuum of possible realizations for $\tilde{r}_t$, we would have infinitely many types.}, i.e., for each $t\in[T]$, $\tba_t$ is supported on a finite set $\mathcal{A}=\{\ba_1,\dots,\ba_n\}$. We call the situation where $\tba_t$ is realized as $\ba_j$ as query $t$ being of type $j$ and we denote by $p_j=P(\tba=\ba_j)$, for each $j\in[n]$.

After query $t$ arrives and the value of $(\tr_t,\tba_t)$ is revealed, the decision maker has to decide immediately and irrevocably whether or not to serve query $t$. {Note that query $t$ can only be served if for every resource $i$ its remaining capacity it at least $\ta_{t,i}$.} The goal of the decision maker is to maximize the total collected reward subject to the resource capacity constraint. 

Any online policy $\pi$ for the decision maker is specified by a set of decision variables $\{\tx_{t}^{\pi}\}_{\forall t\in[T]}$, where $\tx_t^{\pi}$ is a binary variable and denotes whether query $t$ is served, for all $t\in[T]$. Note that $\tx_t^{\pi}$ can be stochastic if $\pi$ is a randomized policy. Any policy $\pi$ is feasible if for all $t\in[T]$, $\tx_t^{\pi}$ depends only on $F(\cdot)$ and $\{(\tr_1, \tba_1), \dots, (\tr_t, \tba_t)\}$, and the following capacity constraint is satisfied:
\begin{equation}\label{eqn:capacityconstraint}
\sum_{t=1}^T \ta_{t,i}\cdot \tx_t^{\pi}\leq C_i,~~\forall i\in[m].
\end{equation}
The total collected value of policy $\pi$ is given by $V^{\pi}(I)=\sum_{t=1}^T\tr_t\cdot \tx_t^{\pi}$, where $I=\{(\tr_t, \tba_t)\}_{t=1}^T$ denotes the problem instance.

The benchmark is the prophet, which is an offline decision maker that is aware of the value of $(\tr_t,\tba_t)$ for all $t\in[T]$ and always makes the optimal decision in hindsight. We denote by $\{\tx^{\text{off}}_t\}_{t=1}^T$ the offline decision of the prophet, which is an optimal solution to the following offline problem:
\begin{eqnarray}
V^{\text{off}}(I)=&\max & \sum_{t=1}^T \tr_t\cdot x_t \label{lp:offlineoptimum}\\
&\mathrm{s.t.\ } & \sum_{t=1}^T \ta_{t,i}\cdot x_t \leq C_i,~~\forall i\in[m] \nonumber\\
& &  x_t\in\{0,1\} \quad \quad \forall t\in[T]. \nonumber
\end{eqnarray}
For any feasible online policy $\pi$, we use \textit{regret} to measure its performance, which is defined as follows:
\begin{equation}\label{eqn:regret}
    \text{Regret}(\pi):= \mathbb{E}_{I\sim F}[V^{\text{off}}(I)]-\mathbb{E}_{ I\sim F}[V^{\pi}(I)]
\end{equation}
where $I=\{(\tr_t,\ta_t)\}_{t=1}^T\sim F$ denotes that $(\tr_t, \ta_a)$ follows distribution $F(\cdot)$ independently for each $t\in[T]$.
In what follows, we describe our general approach to upper bound the regret defined in \eqref{eqn:regret}, and discuss how our approach implies online policies under various settings.

\subsection{$O(\log^2 T)$ Regret with Lower-bounded Densities} \label{sec:introModel1}

Our $O(\log^2 T)$ result tries to extend the approach of \citet{vera2021bayesian} from discrete distributions, which we now recap.
Fix some remaining resource capacities and suppose there are $s$ time steps left.
The offline allocation knows for every possible realization $(r,\ba)$, called a ``type'', the remaining number of queries $t$ with $(\tr_t,\tba_t)=(r,\ba)$, denoted by $\tilde{d}_{(r,\ba)}$.
Meanwhile, consider an online algorithm that solves for an optimal \textit{fluid} packing, which replaces each $\tilde{d}_{(r,\ba)}$ with its expectation to specify the queries that can be served.
The authors compare $\tilde{y}_{(r,\ba)}$, the number of queries of type $(r,\ba)$ accepted by the offline, to $\hat{y}_{(r,\ba)}$, the number of such queries instructed by the fluid packing to accept.
Their key argument is that since the number of queries of type $(r,\ba)$ is growing linearly in $s$ and any two of them are interchangeable, the online algorithm only ``makes a mistake'' if $\tilde{y}_{(r,\ba)}$ and $\hat{y}_{(r,\ba)}$ are distance $\Omega(s)$ apart.  This is a highly unlikely event (over the randomness in the offline's draws of $\tilde{d}_{(r,\ba)}$) because $\tilde{y}_{(r,\ba)}$ and $\hat{y}_{(r,\ba)}$ are generally only $O(\sqrt{s})$ apart, specifically an event with probability $O(e^{-s})$ that when summed over $s$ leads to constant regret.

\textbf{A new, ``semi-fluid'' relaxation of offline.}
Comparing $\tilde{y}_{(r,\ba)}$ to $\hat{y}_{(r,\ba)}$ is meaningless under continuous rewards, because there is zero probability of drawing any specific $(r,\ba)$.
To cope, we introduce a new \textit{semi-fluid} relaxation that amalgamates decisions over queries with the same demand vector.
Specifically, we call each $j=1,\ldots,n$ a \textit{type} in our model of NRM, and let $\tilde{d}_j$ denote the remaining number of queries with $\tba_t=\ba_j$ and $\tr_t$ drawn from $F_j$.
The semi-fluid relaxation knows $\tilde{d}_j$ for all $j$.
However, the semi-fluid relaxation differs from the offline allocation in that it collects exactly the ``fluid'' value
\begin{align} \label{eqn:avgReward}
\tilde{d}_j\int_{1-\tilde{y}_j/\tilde{d}_j}^{1}F^{-1}_j(q)dq
\end{align}
when it accepts $\tilde{y}_j$ queries of type $j$.  Objective~\eqref{eqn:avgReward} integrates over the $\tilde{y}_j/\tilde{d}_j$ proportion of the $\tilde{d}_j$ type-$j$ queries with the highest rewards, as explained in \Cref{sec:knowndistribution}.
Meanwhile, our algorithm is still based on solving the (fully) fluid packing, whose variables can also be amalgamated into acceptance quantities $\hat{y}_j$ for each type $j$ (and $\tilde{d}_j$ will be replaced  $\bE[\tilde{d}_j]$, including in the objective~\eqref{eqn:avgReward}).
We can then compare the acceptance quantities $\tilde{y}_j$ to $\hat{y}_j$ for the amalgamated types $j$.

\textbf{Bounding the myopic regret.}
Unfortunately, because in our model queries of the same type have different rewards, a mistake can be made without requiring $|\tilde{y}_j-\hat{y}_j|=\Omega(s)$.
In fact, a mistake only requires drawing a quantile that is above the acceptance proportion $\tilde{y}_j/\tilde{d}_j$ for the semi-fluid but below $\hat{y}_j/\bE[\tilde{d}_j]$ for the fluid (or vice versa), which occurs with probability $|\tilde{y}_j/\tilde{d}_j-\hat{y}_j/\bE[\tilde{d}_j]|$.
As such, the rough argument that $\tilde{y}_j$ and $\hat{y}_j$ are $O(\sqrt{s})$ apart (and $\bE[\tilde{d}_j]=\Omega(s)$) would lead to a mistake probability of $O(1/\sqrt{s})$, and an undesirable overall regret of $O(\sqrt{T})$.
Therefore, we instead follow \citet{bray2019does} who argues that for continuous distributions one must quantify the ``myopic regret'' at each time step (instead of just bounding the probability that it is non-zero).
We decompose overall regret in way (see \Cref{sec:probDef}) such that the regret at a time step can be quanfied as
\begin{align} \label{eqn:myopicIntro}
\bVH_{\bc}(\tilde{\bd}+\mathbf{e}_j)-\int_{1-\hat{y}_j/\bE[\tilde{d}_j]}^{1}F^{-1}_j(q)dq-\frac{\hat{y}_j}{\bE[\tilde{d}_j]}\cdot\bVH_{\bc-\ba_j}(\tilde{\bd})-\left(1-\frac{\hat{y}_j}{\bE[\tilde{d}_j]}\right)\cdot \bVH_{\bc}(\tilde{\bd}),
\end{align}
where $\bc$ denotes the remaining resource capacities, 
$j$ denotes the type of the current query, and
$\bVH_{\bc}(\bd)$ denotes the optimal objective value of the semi-fluid relaxation given remaining resources $\bc$ and a generic vector $\bd=(d_1,\ldots,d_n)$ counting the remaining queries of each type.

To upper-bound~\eqref{eqn:myopicIntro}, we take an optimal solution $\tilde{\by}=(\tilde{y}_1,\ldots,\tilde{y}_n)$ for $\bVH_{\bc}(\tilde{\bd}+\mathbf{e}_j)$ and modify it into feasible solutions for $\bVH_{\bc-\ba_j}(\tilde{\bd})$ and $\bVH_{\bc}(\tilde{\bd})$, which lower-bounds the latter quantities.
As long as these feasible solutions can be constructed by modifying only the $j$'th coordinate of $\tilde{\by}$, we show that~\eqref{eqn:myopicIntro} is $O((\tilde{y}_j/\tilde{d}_j-\hat{y}_j/\bE[\tilde{d}_j])^2)$.  Per the earlier discussion, this is roughly $O((1/\sqrt{s})^2)=O(1/s)$, which when summed over $s$ would sufficiently lead to logarithmic regret.

\textbf{A boundary-attracted algorithm.}
If the semi-fluid solution $\tilde{y}_j$ is close to 0, however, then $\tilde{\by}$ is difficult to modify into a feasible solution for $\bVH_{\bc-\ba_j}(\tilde{\bd})$---one cannot pack into the reduced capacity $\bc-\ba_j$ by only reducing coordinate $\tilde{y}_j$.
Our strategy is to bypass this boundary situation by tweaking the online algorithm---if there is a risk of this infeasibility, which we show can be identified by the algorithm checking whether $\hat{y}_j=O(\sqrt{s\log s})$, then it \textit{always rejects} the current type-$j$ query.
This effectively sets $\hat{y}_j=0$ in~\eqref{eqn:myopicIntro} and avoids having to lower-bound the quantity $\bVH_{\bc-\ba_j}(\tilde{\bd})$.
On the other extreme, our algorithm \textit{always accepts} if $\hat{y}_j$ is within $O(\sqrt{s\log s})$ of its maximum value $\bE[\tilde{d}_j]$.
All in all, we use this tweaked version of the fluid re-solving algorithm that is ``attracted to boundaries'', which is similar in spirit to the thresholding in \citet{bumpensanti2020re} and the ``conservatism with respect to gaps'' in \citet{besbes2022multisecretary}.  We provide a new explanation for it based on our analysis, and it allows us to always upper-bound~\eqref{eqn:myopicIntro} by $O((\tilde{y}_j/\tilde{d}_j-\hat{y}_j/\bE[\tilde{d}_j])^2)=O(1/s)$ while sacrificing only a log-factor---ultimately achieving $O(\log^2 T)$ regret.

\textbf{Lipschitz property for semi-fluid convex program.}
Finally, it should not be taken for granted that $\tilde{y}_j$ and $\hat{y}_j$ are nearby, an intuition we have been frequently using.
Indeed, they correspond to optimal solutions of mathematical programs with different objective functions---the fluid problem replaces $\tilde{d}_j$ with $\bE[\tilde{d}_j]$ in~\eqref{eqn:avgReward}---and a simple example \citep[Remark~2.7]{mangasarian1987lipschitz} reveals that optimal solution sets are highly sensitive to small perturbations in the objective.
Nonetheless, we extend (\Cref{lem:perturbationLP}) the Lipschitz analysis of \citet{mangasarian1987lipschitz} to show that the specific objective function~\eqref{eqn:avgReward} is well-behaved.
We also note that due to degeneracy, it is necessary for the Lipschitz property to be of the form ``given any optimal solution $\hat{\by}=(\hat{y}_1,\ldots,\hat{y}_n)$ to the fluid, \textit{there exists} a nearby optimal solution $\tilde{\by}$ to the semi-fluid''.
General perturbation analysis results for convex programs \citep{bonnans2013perturbation}, which try to argue that \textit{all} optimal solutions $\tilde{\by}$ are nearby, do not apply in our setting with degeneracy.

\subsection{$O(\log T)$ Regret with Bounded Densities and Second-order Growth Assumption} \label{sec:introModel2}

Based on the structural form introduced previously, we further assume that the reward density is also upper bounded, conditional on $\tba_t=\ba_j$, and additionally, we impose a second-order growth condition over the Lagrangian dual function of the ex-ante relaxation, which requires the dual function to be strongly convex. As discussed earlier, the second-order growth condition, together with the non-degeneracy assumption, lead to the $O(\log T\log\log T)$ regret bound in \cite{li2021online} and the $O(\log T)$ regret bound in \cite{balseiro2021survey} and \cite{bray2022logarithmic}. In comparison to the aforementioned papers, our contribution here is to derive an $O(\log T)$ regret bound by relaxing the non-degeneracy assumption. We now explain at a high level our approach.

\textbf{A tighter relaxation of the offline allocation.}
Our improvement comes from using a tighter relaxation than the ex-ante relaxation as an upper bound of the offline allocation. Our relaxation is that for each sample path, we relax the integral decision of the offline allocation to be fractional, and we take an expectation over the sample path. Such an LP relaxation of the offline allocation has been derived in \cite{bumpensanti2020re} and \cite{vera2021bayesian} in the discrete setting and we derive it here for general distributions. Then, by following a myopic regret approach described in \Cref{sec:GDescription}, we are able to bound the regret incurred at each period by the variance of the dual variable of the LP relaxation of the offline allocation, no matter what the remaining capacities are. Note that we do not need to consider the optimal basis of the ex-ante relaxation to bound the myopic regret. This is the key distinction between our approach and the martingale-based approach in \cite{li2021online, balseiro2021survey, bray2022logarithmic}, which would require a non-degeneracy assumption to guarantee the optimal basis remains fixed in their analysis.

\textbf{A dual convergence bound.}
A second element of our approach is the dual convergence bound, which regards the variance of the dual variable of the LP relaxation of the offline allocation. To be specific, when there are $s$ data points, we prove the dual convergence bound to be at the order of $O(\frac{1}{s})$. To obtain this result, we utilize both ways of splitting the whole space into a set of small cubes with exponentially increasing edge lengths in \cite{huber1967under} and \cite{li2021online}. Denote by $\tbmu$ the dual variable of the sample average problem and $\hbmu$ the dual variable of the ex-ante relaxation. Then, we apply the approach in \cite{huber1967under} to obtain a bound on $P(|\tbmu-\hbmu|\geq\eps)$. However, we note that when $\eps>\sqrt{\frac{1}{s}}$, the approach in \cite{li2021online} would give us a tighter probability bound. Therefore, by applying different ways to bound $P(|\tbmu-\hbmu|\geq\eps)$ for different ranges of $\eps$, we get a $O(\frac{1}{s})$ bound on the dual convergence $\mathbb{E}[\|\tbmu-\hbmu\|_2^2]$.

We do note that the ``dual convergence'' bound is used in different ways between our analysis vs.\ \cite{li2021online}. The data points for ``dual convergence'' in \cite{li2021online} comes from past periods. By contrast, the ``data points'' for ``dual convergence'' in our analysis comes from the future periods. This is because our online decision is made based on the ex-ante relaxation, while our benchmark, the LP relaxation of the offline allocation, makes the decision based on each sample path of future periods. The myopic regret caused by this distinction is shown to be bounded by the variance of the dual variable, where the randomness comes from the sample path of future periods. Though the future sample path is convoluted, we only use its distribution to give a bound and our algorithm does not require any knowledge of the realization. {\cite{bray2022logarithmic} presents another way to derive the dual convergence bound and obtains the $O(\frac{1}{s})$ bound independently.}

\textbf{Discussion on the second-order growth condition.} We do acknowledge that the second-order growth condition in \Cref{sec:Logsection} is somewhat stronger than the one in existing literature \citep{li2021online, balseiro2021survey, bray2022logarithmic}. To be specific, the second-order growth condition assumed in \Cref{sec:Logsection} holds for the Lagrangian dual function given \textit{any} remaining average capacities, while the condition in the existing literature holds given remaining average capacities belonging to a neighborhood of the initial average capacities. However, the second-order growth condition in \Cref{sec:Logsection} is a consequence of our problem formulation instead of a primitive assumption. Moreover, since it is assumed that the ex-ante relaxation admits a unique optimal dual variable in \citet{li2021online, balseiro2021survey, bray2022logarithmic}, their second-order growth conditions are stated as the strong-convexity of the Lagrangian dual function. In contrast, the second-order growth condition in our setting is stated as the strong-convexity after projecting every variable into the subspace that is spanned by the set of possible query sizes and we do not require the uniqueness of the optimal dual variable.

\section{Policy with Log-squared Regret}\label{sec:knowndistribution}
In this section, we derive a log-squared bound for \eqref{eqn:regret} under the following assumption over the distribution $F(\cdot)$.
\begin{assumption}\label{assump:finitesize}
We assume that for each $j\in[n]$, conditional on $\tba_t$ being realized as any $\ba_j\in\mathcal{A}$, the reward distribution of $\tr$ is supported on the interval $[l_j, u_j]$ with a density function $f(\cdot|\ba_j)$, where $u_j\geq l_j\geq 0$, and it satisfies $f(r|\ba_j)\geq\alpha$, for a constant $\alpha>0$, for any $r\in[l_j, u_j]$.
\end{assumption}
For notation simplicity, we denote by $F_j(\cdot)=F(\cdot|\ba_j)$ and $f_j(\cdot)=f(\cdot|\ba_j)$.
Note that in the above \Cref{assump:finitesize}, we allow $l_j=u_j$ for a type $j$, i.e., the reward distribution for type $j$ query is a point mass. In this case, we let the density be $f_j(r)=\infty$ for $r=l_j=u_j$ and any constant $\alpha$ would satisfy $f_j(r)\geq\alpha$ for $r\in[l_j, u_j]$. 

\subsection{General Description of Our Approach}\label{sec:GDescription}
We now give a general description of our approach. We denote by $\bc=(c_{1},\dots,c_{m})\in\mathbb{R}^m$ any vector of remaining capacities of the resources at the beginning of a period $t$. Then, on problem instance $I_t=\{(\tr_t,\tba_t),\dots, (\tr_T,\tba_T)\}$, we denote by $\bV_{\bc}(I_t)$ a relaxation of the total reward collected by the prophet from period $t$ up to period $T$, given the remaining capacity $\bc$, where the decision variable $x_{\tau}\in\{0,1\}$ is relaxed into $x_{\tau}\in[0,1]$ for $\tau=t,\dots,T$. We specify various formulations of the relaxation $\bV_{\bc}(I_t)$ to deal with various settings in the following sections. 
Then, the regret of any online policy $\pi$ can be upper bounded by the gap between $\mathbb{E}_{I_1\sim F}[\bV_{\mathbf{C}}(I_1)]$, where $\bm{C}=(C_1,\dots, C_m)$ is a vector of initial capacity for all resources, and $\mathbb{E}_{\pi, I_1\sim F}[V^{\pi}(I)]$, i.e., 
\begin{equation}\label{eqn:firstupper}
\text{Regret}(\pi)\leq \mathbb{E}_{I_1\sim\bm{F}}[\bV_{\mathbf{C}}(I_1)]-\mathbb{E}_{\pi, I_1\sim\bm{F}}[V^{\pi}(I_1)].
\end{equation}
Our approach relies on the following decomposition of the upper bound in \eqref{eqn:firstupper}. For each $t\in[T]$, we denote by $\tilde{\bm{c}}^{\pi}_t=(\tc^{\pi}_{t,1},\dots,\tc^{\pi}_{t,m})\in\mathbb{R}^m$ the remaining capacities at the beginning of period $t$ during the execution of the policy $\pi$. Note that $\tilde{\bm{c}}^{\pi}_t$ is random for each $t\in[T]$, where the randomness comes from the randomness in the problem instance $I$ and any randomness in the policy $\pi$. Then, the term $\bV_{\mathbf{C}}(I_1)$ can be telescoped as follows by noting that $\tilde{\bm{c}}^{\pi}_1=\bm{C}$ and $\bV_{\bc}(I_{T+1})=0$ for every $\bc$:
\begin{equation}\label{eqn:telescopeoff}
\bV_{\mathbf{C}}(I_1)=\bV_{\tilde{\bm{c}}^{\pi}_1}(I_1)-\bV_{\tilde{\bm{c}}^{\pi}_{T+1}}(I_{t+1})=\sum_{t=1}^T \left( \bV_{\tilde{\bm{c}}^{\pi}_t}(I_t)-\bV_{\tilde{\bm{c}}^{\pi}_{t+1}}(I_{t+1}) \right).
\end{equation}
Thus, the regret upper bound \eqref{eqn:firstupper} can be decomposed as:
\[\begin{aligned}
\mathbb{E}_{I_1\sim\bm{F}}[\bV_{\mathbf{C}}(I_1)]-\mathbb{E}_{\pi, I_1\sim\bm{F}}[V^{\pi}(I_1)]&=\mathbb{E}_{\pi,I_t\sim\bm{F}}\left[\sum_{t=1}^T \left( \bV_{\tilde{\bc}^{\pi}_t}(I_t)-\bV_{\tilde{\bc}^{\pi}_{t+1}}(I_{t+1})-\tr_t\cdot \tx^{\pi}_t \right)\right]\\
&=\sum_{t=1}^T \mathbb{E}_{\pi,I_t\sim\bm{F}}\left[ \bV_{\tilde{\bc}^{\pi}_t}(I_t)-\bV_{\tilde{\bc}^{\pi}_{t+1}}(I_{t+1})-\tr_t\cdot \tx^{\pi}_t \right]\\
&=\sum_{t=1}^T \mathbb{E}_{\pi,I_t\sim\bm{F}}\left[ \bV_{\tilde{\bc}^{\pi}_t}(I_t)-\bV_{\tilde{\bc}^{\pi}_{t}-\tba_t\cdot \tx^{\pi}_t}(I_{t+1})-\tr_t\cdot \tx^{\pi}_t \right]
\end{aligned}\]
where the third equality follows from the identity that $\tilde{\bc}^{\pi}_{t+1}=\tilde{\bc}^{\pi}_t-\tba_t\cdot \tx^{\pi}_t$.
We proceed to analyze the term for each $t\in[T]$ in the above summation. 
For each $\bc\geq0$, we now denote by
\begin{align}
\text{Myopic}_t(\pi,\bc)&=\mathbb{E}_{\pi,I_t}\left[\bV_{\bc}(I_t)-\bV_{\bc-\tba_t\cdot \tx^{\pi}_t}(I_{t+1})-\tr_t\cdot \tx^{\pi}_t\right].\label{def:myopicregret}
\end{align}
It is clear that in order to upper bound Regret$(\pi)$, it is sufficient to upper bound Myopic$_t(\pi,\bc)$ for each $t\in[T]$ and each $\bc\geq0$. We summarize the above arguments in the following lemma.
\begin{lemma}\label{lem:upperregret}
For any feasible online policy $\pi$, the regret is upper bounded by
\[
\text{Regret}(\pi)\leq \sum_{t=1}^T\mathbb{E}_{\tilde{\bc}^{\pi}_t}\left[\text{Myopic}_t(\pi,\tilde{\bc}_t^{\pi})\right]
\]
where the myopic term $\text{Myopic}_t(\pi,\tilde{\bc}_t^{\pi})$ is defined in \eqref{def:myopicregret}.
\end{lemma}
We now motivate our policy $\pi$ such that the myopic term Myopic$_t(\pi,\bc)$ can be minimized for each $\bc$. Now suppose that the online decision maker is allowed to ``foresee'' the sample path $I_t$ and we denote by 
\[
M_{\bc, \tba_t}(I_{t+1})=\bV_{\bc}(I_{t+1})-\bV_{\bc-\tba_t}(I_{t+1})
\]
the marginal increase for the relaxation $\bV$ to have an extra $\tba_t$ resources from period $t+1$ to $T$. Clearly, in order to minimize $\text{Myopic}_t(\pi,\bc)$ in \eqref{def:myopicregret}, we set $\tx^{\pi}_t=1$ if and only if
\[
\bV_{\bc}(I_{t})-\bV_{\bc-\tba_t}(I_{t+1})-\tr_t\leq \bV_{\bc}(I_{t})-\bV_{\bc}(I_{t+1})
\]
which implies that
\[
\tx^{\pi}_t=\left\{\begin{aligned}
&1, &\text{if~}\tr_t\geq M_{\bc,\tba_t}(I_{t+1})\\
&0, &\text{if~}\tr_t< M_{\bc,\tba_t}(I_{t+1}).
\end{aligned}\right.
\]
However, note that in order for $\pi$ to be feasible, $\tx^{\pi}_t$ must be independent of $I_{t+1}$. Therefore, instead of comparing $\tr_t$ to the marginal increase $M_{\bc, \tba_t}(I_{t+1})$, we compare $\tr_t$ to an estimator $\hat{M}_{\bc, \tba_t}$ that is independent of $I_{t+1}$. Our policy is formalized in \Cref{alg:Mestimator}, which takes as an input an exogenous estimator $\hat{M}$ that we further specify in the following sections on different settings. 
In what follows, we first specify the relaxation $\bV$ that will be used in this section, and then we specify the $\hM$-estimator for our algorithm and derive the corresponding regret bound.

\begin{algorithm}[ht!]
\caption{$\hat{M}$-estimator policy ($\pi_{\hat{M}}$)}
\label{alg:Mestimator}
\begin{algorithmic}[1]
\State Input: an estimator $\hat{M}$.
\State Initialize the initial capacities $\bm{c}_1=\bm{C}$.
\For {$t=1,..., T$}
\State Observe the value of $(\tr_t,\tba_t)$ and obtain the value of $\hat{M}_{\bm{c}_t,\tba_t}$.
\State if $\tr_t\geq\hat{M}_{\bm{c}_t,\tba_t}$ and $\mathbf{c}_t\geq \tba_t$, then we set $x_t=1$ and update $\bc_{t+1}=\bc_t-\tba_t$;
\State otherwise, set $x_t=0$ and update $\bc_{t+1}=\bc_t$.
\EndFor
\State Output: online decisions $\bm{x} = (x_1,...,x_T)$.
\end{algorithmic}
\end{algorithm}

\subsection{Semi-fluid Relaxation}
We now specify the semi-fluid relaxation $\bV$ that will be used in this section. For each $j\in[n]$, we let {$d_j$ denote a generic non-negative integer that should be interpreted as
the number of type $j$ query arrivals remaining, regardless of the current time period. Meanwhile, we let $\tilde{d}_{j,t}$ be the random variable} for the number of type $j$ query arrivals from period $t$ to period $T$, i.e., the number of times that $\tba_{\tau}=\ba_j$ for $\tau=t,\dots,T$.
We introduce $\bV$ is as follows, for a fixed $\bd=(d_1,\dots, d_n)\in\mathbb{R}^n$: 
\begin{eqnarray}
\bVH_{\bc}(\bd)=&\max_{\bm{x}} &\sum_{j=1}^n d_{j}\cdot\mathbb{E}_{r\sim F_j}[r\cdot x_j(r)] \label{lp:Prophrelax}\\
&\mbox{s.t.} & \sum_{j=1}^n d_{j}\cdot a_{j,i}\cdot \mathbb{E}_{r\sim F_j}[ x_j(r)]\leq c_i,~~\forall i\in[m]\nonumber\\
&& x_j(r)\in[0,1],~~\forall j\in[n], \forall r\in[l_j, u_j]. \nonumber
\end{eqnarray}
In the following lemma, we show that the formulation of $\bV$ introduced in \eqref{lp:Prophrelax} implies an upper bound of the offline optimum $V^{\text{off}}$ in \eqref{lp:offlineoptimum}.
\begin{lemma}\label{lem:UB1}
It holds that $\mathbb{E}_{I}[\bVH_{\bm{C}}(\tilde{\bd}_1)]\geq\mathbb{E}_I[V^{\text{off}}(I)]$, where $\tilde{\bd}_1=(\tilde{d}_{1,1},\dots, \tilde{d}_{n,1})$ depends on the sample path $I$.
\end{lemma}
In order to see that $\bVH_{\bm{C}}(\tilde{\bd}_1)$ is an upper bound, we first fix the type arrivals of the queries that is implied by the sample path $I$. We then take an ex-ante relaxation over the reward distribution for each type.

\noindent\textbf{Comparison with other relaxations.} There are also other relaxations of the prophet \eqref{lp:offlineoptimum} existing in the literature and we now compare $\bVH_{\bc}(\bd)$ \eqref{lp:Prophrelax} with them. 

The most natural relaxation of the prophet \eqref{lp:offlineoptimum} is an LP relaxation, which is defined for each $t\in[T]$, any $\bc\geq0$, and any sample path $I$.
\begin{eqnarray}
\bV^{\text{Off}}_{\bc}(I_t)=&\max_{\bm{x}} & \sum_{\tau=t}^T \tr_{\tau}\cdot x_{\tau} \label{lp:relaxoffline}\\
&\mathrm{s.t.\ } & \sum_{\tau=t}^T \ta_{\tau,i}\cdot x_{\tau} \leq c_{i},~~\forall i\in[m] \nonumber\\
& &  x_{\tau}\in[0,1] \quad \quad \forall \tau=t,\dots,T. \nonumber
\end{eqnarray}
The only difference between the formulation of $V^{\text{off}}(I)$ and $\bVL_{\bm{C}}(I_1)$ is that the integral decision variables of $V^{\text{off}}(I)$ are relaxed to be fractional in $\bVL_{\bm{C}}(I_1)$.

Another common relaxation in the literature is the so-called ex-ante relaxation, which can be obtained {from replacing $\tilde{\bd}_t$ by its expectation in the formulation of $\bVH_{\bm{c}}(\tilde{\bd}_t)$}. For any $t\in[T]$ and any $\bc$, we denote by $\bVF_{t,\bc}$ the ex-ante relaxation with a formulation given as follows:
\begin{eqnarray}
\bVF_{t,\bc}=&\max_{\bm{x}} &\sum_{j=1}^n p_j\cdot s\cdot\mathbb{E}_{r\sim F_j}[r\cdot x_j(r)] \label{lp:exante1}\\
&\mbox{s.t.} & \sum_{j=1}^n p_j\cdot s\cdot a_{j,i}\cdot \mathbb{E}_{r\sim F_j}[ x_j(r)]\leq c_i,~~\forall i\in[m]\nonumber\\
&& x_j(r)\in[0,1],~~\forall j\in[n], \forall r\in[l_j, u_j] \nonumber
\end{eqnarray}
where we denote by $s=T-t+1$ for notation brevity.

\subsection{Policy and Regret Analysis}
We now develop the estimator that will be used in \Cref{alg:Mestimator} and analyze the regret bound. It is easy to see that the optimal solution of \eqref{lp:Prophrelax} preserves a ``threshold'' property {as formalized in the following lemma, where the proof is relegated to \Cref{sec:missingpf}.}
\begin{lemma}\label{lem:threshold}
Denote by $\{x^*_j(r), \forall j\in[n], \forall r\}$ an optimal solution to \eqref{lp:Prophrelax}. Then, there exists a set of thresholds $\{\kappa_j\}_{j=1}^n$ such that it is optimal to set $x^*_{j}(r)=1$ if and only if $r\geq \kappa_j$ and $x^*_{j}(r)=0$ if and only if $r<\kappa_j$, for any $j\in[n]$. 
\end{lemma}
Note that Topkis' theorem \citep{topkis1978minimizing} can also be used to derive \Cref{lem:threshold}. Following \Cref{lem:threshold}, $\bVH_{\bc}(\bd)$ in \eqref{lp:Prophrelax} can be re-written into the following formulation:
\begin{eqnarray}
\bVH_{\bc}(\bd)=&\max_{\bm{q}} &\sum_{j=1}^n d_{j}\cdot\int_{q=1-q_j}^{1}F_j^{-1}(q)  dq \label{lp:newProphrelax}\\
&\mbox{s.t.} & \sum_{j=1}^n d_{j}\cdot a_{j,i}\cdot q_j\leq c_i,~~\forall i\in[m]\nonumber\\
&& q_j\in[0,1],~~\forall j\in[n]. \nonumber
\end{eqnarray}
Here, the decision variable $q_j$ can be interpreted as the probability of serving type $j$ query, for each $j\in[n]$. Denote by $\{\tilde{q}^*_j\}$  one optimal solution to $\bVH_{\bc}(\tilde{\bd}_t)$. Since $\{\tilde{q}^*_j\}$ depends on the sample path $I_t$, clearly, one cannot directly use $\{\tilde{q}^*_j\}$ to derive a feasible online policy that is ``agnostic'' about $I_{t+1}$. Therefore, we will consider using the optimal solution of the ex-ante problem \eqref{lp:exante1}
to ``approximate'' $\{\tilde{q}^*_j\}$. It is clear to see that the optimal solution of the ex-ante problem \eqref{lp:exante1} also preserves a threshold property and $\bVF_{t,\bc}$ \eqref{lp:exante1} can be re-written into the following formulation: 
\begin{eqnarray}
\bVF_{t,\bc}=&\max &\sum_{j=1}^n p_j\cdot s\cdot\int_{q=1-q_j}^{1}F_j^{-1}(q)  dq \label{lp:exante}\\
&\mbox{s.t.} & \sum_{j=1}^n p_j\cdot s\cdot a_{j,i}\cdot q_j\leq c_i,~~\forall i\in[m]\nonumber\\
&& q_j\in[0,1],~~\forall j\in[n]. \nonumber
\end{eqnarray}
Note that the formulation of $\bVF_{t,\bc}$ deviates from the formulation of $\bVH_{\bc}(\tilde{\bd}_t)$ only in that the random variable $\tilde{d}_{j,t}$ is changed into its expectation $p_j\cdot s$. We can bound how this change of parameter would result in a change of the expected reward that we can gain from each type of query in the optimization problem \eqref{lp:newProphrelax} and \eqref{lp:exante}. Our analysis generalizes the Lipschitz analysis in \citet{mangasarian1987lipschitz} from linear programming to a general convex optimization problem. Our argument is formalized in the following lemma, where the proof is relegated to \Cref{sec:missingpf}.
\begin{lemma}\label{lem:perturbationLP}
There exists a constant $\kappa_1$ such that for any $\bc\geq0$ and any optimal solution $\hat{\bm{q}}^*=(\hat{q}^*_j)_{j=1}^n$ to \eqref{lp:exante}, it holds that
\begin{equation}\label{eqn:102601}
\|\hat{\bm{q}}^*-\tilde{\bm{q}}^*\|_{\infty}
\leq \kappa_1\cdot\max_{j\in[n]}\{|d_j/s-p_j|\}
\end{equation}
for any sample path $I_t$, where $s=T-t+1$, and $\tilde{\bm{q}}^*=(\tilde{q}^*_j)_{j=1}^n$ denotes one optimal solution to \eqref{lp:newProphrelax}. Moreover, the constant $\kappa_1$ can be set as follows,
\begin{equation}\label{def:kappa}
    \kappa_1=\max_{j\in[n]}\left\{\frac{1}{p_j}\right\} \cdot (m+2n)\cdot \bar{a}^n.
\end{equation}
\end{lemma}
The formal policy is given in \Cref{alg:M1}. 
We now provide the regret analysis.

\begin{algorithm}[]
\caption{Algorithm achieving $O(\log^2 T)$ Regret}
\label{alg:M1}
\begin{algorithmic}[1]
\State Input: the remaining inventory $\bc$, a constant $\kappa_1$ given in \eqref{def:kappa}, and the type of query $t$, denoted by $j_t$.
\State Obtain $\{\hat{q}^*_{j,t}\}$ by solving the optimization problem \eqref{lp:exante}.
\State \textbf{if} $\hat{q}^*_{j_t,t}\geq 1-2\kappa_1\cdot\sqrt{\frac{\log(T-t+1)}{T-t+1}}$, then we set $\hat{M}_{\bc, \ba_{j_t}}=l_{j_t}$.
\State \textbf{else if} $\hat{q}^*_{j_t,t}\leq2\kappa_1\cdot \sqrt{\frac{\log(T-t+1)}{T-t+1}}$, then we set $\hat{M}_{\bc, \ba_{j_t}}=u_{j_t}+1$.
\State \textbf{else if} $2\kappa_1\cdot\sqrt{\frac{\log(T-t+1)}{T-t+1}}\leq \hat{q}^*_{j_t,t}\leq 1-2\kappa_1\cdot\sqrt{\frac{\log(T-t+1)}{T-t+1}}$, then we set $\hat{M}_{\bc, \ba_{j_t}}=F^{-1}_{j_t}(1-\hat{q}^*_{j_t,t})$.
\State Output: $\hat{M}_{\bc, \ba_{j_t}}$
\end{algorithmic}
\end{algorithm}

\begin{theorem}\label{thm:Regretthm}
{Denote by $\pi$ \Cref{alg:Mestimator} with the estimator $\hat{M}$ given in \Cref{alg:M1}, and a constant $\kappa_1$ defined in \eqref{def:kappa}.} 
Then, it holds that
\[
\text{Regret}(\pi) \leq \left( \frac{2\kappa_1+2}{\alpha}+\frac{4}{\alpha}\cdot\sum_{j=1}^n\frac{1}{p_j}\right)\cdot\log^2 T + s_0\cdot r_{\max}.
\]
where $r_{\max}=\max_{j\in[n]}\{u_j\}$ denotes the upper bound of the reward, and $s_0$ is a constant that depends on $\kappa_1$, $n$ and $\{p_j\}_{j=1}^n$.
\end{theorem}
\begin{myproof}[Proof of \Cref{thm:Regretthm}]
We have that
\begin{equation}\label{eqn:09131}
\bVH_{\bc}(\tilde{\bd}_t)=\sum_{j=1}^n \td_{j,t+1}\cdot\int_{q=1-\tilde{q}^*_{j,t}}^{1} F^{-1}_j(q)dq+ \int_{q=1-\tilde{q}^*_{j_t,t}}^{1} F^{-1}_{j_t}(q)dq.
\end{equation}
where we denote by $j_t$ the type of query $t$ in the instance $I$.
Now we plug \eqref{eqn:09131} into the formulation \eqref{def:myopicregret} {where the term $\bV_{\bc}(I_t)$ is set to be $\bVH_{\bc}(\tilde{\bd}_t)$}. We get 
\begin{equation}\label{lp:newMyopic}
\begin{aligned}
\text{Myopic}_t(\pi,\bc)=&\mathbb{E}_{j_t, I_{t+1}}\left[ \sum_{j=1}^n \td_{j,t+1}\cdot\int_{q=1-\tilde{q}^*_{j,t}}^{1} F^{-1}_j(q)dq+ \int_{q=1-\tilde{q}^*_{j_t,t}}^{1} F^{-1}_{j_t}(q)dq\right.\\
&\left.-\int_{q=1-{q}^{\pi}_{j_t,t}}^{1}F^{-1}_{j_t}(q)dq-\mathbb{E}_{r\sim F_{j_t}}[\bVH_{\bc-\ba_{j_t}\cdot\tx_t^{\pi}(r)}(I_{t+1})] \right]    
\end{aligned}
\end{equation}
where we denote by $\pi$ our online policy \Cref{alg:Mestimator} with the estimator given in \Cref{alg:M1}. Then, ${q}^{\pi}_{j,t}$ denotes the ex-ante probability that query $t$ will be served by the online policy $\pi$. With these notations, \eqref{lp:newMyopic} can be re-written as
\[\begin{aligned}
\text{Myopic}_t(\pi,\bc)=&\mathbb{E}_{j_t, I_{t+1}}\left[ 
\sum_{j=1}^n \td_{j,t+1}\cdot \int_{q=1-\tilde{q}^*_{j,t}}^{1}F^{-1}_j(q)dq+\int_{q=1-\tilde{q}^*_{j_t,t}}^{ 1-{q}^{\pi}_{j_t,t} }F^{-1}_{j_t}(q)dq \right.\\
&\left. -{q}^{\pi}_{j_t,t}\cdot\bVH_{\bc-\ba_{j_t}}(\tilde{\bd}_{t+1})-(1-{q}^{\pi}_{j_t,t})\cdot \bVH_{\bc}(\tilde{\bd}_{t+1})
\right]. 
\end{aligned}\]
We construct feasible solutions to $\bVH_{\bc-\ba_{j_t}}(\tilde{\bd}_{t+1})$ and $\bVH_{\bc}(\tilde{\bd}_{t+1})$ to upper bound the myopic regret. We identify a ``good'' event {as one in which} $\tilde{q}^*_{j,t}$ is close to $\hat{q}^*_{j,t}$ for each $j\in[n]$. Following \Cref{lem:perturbationLP}, we know that
\begin{equation}\label{eqn:101201}
|\tilde{q}^*_{j,t}-\hat{q}^*_{j,t}|\leq\kappa_1\cdot\max_{j'}\{|p_{j'}-\td_{j',t+1}/(s-1)|\}
\end{equation}
for a constant $\kappa_1>0$.
We note that for each $j\in[n]$, $\td_{j,t+1}$ is a binomial distribution with mean $p_j\cdot(s-1)$. Then, from Hoeffding's inequality (\Cref{lem:Hoeffding}), we have
\[
P(|\td_{j,t+1}- p_j(s-1)|\leq\sqrt{(s-1)\log(s-1)})\geq 1-2\exp(-2\log(s-1))=1-\frac{2}{(s-1)^2}\geq1-\frac{1}{n(s-1)}
\]
as long as $s\geq s_0$ for a constant $s_0\geq2n+1$. We denote by the event 
\[
\mathcal{G}=\{|\td_{j,t+1}- p_j(s-1)|\leq\sqrt{(s-1)\log(s-1)}, \forall j\in[n]\},
\]
which is the "good" event that $\td_{j,t+1}$ is close to its mean. From union bound, we know that
\[
P(\mathcal{G})\geq 1-\sum_{j=1}^n(1-P(|\td_{j,t+1}- p_j(s-1)|\leq\sqrt{(s-1)\log(s-1)}))\geq 1-\frac{1}{s-1}.
\]
We denote by
\begin{equation}\label{eqn:231801}
\begin{aligned}
\text{Myopic}_t(\pi,\bc,\mathcal{G})=&P(\mathcal{G})\cdot\mathbb{E}_{j_t, I_{t+1}}\left[ 
\sum_{j=1}^n \td_{j,t+1}\cdot \int_{q=1-\tilde{q}^*_{j,t}}^{1}F^{-1}_j(q)dq+\int_{q=1-\tilde{q}^*_{j_t,t}}^{ 1-{q}^{\pi}_{j_t,t} }F^{-1}_{j_t}(q)dq \right.\\
&\left. -{q}^{\pi}_{j_t,t}\cdot\bVH_{\bc-\ba_{j_t}}(\tilde{\bd}_{t+1})-(1-{q}^{\pi}_{j_t,t})\cdot \bVH_{\bc}(\tilde{\bd}_{t+1})\mid\mathcal{G}
\right] 
\end{aligned}
\end{equation}
and
\[\begin{aligned}
\text{Myopic}_t(\pi,\bc,\mathcal{G}^c)=&P(\mathcal{G}^c)\cdot\mathbb{E}_{j_t, I_{t+1}}\left[ 
\sum_{j=1}^n \td_{j,t+1}\cdot \int_{q=1-\tilde{q}^*_{j,t}}^{1}F^{-1}_j(q)dq+\int_{q=1-\tilde{q}^*_{j_t,t}}^{ 1-{q}^{\pi}_{j_t,t} }F^{-1}_{j_t}(q)dq \right.\\
&\left. -{q}^{\pi}_{j_t,t}\cdot\bVH_{\bc-\ba_{j_t}}(\tilde{\bd}_{t+1})-(1-{q}^{\pi}_{j_t,t})\cdot \bVH_{\bc}(\tilde{\bd}_{t+1})\mid\mathcal{G}^c
\right] 
\end{aligned}\]
where $\mathcal{G}^c$ is the complement of $\mathcal{G}$. It is clear that
\[
\text{Myopic}_t(\pi,\bc)=\text{Myopic}_t(\pi,\bc,\mathcal{G})+\text{Myopic}_t(\pi,\bc,\mathcal{G}^c).
\]
We have a direct upper bound 
\begin{equation}\label{eqn:100601}
\text{Myopic}_t(\pi,\bc,\mathcal{G}^c)\leq\frac{\max_{j\in[n]}\{u_j\}}{s-1}.
\end{equation}
In what follows, we condition on the event $\mathcal{G}$ happening, and we bound $\text{Myopic}_t(\pi,\bc,\mathcal{G})$.

\noindent\textbf{Case (i) when $\hat{q}^*_{j_t,t}\geq 1-2\kappa_1\cdot\sqrt{\frac{\log s}{s}}$:} Note that conditioning on the event $\mathcal{G}$, following \eqref{eqn:101201}, we have
\[
|\tilde{q}^*_{j_t,t}-\hat{q}^*_{j_t,t}|\leq \kappa_1\cdot \sqrt{\frac{\log(s-1)}{s-1}}.
\]
Such a case implies that $\tilde{q}^*_{j_t,t}\geq 1-3\kappa_1\cdot\sqrt{\frac{\log s}{s}}\geq \frac{1}{2}$ when $s\geq s_0$ for a constant $s_0$ satisfying $\frac{s_0}{\log s_0}\geq36\kappa_1^2$. Recall that $\tilde{q}^*_{j_t, t}$ denotes the quantile for query $t$ to be accepted by the offline optimum. We know that query $t$ of type $j_t$ should also be accepted by our algorithm in a high quantile. In fact, since $\tilde{q}^*_{j_t,t}\geq 1-O\left(\sqrt{\frac{\log s}{s}}\right)$, we can set the quantile $q^{\pi}_{j_t, t}$ of our online algorithm to be any value at the order of $1-O\left(\sqrt{\frac{\log s}{s}}\right)$, and the order of the myopic regret will be bounded at the order of $O\left(\sqrt{\frac{\log s}{s}}\right)$. For simplicity, we set $q^{\pi}_{j_t, t}=1$ and we summarize our results for Case (i) in the following lemma, where the proof is relegated to \Cref{sec:missingpf}. 

\begin{lemma}\label{lem:proof1}
as long as $s\geq s_0$ for $s_0$ satisfying $\frac{s_0}{\log s_0}\geq16\kappa_1^2$ and $s_0\geq2\cdot\max_{j\in[n]}\{\frac{1}{p_j}\}$, we can obtain the following result.
\begin{equation}\label{eqn:091804}
\begin{aligned}
\text{Myopic}_t(\pi,\bc,\mathcal{G})\leq \frac{2\log s}{\alpha\cdot s}+\frac{2}{\alpha}\cdot\mathbb{E}_{j_t, I_{t+1}}[(\tilde{q}^*_{j_t,t}-\hat{q}^*_{j_t,t})^2]+\frac{2}{\alpha}\cdot\mathbb{E}_{j_t, I_{t+1}}\left[\frac{1}{\td_{j_t, I_{t+1}}}\mid\mathcal{G}\right].
\end{aligned}
\end{equation}
\end{lemma}

\noindent\textbf{Case (ii) when $\hat{q}^*_{j_t,t}\leq2\kappa_1\cdot\sqrt{\frac{\log s}{s}}$:} Note that conditioning on the event $\mathcal{G}$, following \eqref{eqn:101201}, we have
\[
|\tilde{q}^*_{j_t,t}-\hat{q}^*_{j_t,t}|\leq \kappa_1\cdot \sqrt{\frac{\log(s-1)}{s-1}}
\]
which implies $\tilde{q}^*_{j_t, t}\leq 3\kappa_1\cdot\sqrt{\frac{\log s}{s}}\leq\frac{1}{2}$ as long as $s\geq s_0$ for a constant $s_0$ satisfying $\frac{s_0}{\log s_0}\geq36\kappa_1^2$. We know that query $t$ will be rejected by the offline optimum with a high probability. Therefore, we can simply set ${q}^{\pi}_{j_t,t}=0$ such that query $t$ is rejected by our online algorithm. Following such a way, we obtain the following result, where the formal proof is relegated to \Cref{sec:missingpf}.
\begin{lemma}\label{lem:proof2}
As long as $s\geq s_0$ for a constant $s_0$ satisfying $\frac{s_0}{\log s_0}\geq36\kappa_1^2$, we know that
\begin{equation}\label{eqn:091808}
\begin{aligned}
\text{Myopic}_t(\pi,\bc,\mathcal{G})\leq \frac{2\log s}{\alpha\cdot s}+\frac{2}{\alpha}\cdot\mathbb{E}_{j_t, I_{t+1}}[(\tilde{q}^*_{j_t,t}-\hat{q}^*_{j_t,t})^2]+\frac{2}{\alpha}\cdot\mathbb{E}_{j_t, I_{t+1}}\left[\frac{1}{\td_{j_t, t+1}}\mid\mathcal{G}\right].
\end{aligned}\end{equation}
\end{lemma}

\noindent\textbf{Case (iii) when $2\kappa_1\cdot\sqrt{\frac{\log s}{s}}\leq \hat{q}^*_{j_t,t}\leq 1-2\kappa_1\cdot\sqrt{\frac{\log s}{s}}$:} Note that conditioning on the event $\mathcal{G}$, following \eqref{eqn:101201}, we have
\[
|\tilde{q}^*_{j_t,t}-\hat{q}^*_{j_t,t}|\leq \kappa_1\cdot \sqrt{\frac{\log(s-1)}{s-1}}
\]
which implies $\kappa_1\cdot\sqrt{\frac{\log s}{s}}\leq \tilde{q}^*_{j_t,t}\leq1-\kappa_1\cdot\sqrt{\frac{\log s}{s}}$.
In this case, query $t$ is accepted by the offline optimum with a probability not too close to either $0$ or $1$. We mimic this point in our online algorithm by setting $q^{\pi}_{j_t, t}=\hat{q}^*_{j_t,t}$ and we obtain the following result, where the formal proof is relegated to \Cref{sec:missingpf}.
\begin{lemma}\label{lem:proof3}
As long as $s\geq s_0$ for a constant $s_0$ satisfying $s_0\log s_0\geq \frac{4}{\kappa_1^2}\cdot\max_{j\in[n]}\{\frac{1}{p_j^2}\}$ and $\frac{s_0}{\log s_0}\geq\max_{j\in[n]}\{\frac{4}{p_j^2}\}$, we have that
\begin{equation}\label{eqn:091811}
\begin{aligned}
\text{Myopic}_t(\pi,\bc,\mathcal{G})\leq\mathbb{E}_{j_t, I_{t+1}}\left[ \frac{1}{\alpha\cdot d_{j_t, t+1}}+\frac{(\tilde{q}^*_{j_t,t}-\hat{q}^*_{j_t,t})^2}{\alpha}\mid\mathcal{G} \right]\cdot P(\mathcal{G}).
\end{aligned}
\end{equation}
\end{lemma}

From \eqref{eqn:091804}, \eqref{eqn:091808} and \eqref{eqn:091811}, for all cases, it holds that
\[\begin{aligned}
\text{Myopic}_t(\pi,\bc,\mathcal{G})&\leq \frac{2\log s}{\alpha\cdot s}+\frac{2}{\alpha}\cdot\mathbb{E}_{j_t, I_{t+1}}[(\tilde{q}^*_{j_t,t}-\hat{q}^*_{j_t,t})^2]+\frac{2}{\alpha}\cdot\mathbb{E}_{j_t, I_{t+1}}\left[\frac{1}{\td_{j_t, I_{t+1}}}\mid\mathcal{G}\right]\\
&\leq \frac{2\log s}{\alpha\cdot s}+\frac{2}{\alpha}\cdot\mathbb{E}_{j_t, I_{t+1}}[(\tilde{q}^*_{j_t,t}-\hat{q}^*_{j_t,t})^2]+\frac{4}{\alpha(s-1)}\cdot\sum_{j=1}^n\frac{1}{p_j}.
\end{aligned}\]
From \Cref{lem:perturbationLP}, we know that there exists a constant $\kappa_1$ with formulation given in \eqref{def:kappa} such that
\[
\mathbb{E}_{j_t, I_{t+1}}[(\tilde{q}^*_{j_t,t}-\hat{q}^*_{j_t,t})^2]\leq \kappa_1\cdot\sum_{j=1}^n \mathbb{E}[(p_j-\td_{j,t}/s)^2]\leq\frac{\kappa_1}{s}.
\]
Therefore, as long as $s\geq s_0$ for a constant $s_0$ satisfying
\[
s_0\geq 2n+1,~\frac{s_0}{\log s_0}\geq 36\kappa_1^2,~s_0\geq\max_{j\in[n]}\{\frac{2}{p_j}\},~s_0\cdot\log s_0\geq\frac{4}{\kappa_1^2}\cdot\max_{j\in[n]}\{\frac{1}{p_j^2}\}\text{~and~}\frac{s_0}{\log s_0}\geq\max_{j\in[n]}\{\frac{4}{p_j^2}\},
\]
we have that
\[
\text{Myopic}_t(\pi,\bc,\mathcal{G})\leq \frac{2\log s}{\alpha\cdot s}+\frac{2\kappa_1}{\alpha\cdot s}+\frac{4}{\alpha(s-1)}\cdot\sum_{j=1}^n\frac{1}{p_j}.
\]
Together with \eqref{eqn:100601}, we have the following regret bound over the final regret bound of our algorithm,
\[
\text{Regret}(\pi)\leq \left( \frac{2\kappa_1+2}{\alpha}+\frac{4}{\alpha}\cdot\sum_{j=1}^n\frac{1}{p_j}\right)\cdot\log^2 T + s_0\cdot r_{\max}.
\]
Our proof is completed.
\end{myproof}


\section{Policy with Logarithmic Regret under Second-order Growth} \label{sec:Logsection}
In this section, we derive an improved logarithmic regret bound for \eqref{eqn:regret} under the following stronger assumption.

\begin{assumption}\label{assump:contiknown2}
There exists a compact convex set $\Omega\subset\mathbb{R}^m_{\geq0}$ such that for any $t\in[T]$, any $\bc$ and any problem instance $I_t$, the relaxed offline optimum $\bV^{\text{Off}}_{\bc}(I_t)$ \eqref{lp:relaxoffline} possesses one optimal dual solution $\tbmu$ satisfying $\tbmu\in\Omega$. Moreover, there exists two positive constants $\underline{\alpha}, \bar{\alpha}$ such that for any $\bmu', \bmu''\in\Omega$, it holds that
\begin{equation}\label{cond:contiknown2}
\underline{\alpha}\cdot\mathbb{E}_{\tba}\left[(\tba^\top\bmu'-\tba^\top\bmu'')^2\right]\leq\mathbb{E}_{\tba}\left[ \left(F(\tba^\top\bmu'|\tba)-F(\tba^\top\bmu''|\tba)\right)\cdot (\tba^\top\bmu'-\tba^\top\bmu'')\right]\leq\bar{\alpha}\cdot\mathbb{E}_{\tba}\left[(\tba^\top\bmu'-\tba^\top\bmu'')^2\right]
\end{equation}
where $\tba$ is a random variable denoting the size of one query and is realized as $\ba_j$ with probability $p_j$, for $j\in[n]$.
\end{assumption}
We now provide an example to illustrate under which condition \Cref{assump:contiknown2} holds or not. 

\begin{example}
Consider a special case of our model where there is a single resource, i.e., $m=1$, and two types of queries, i.e., $n=2$. For each type $j=1\text{~or~}2$, the size $a_j=1$ and the reward distribution is a uniform distribution over the interval $[l_j, u_j]$ with $u_1 \leq u_2$. Note that \Cref{assump:contiknown2} essentially requires that $u_1\geq l_2$, i.e., the support for the reward distribution of each type of query overlaps with each other. In order to see this point, the set $\Omega$ can be specified as $[l_1, u_2]$ and it is clear to see that \eqref{cond:contiknown2} will be satisfied. 
In contrast, if $u_1< l_2$, then we can see that \eqref{cond:contiknown2} will be violated by setting $\mu'=l_2$ and $\mu''=u_1$. However, the results derived in \Cref{sec:knowndistribution} will still apply and a $O(\log^2 T)$ regret bound can be obtained.
\end{example}

The only reason why we make \Cref{assump:contiknown2} is to establish a second-order growth condition of a dual function (we formalize in \Cref{lem:secondorder}), which is a standard condition in the stochastic programming literature. Then, we apply results from the stochastic programming literature to derive our logarithmic regret bound (we formalize in \Cref{lem:boundIandII}). Note that the second-order growth condition has been assumed frequently in the previous literature (e.g. \cite{bray2022logarithmic, li2021online, balseiro2021survey}) {in various formulations. To be specific, conditions on the density function are imposed to establish second-order growth, as in Assumption 2 in \cite{li2021online}. \cite{balseiro2021survey} directly assumes the strong-convexity of the dual function while \cite{bray2022logarithmic} assumes the Jacobian matrix of the dual function to be full rank and continuous in a neighborhood. All these conditions ensure the second-order growth condition}. By deriving our results simply under \Cref{assump:contiknown2}, our contribution would be to get rid of a so-called ``non-degeneracy'' assumption, which concerns the ``position'' of the ex-ante relaxation $\bVF_{1,\bm{C}}$ and is different from the second-order growth condition. In the following part, we briefly illustrate the ``non-degeneracy'' and we provide a thorough comparison between our \Cref{assump:contiknown2} and the assumptions made in previous literature in \Cref{sec:conclusion}.

\noindent\textbf{Comparison with assumptions made in previous literature.} Notably, a common assumption made in the previous literature \citep{balseiro2021survey, li2021online, bray2022logarithmic} regarding logariathmic regret with continuous reward distribution is about the ``non-degeneracy'' of the ex-ante relaxation $\bVF_{1,\bm{C}}$. The ``non-degeneracy'' assumption not only requires the optimal solution to $\bVF_{1,\bm{C}}$ to be unique, but also requires \textit{strict complementary slackness} condition to be satisfied by $\bVF_{1,\bm{C}}$, i.e., for any binding resource constraint in the optimal solution of $\bVF_{1,\bm{C}}$, the corresponding optimal dual variable must be strictly positive. In this way, the optimal basis of $\bVF_{1,\bm{C}}$ will remain unchanged if $\bm{C}$ is perturbed by a certain amount, which drives all the analysis in \citet{balseiro2021survey, li2021online, bray2022logarithmic}. In contrast, our \Cref{assump:contiknown2} simply requires the support of the reward distribution of each type to ``overlap'' with each other, as shown in Example 1, and our goal of \Cref{assump:contiknown2} is to establish the standard second-order growth condition of the dual function. We require nothing over the ``position'' of the ex-ante relaxation $\bVF_{1,\bm{C}}$, including unique optimal solution condition, strict complementary slackness condition, and so on.

We now provide an example where the strict complementary slackness condition of the ex-ante relaxation $\bVF_{1,\bm{C}}$ is violated, while our \Cref{assump:contiknown2} is still satisfied.

\begin{example} \label{ex:2}
Consider a model with 3 resources, each with an initial capacity of $T\cdot\frac{5\eps(1+\eps)}{2(1+6\eps)}$. There are 4 types of queries, denoted by $j=1,2,3, 4$. The size of each type of query is $\ba_1=(0,1,1)$, $\ba_2=(1,0,1)$, $\ba_3=(1,1,0)$ and $\ba_4=(1,1,1)$. The reward distribution of types 1, 2, and 3 of the query is a uniform distribution over $[0,1]$, and the reward distribution of type 4 of the query is a uniform distribution over $[0,2]$. The arrival probability is $p_1=\frac{2\eps}{1+6\eps}$, $p_2=\frac{1+\eps}{1+6\eps}$, $p_3=\frac{2\eps}{1+6\eps}$ and $p_4=\frac{\eps}{1+6\eps}$. 
\end{example}

We can show that our \Cref{assump:contiknown2} is satisfied in \Cref{ex:2} (formalized in \Cref{lem:Example}). On the other hand, following \eqref{lp:exante}, the ex-ante relaxation $\bVF_{1,\bm{C}}/T$ can be formulated as:
\begin{eqnarray}
&\max &\sum_{j=1}^3 p_j\cdot q_j\cdot(1-\frac{q_j}{2})+p_4\cdot q_4\cdot(2-q_4)\label{lp:Example2}\\
&\mbox{s.t.} & p_2q_2+p_3q_3+p_4q_4\leq \frac{5\eps(1+\eps)}{2(1+6\eps)}\nonumber\\
& & p_1q_1+p_3q_3+p_4q_4\leq \frac{5\eps(1+\eps)}{2(1+6\eps)}\nonumber\\
& & p_1q_1+ p_2q_2+p_4q_4\leq \frac{5\eps(1+\eps)}{2(1+6\eps)}\nonumber\\
& & q_1, q_2, q_3, q_4\in[0,1]. \nonumber
\end{eqnarray}
Denote by $\mu_i$ the dual variable for constraint for resource $i$. Then, we can show that the primal-dual pair $\mu_1^*=\mu^*_3=\frac{1-\eps}{2}$, $\mu^*_2=0$ and $q^*_1=q^*_3=q^*_4=\frac{1+\eps}{2}$, $q^*_2=\eps$ is optimal to \eqref{lp:Example2} by checking that the saddle point condition is satisfied, for any $\eps>0$. However, while the resource constraint is binding for every $i=1,2,3$, the optimal dual variable $\mu^*_2=0$, which shows that the \textit{strict complementary slackness} condition is not satisfied for every $\eps>0$. Therefore, the ``non-degeneracy'' assumption is violated for \eqref{lp:Example2}. The above argument is formalized in the following lemma, with the formal proof relegated to \Cref{sec:missingpf2}.
\begin{lemma}\label{lem:Example}
The problem instance described in \Cref{ex:2} satisfies \Cref{assump:contiknown2}, while the strict complementary slackness condition is violated by the corresponding ex-ante relaxation in \eqref{lp:Example2}.
\end{lemma}

\subsection{Decomposition of the Myopic Regret}\label{sec:Decomposition}
We still follow the general approach described in \Cref{sec:GDescription}. Following \Cref{lem:upperregret}, we proceed to bound the myopic regret Myopic$_t(\pi,\tbc_t^{\pi})$ in \eqref{def:myopicregret} under \Cref{assump:finitesize} and \Cref{assump:contiknown2}. The difference from \Cref{sec:knowndistribution} is that we now use $\bVL_{\tbc^{\pi}_t}(I_t)$ \eqref{lp:relaxoffline}, which is a LP relaxation of $V^{\text{off}}(I)$ \eqref{lp:offlineoptimum}, to serve as the benchmark $\bar{V}$ and thus provides more tractablity. Now the definition of Myopic$_t(\pi,\tbc_t^{\pi})$ becomes
\[
\text{Myopic}_t(\pi,\bc)=\mathbb{E}_{\pi,I_t}\left[\bVL_{\bc}(I_t)-\bVL_{\bc-\tba_t\cdot \tx^{\pi}_t}(I_{t+1})-\tr_t\cdot \tx^{\pi}_t\right].
\]
We denote by $\{\tx^*_{\tau}\}$ one optimal solution of $\bVL_{\tbc^{\pi}_t}(I_t)$ \eqref{lp:relaxoffline}, where $\tx^*_{\tau}\in[0,1]$ for each $\tau=t,\dots,T$. Then, a gap arises from the fact that the online decision for our policy $\pi$ must be binary as required by problem formulation, while the optimal solution $\tx^*_t$ in the relaxation \eqref{lp:relaxoffline} can be fractional. In order to deal with this gap, we introduce a ``rounded'' relaxed offline optimum as an intermediate. To be specific, we denote by
\begin{equation}\label{def:roundedsolution}
\tx^{\text{round}}_t=\left\{\begin{aligned}
&1, &&\text{if~}\tr_t\geq M_{\tbc^{\pi}_t,\tba_t}(I_{t+1})= \bVL_{\tbc_t^{\pi}}(I_{t+1})-\bVL_{\tbc_t^{\pi}-\tba_t}(I_{t+1})\text{~and~}\tbc^{\pi}_t\geq \tba_t\\
&0, &&\text{otherwise},
\end{aligned}\right.
\end{equation}
as a rounding of $\tx^*_t$.
Then, we have that 
\begin{equation}\label{eqn:roundViterate}
\bVL_{\tbc_t^{\pi}}(I_t)=\tx^{\text{round}}_t\cdot(\tr_t+\bVL_{\tbc_t^{\pi}-\tba_t}(I_{t+1}))+(1-\tx^{\text{round}})\bVL_{\tbc_t^{\pi}}(I_{t+1})+G_{\tbc^{\pi}_t}(I_t)
\end{equation}
where $G_{\tbc^{\pi}_t}(I_t)$ denotes the gap caused by rounding $\tx^*_t$ to be $\tx^{\text{round}}_t$, which can be formulated as follows
\begin{equation}\label{eqn:newroundedgap}
    G_{\tbc^{\pi}_t}(I_t)=\tr_t(\tx^*_t-\tx^{\text{round}}_t)+\bVL_{\tbc_t^{\pi}-\tba_t \tx^*_t}(I_{t+1})-\left(\tx^{\text{round}}_t \bVL_{\tbc_t^{\pi}-\tba_t}(I_{t+1}))+ (1-\tx^{\text{round}}_t) \bVL_{\tbc_t^{\pi}}(I_{t+1})\right).
\end{equation}
Introducing the rounded gap $G_{\tbc^{\pi}_t}(I_t)$ allows us to further decompose the regret of our policy into three terms, as formalized in the following lemma, where the proof is relegated to \Cref{sec:missingpf2}.
\begin{lemma}\label{lem:decompose}
For any $t\in[T]$, {under \Cref{assump:finitesize} and \Cref{assump:contiknown2},} it holds that
\begin{equation}\label{eqn:derivation03}
\begin{aligned}
\text{Myopic}_t(\pi,\tbc^{\pi}_t)\leq&2\bar{\alpha}\cdot\mathbb{E}_{\tba_t}[\bI_{\{\tbc^{\pi}_t\geq \tba_t\}}\cdot\Var(M_{\tbc^{\pi}_t,\tba_t}(I_{t+1}))]+\mathbb{E}_{I_t}[G_{\tbc_t^{\pi}}(I_t)]\\
&+2\bar{\alpha}\cdot\mathbb{E}_{\tba_t}\left[\bI_{\{\tbc^{\pi}_t\geq \tba_t\}}\cdot\left(\hat{M}_{\tbc^{\pi}_t,\tba_t}-\mathbb{E}_{I_{t+1}}[M_{\tbc^{\pi}_t,\tba_t}(I_{t+1})]\right)^2\right]
\end{aligned}
\end{equation}
where the variance $\Var$ is taken over the problem instance $I_{t+1}$.
\end{lemma}
A key step in deriving \Cref{lem:decompose} is to utilize the relationship
\begin{equation}\label{eqn:101001}
\bVL_{\bc}(I_t)=\tr_t\cdot \tx^*_t+\bVL_{\bc-\tba_t\cdot \tx^*_t}(I_{t+1}).
\end{equation}
We note that the above backward induction holds on each sample path $I$, which is the reason why we use a sample-path based relaxed offline optimum  $\bVL_{\bc}(I_t)$ \eqref{lp:relaxoffline} as the benchmark. As we will show in the next section, the backward induction \eqref{eqn:101001} enables us to reduce bounding the myopic regret into bounding the dual convergence, without requiring an additional ``non-degeneracy'' assumption adopted in the previous literature (we further discuss in \Cref{sec:conclusion}).
This discussion reveals the benefits of considering a sample-path based benchmark $\bVL_{t,\bc}(I_t)$. The same idea of using the relaxation $\bVL_{\bc}(I_t)$ and relationship \eqref{eqn:101001} has been developed in \cite{vera2021bayesian} to get rid of the ``non-degeneracy'' assumption when the reward for each type is deterministic. We generalize this idea to allow the reward for each type having a continuous distribution.

For the RHS of \eqref{eqn:derivation03}, we refer the first term as the variation gap, the second term as the rounding gap, and the third term as the estimator gap. In what follows, we proceed to bound the three terms separately. Our goal is to show that each gap can be bounded at the order of $O(\frac{1}{T-t})$, and these bounds together will imply a $O(\log T)$ bound for \Cref{alg:Mestimator}. 

Our analysis relies on considering the dual problem of $\bVL_{\tbc^{\pi}_t}(I_{t+1})$ and $\bVL_{\tbc^{\pi}_t-\tba_t}(I_{t+1})$ that define $M_{\tbc^{\pi}_t,\tba_t}(I_{t+1})$ in \eqref{def:myopicregret}, for each fixed $\tbc^{\pi}_t$ and $\tba_t$ satisfying $\tbc^{\pi}_t\geq \tba_t$. Now for any $\bc\geq0$, we introduce a dual variable $\bmu$ for the constraints of $\bVL_{\bc}(I_{t+1})$ \eqref{lp:relaxoffline} and we denote by the function
\begin{equation}\label{eqn:Lagrangian}
\begin{aligned}
\LL_{\bc, I_{t+1}}(\mu):&=\max_{\tx_{\tau}\in[0,1],\forall \tau=t+1,\dots,T}\left(\frac{\bc}{s-1}\right)^{\top}\mu+\frac{1}{s-1}\cdot\sum_{\tau=t+1}^T[\tr_{\tau}-\tba_{\tau}^{\top}\mu]\cdot\tx_{\tau}\\
&=\left(\frac{\bc}{s-1}\right)^{\top}\mu+\frac{1}{s-1}\cdot\sum_{\tau=t+1}^T[\tr_{\tau}-\tba_{\tau}^{\top}\mu]^+.
\end{aligned}
\end{equation}
as the dual function of $\bVL_{\bc}(I_{t+1})$, scaled by $\frac{1}{s-1}$, with $s=T-t+1$. 
We now proceed to bound the variation gap, the rounding gap, and the estimator gap. In \Cref{sec:ReductDual}, we show that bounding the first two gaps can be reduced to bounding a so-called ``dual convergence'', which concerns the variance of one optimal dual variable for minimizing the dual function $\LL_{\bc, I_{t+1}}(\mu)$ \eqref{eqn:Lagrangian}. Then, we propose our $\hat{M}-$estimator and bound the ``dual convergence'' in \Cref{sec:BoundDual} to complete our final bound over the myopic regret.

\subsection{Reduction to Dual Convergence}\label{sec:ReductDual}
In this section, we show how to reduce bounding each term in \eqref{eqn:derivation03} to bounding the ``dual convergence'', and we also propose our $\hat{M}-$estimator. 

We first bound the variation gap $\Var(M_{\bc,\ba_t}(I_{t+1}))$ for each fixed $\bc$ and $\ba_t$ satisfying $\bc\geq \ba_t$. We denote by 
\begin{equation}\label{eqn:dualsolution}
\tbmu_1\in\argmin_{\bmu\in\Omega} \LL_{\bc,I_{t+1}}(\bmu) \text{~~and~~}\tbmu_2\in\argmin_{\bmu\in\Omega} \LL_{\bc-\ba_t,I_{t+1}}(\bmu).
\end{equation}
Note that $\tbmu_1$ (resp. $\tbmu_2$) is one optimal dual variable of $\bVL_{\bc}(I_{t+1})$ (resp. $\bVL_{\bc-\ba_t}(I_{t+1})$). Also, $\tbmu_1$ and $\tbmu_2$ are random variables, where the randomness comes from the randomness of the problem instance $I_{t+1}$. The goal of introducing $\tbmu_1$ and $\tbmu_2$ is to lower-bound and upper-bound $M_{\bc,\ba_t}(I_{t+1})$ in \eqref{def:myopicregret}, as in the following lemma.
\begin{lemma}\label{lem:dualbound}
For any problem instance $I_{t+1}$ and any $\bc\geq\ba_t$, it holds that
\begin{equation}\label{eqn:dualboundvariance}
\ba_t^{\top}\tmu_1\leq M_{\bc,\ba_t}(I_{t+1})=\bVL_{\bc}(I_{t+1})-\bVL_{\bc-\ba_t}(I_{t+1})\leq \ba_t^{\top}\tmu_2
\end{equation}
where $\tbmu_1$ and $\tbmu_2$ are defined in \eqref{eqn:dualsolution}.
\end{lemma}
We proceed to bound $\Var(M_{\bc,\ba_t}(I_{t+1}))$ with the help of \Cref{lem:dualbound}. We note that the function $\LL_{\bc,I_{t+1}}(\bmu)$ can be regarded as a sample average approximation of the following stochastic optimization problem:
\begin{equation}\label{eqn:limitingproblem}
   \min_{\bmu\in\Omega} \LF_{\bc,t+1}(\bmu):=\left(\frac{\bc}{s-1}\right)^{\top}\bmu+\mathbb{E}_{(\tr,\tba)\sim F}[\tr-\tba^{\top}\bmu]^+
\end{equation}
with $s-1=T-t$ samples. 
Our analysis relies on showing the second-order growth condition of the ``limiting'' dual function $\LF_{\bc,t+1}$ defined in \eqref{eqn:limitingproblem}. Denote by $\mathcal{S}$ the subspace spanned by the resource consumption vector $\ba$ of each query:
\begin{equation}\label{def:subspace}
    \mathcal{S}=\left\{\sum_{j\in[n]}\alpha_{j}\cdot \ba_j: \forall \alpha_{j}\in\mathbb{R} \right\}.
\end{equation}
As a result of \Cref{assump:contiknown2}, we have the following second-order growth condition of the limiting dual function $\LF_{\bc,t+1}$ for any $\bc\geq0$.
\begin{lemma}\label{lem:secondorder}
For any $\bc\geq0$, we denote by $\bmu^*\in\argmin~\LF_{\bc,t+1}(\bmu)$ such that $\bmu^*\in\Omega$ with $\Omega$ specified in \Cref{assump:contiknown2}. We also denote by $\mathcal{P}_S$ the projection of any vector to the subspace $\mathcal{S}$ (defined in \eqref{def:subspace}) spanned by the resource consumption vector $\ba$ of each query. Then for any $\bmu\in\Omega$, it holds that
\[
\LF_{\bc,t+1}(\bmu)-\LF_{\bc,t+1}(\bmu^*)\geq \frac{\underline{\alpha}\underline{\beta}}{2}\cdot\|\mathcal{P}_S(\bmu-\bmu^*)\|_2^2
\]
where $\underline{\beta}$ denotes the smallest positive eigenvalue of $\mathbb{E}_{\tba}[\tba\cdot \tba^\top]$. 
\end{lemma}
The proof is relegated to \Cref{sec:missingpf2}. Denote by 
\begin{equation}\label{eqn:limitingsolution}
\hbmu_1\in\argmin_{\bmu\in\Omega} \LF_{\bc,t+1}(\bmu)\text{~~and~~}\hbmu_2\in\argmin_{\bmu\in\Omega} \LF_{\bc-\ba_t,t+1}(\bmu).
\end{equation}
Following classical results from sample average approximation for stochastic programming \citep{shapiro1993asymptotic}, we know that $\tbmu_1$ (resp. $\tbmu_2$) converges to $\hbmu_1$ (resp. $\hbmu_2$) in probability, as $s-1=T-t\rightarrow\infty$. Thus, we can use $\hbmu_1$ (resp. $\hbmu_2$) as an approximation of $\mathbb{E}[\tbmu_1]$ (resp. $\mathbb{E}[\tbmu_2]$) and obtain a bound over $\Var(\tbmu_1)$ and $\Var(\tbmu_2)$, which finally implies an upper bound of $\Var(M_{\bc,\ba_t}(I_{t+1}))$. We summarize the above arguments in the following lemma, which shows that we can reduce bounding the variation gap to bounding the ``dual convergence'' terms $\mathbb{E}_{I_{t+1}}[(\ba_t^\top\tbmu_1-\ba_t^\top\hbmu_1)^2]$ and $\mathbb{E}_{I_{t+1}}[(\ba_t^\top\tbmu_2-\ba_t^\top\hbmu_2)^2]$. The proof is relegated to \Cref{sec:missingpf2}.
\begin{lemma}\label{lem:Decomposevariation}
Denote by $\bar{d}=\max_{j\in[n]}\{\|\ba_j\|_2\}$ and $\gamma=\max_{i\in[m], j\in[n]:a_{j,i}>0}\frac{u_j}{a_{j,i}}$. Then, for any $\bc\geq0$ and any $\ba_t$ satisfying $\bc\geq\ba_t$, it holds that 
\[
\Var(M_{\bc,\ba_t}(I_{t+1}))\leq \frac{12}{(s-1)\cdot\underline{\alpha}\underline{\beta}}\cdot\bar{d}^{3}\cdot m^{1/2}\cdot\gamma+ 14\cdot\mathbb{E}_{I_{t+1}}[(\ba_t^\top\tbmu_1-\ba_t^\top\hbmu_1)^2]+14\cdot\mathbb{E}_{I_{t+1}}[(\ba_t^\top\tbmu_2-\ba_t^\top\hbmu_2)^2]
\]
with $s=T-t+1$,
where $\tbmu_1$ and $\tbmu_2$ are defined in \eqref{eqn:dualsolution}, $\hbmu_1$ and $\hbmu_2$ are defined in \eqref{eqn:limitingsolution}, and $\mathbb{E}_{I_{t+1}}[\cdot]$ denotes taking expectation over $I_{t+1}$ that decides the value of $\tbmu_1$ and $\tbmu_2$.
\end{lemma}

We then reduce bounding the rounding gap (second term in \eqref{eqn:derivation03}) to bounding the ``dual convergence''. Following previous notations, we denote by $\{\tx^*_{\tau}\}_{\tau=t}^T$ an optimal solution to $\bVL_{\bc}(I_t)$ \eqref{lp:relaxoffline}. Note that $\tx^*_t\in[0,1]$ can be fractional. We denote by $\tx_t^{\text{round}}$
as the rounding of $\tx^*_t$ as in \eqref{def:roundedsolution}. We bound the rounding gap $\sum_{t=1}^T\mathbb{E}[G_{\tbc^{\pi}_t}(I_t)]$, where the formulation of $G_{\tbc^{\pi}_t}(I_t)$ is given in \eqref{eqn:newroundedgap}. Clearly, when $\tx^*_t=\tx_t^{\text{round}}$, $G_{\tbc^{\pi}_t}(I_t)=0$. Thus, the rounding gap arises from the fact that $\tx^*_t\in[0,1]$ can be different from $\tx_t^{\text{round}}\in\{0,1\}$. 

A key observation can be summarized as follows: i) if $\bc\geq \tba_t$, then both $\tbmu_1$ and $\tbmu_2$ are well-defined in \eqref{eqn:dualsolution} and $\tx^*_t\neq\tx_t^{\text{round}}$ happens if only $\tba^\top_t\tbmu_1\leq \tr_t\leq \tba^\top_t\tbmu_2$; ii) if $\bc\geq \tba_t$ does not hold and only $\tbmu_1$ is well-defined in \eqref{eqn:dualsolution}, then $\tx^*_t\neq\tx_t^{\text{round}}$ happens if only $\tba^\top_t\tbmu_1\leq \tr_t$. Therefore, we can again involve the dual variables $\tbmu_1$ and $\tbmu_2$ (if well-defined) in bounding the rounding gap $G_{\bc}(I_t)$ for any $\bc\geq0$, which reduces bounding the rounding gap into bounding the ``dual convergence''. We formalize the above arguments in the following lemma, where the proof is relegated to \Cref{sec:missingpf2}. 
\begin{lemma}\label{lem:firstvariation}
For any $\bc$, it holds that
\[
\mathbb{E}_{I_{t}}\left[G_{\bc}(I_t)\right]\leq\frac{\kappa_2}{s-1}+\kappa_3\cdot \mathbb{E}_{\tba_t}\left[\mathbb{E}_{I_{t+1}}[(\tba^\top_t\tbmu_1-\tba^\top_t\hbmu_1)^2]\right]+3\bar{\alpha}\cdot\mathbb{E}_{\tba}\left[\bI_{\{\bc\geq\tba_t\}}\cdot\mathbb{E}_{I_{t+1}}[(\tba^\top_t\tbmu_2-\tba^\top_t\hbmu_2)^2]\right]
\]
for a constant $\kappa_2=\max\left\{ \frac{6\bar{\alpha}\bar{d}^{3} m^{1/2}\gamma}{\underline{\alpha}\underline{\beta}}, \frac{u_{\max}}{\alpha\cdot p_{\min}} \right\}$ where $\bar{d}=\max_{j\in[n]}\{\|\ba_j\|_2\}$ and $\gamma=\max_{i\in[m], j\in[n]:a_{j,i}>0}\frac{u_j}{a_{j,i}}$, $u_{\max}=\max_j\{u_j\}$ and $p_{\min}=\min_j\{p_j\}$, and a constant $\kappa_3=\max\{1, 3\bar{\alpha}\}$. The variable $\tbmu_1$ and $\tbmu_2$ (if $\bc\geq\tba_t$ and well-defined) are defined in \eqref{eqn:dualsolution}.
\end{lemma}

\subsection{Bound on Dual Convergence and Policy}\label{sec:BoundDual}

In this section, we first bound the ``dual convergence'' and then propose our $\hat{M}-$estimator. By further utilizing the second-order growth condition established in \Cref{lem:secondorder}, we can show the following bound over the term $\mathbb{E}[(\ba_t^\top\tbmu_1-\ba_t^\top\hbmu_1)^2]$. We only state the ``dual convergence'' result for $\hbmu_1$ and $\tbmu_1$. Since the bound is independent of $\bc$, it is clear that the same bound also holds for $\hbmu_2$ and $\tbmu_2$ whenever well-defined (i.e. $\bc\geq\tba_t$).
\begin{lemma}\label{lem:boundIandII}
For any $\bc\geq0$, let $\hbmu_1$ be defined in \eqref{eqn:limitingsolution} and $\tbmu_1$ be defined in \eqref{eqn:dualsolution}. Then, it holds that 
\[
\mathbb{E}[(\ba_t^\top\tbmu_1-\ba_t^\top\hbmu_1)^2]\leq\frac{8\bar{d}^2}{\underline{\alpha}^2\underline{\beta}^2\cdot(s-1)}+\frac{1}{9\bar{\alpha}\bar{d}^2\cdot(s-1)}+ \frac{2}{s-1}
\]
as long as $s-1=T-t\geq t_0$, where $t_0$ is a constant that is determined polynomially by the problem parameters $\underline{\alpha}, \underline{\beta}, \bar{\alpha}, \bar{d}, m$.
\end{lemma}
The key of the proof of \Cref{lem:boundIandII} is to regard $\tbmu_1$ as the solution of the sample average approximation (definition in \eqref{eqn:dualsolution}) of the stochastic programming \eqref{eqn:limitingproblem} with parameter $\bc$, whose optimal solution is $\hbmu_1$ by definition \eqref{eqn:limitingsolution}. The second-order growth condition in \Cref{lem:secondorder} enables us to apply Theorem 2.1 in \citet{shapiro1993asymptotic} showing that the gap between $\tbmu_1$ and $\hbmu_1$ is at the order of $\sqrt{\frac{1}{s-1}}$ with high probability, as $s-1=T-t\rightarrow\infty$. This result is also known as asymptotic normality of sample average solution in the stochastic programming literature. Note that in order to apply Theorem 2.1 in \citet{shapiro1993asymptotic}, there are some additional conditions need to be satisfied. We verified that all these conditions are satisfied by our problem in the proof of \Cref{lem:boundIandII}, which is relegated to \Cref{sec:missingpf2}. However, the high probability result cannot be directly translated into a bound over the $L_2$ norm. In order to obtain the bound over the $L_2$ norm, we further utilize the method developed in \citet{li2021online}. As a result, a combination of the methods from \citet{shapiro1993asymptotic} and \citet{li2021online} enables us to bound the $L_2$ norm at the order of $\frac{1}{s-1}$, which improves the $\frac{\log\log (s-1)}{s-1}$ bound established in Theorem 1 in \citet{li2021online}. 

We now present our $\hat{M}-$estimator to complete our algorithm and provide the final regret bound. From \Cref{lem:dualbound}, we know that $M_{c,\ba_t}(I_{t+1})$ is close to $\ba_t^\top\tbmu_1$ and from the ``dual convergence'' established in \Cref{lem:boundIandII}, we know that $\hbmu_1$ is a good ``approximate'' of $\tbmu_1$. Therefore, we use $\ba_t^\top\hbmu_1$ as our $\hat{M}-$estimator, which is formalized in \Cref{alg:Mestimator}.
\begin{algorithm}[ht!]
\caption{Algorithm achieving $O(\log T)$ Regret}
\label{alg:MLog}
\begin{algorithmic}[1]
\State Input: the remaining inventory $\tbc^{\pi}_t$ and size $\tba_t$.
\State Obtain $\tbmu_1$ by solving
\[
\min_{\bmu\in\Omega} \LF_{\tbc^{\pi}_t,t+1}(\bmu):=\left(\frac{\tbc^{\pi}_t}{T-t}\right)^{\top}\bmu+\mathbb{E}_{(\tr,\tba)\sim F}[\tr-\tba^{\top}\bmu]^+.
\]
\State Output: $\hat{M}_{\tbc^{\pi}_t,\tba_t}=\ba_t^\top\tbmu_1$.
\end{algorithmic}
\end{algorithm}
We provide the following regret bound for the $\hat{M}-$estimator in \Cref{alg:MLog}.
\begin{theorem}\label{thm:LogRegret}
Suppose that the estimator $\hat{M}$ is given in \Cref{alg:MLog} and we denote by $\pi$ the policy given in \Cref{alg:Mestimator}. Then, it holds that
\[
\text{Regret}(\pi)\leq C_1\cdot \log T+ C_2
\]
where $C_1$ and $C_2$ are two constants.
\end{theorem}
\begin{myproof}[Proof of \Cref{thm:LogRegret}]
From \Cref{lem:decompose}, we have
\begin{equation}\label{eqn:100903}
\begin{aligned}
\text{Myopic}_t(\pi,\tbc^{\pi}_t)\leq&2\bar{\alpha}\cdot\mathbb{E}_{\tba_t}[\bI_{\{\tbc^{\pi}_t\geq \tba_t\}}\cdot\Var(M_{\tbc^{\pi}_t,\tba_t}(I_{t+1}))]+\mathbb{E}_{I_t}[G_{\tbc_t^{\pi}}(I_t)]\\
&+2\bar{\alpha}\cdot\mathbb{E}_{\tba_t}\left[\bI_{\{\tbc^{\pi}_t\geq \tba_t\}}\cdot\left(\hat{M}_{\tbc^{\pi}_t,\tba_t}-\mathbb{E}_{I_{t+1}}[M_{\tbc^{\pi}_t,\tba_t}(I_{t+1})]\right)^2\right].
\end{aligned}
\end{equation}
Then, from \Cref{lem:Decomposevariation} and \Cref{lem:firstvariation}, we know that
\begin{equation}\label{eqn:100904}
\begin{aligned}
\text{Myopic}_t(\pi,\tbc^{\pi}_t)\leq& 
\frac{24\bar{\alpha}}{(T-t)\cdot\underline{\alpha}\underline{\beta}}\cdot\bar{d}^{3}\cdot m^{1/2}\cdot\gamma+\frac{\kappa_2}{T-t}+(28\bar{\alpha}+2\bar{\alpha}\kappa_3)\cdot\mathbb{E}_{\tba_t}\left[\mathbb{E}_{I_{t+1}}[(\tba_t^\top\tbmu_1-\tba_t^\top\hbmu_1)^2]\right]\\
&+(28\bar{\alpha}+6\bar{\alpha}^2)\cdot\mathbb{E}_{\tba_t}\left[\bI_{\tbc^{\pi}_t\geq\tba_t}\cdot\mathbb{E}_{I_{t+1}}[(\tba_t^\top\tbmu_2-\tba_t^\top\hbmu_2)^2]\right]\\
&+2\bar{\alpha}\cdot\mathbb{E}_{\tba_t}\left[\bI_{\{\tbc^{\pi}_t\geq \tba_t\}}\cdot\left(\hat{M}_{\tbc^{\pi}_t,\tba_t}-\mathbb{E}_{I_{t+1}}[M_{\tbc^{\pi}_t,\tba_t}(I_{t+1})]\right)^2\right]
\end{aligned}
\end{equation}
where $\tbmu_1, \tbmu_2$ are defined in \eqref{eqn:dualsolution} and $\hbmu_1, \hbmu_2$ are defined in \eqref{eqn:limitingsolution}, with $\bc=\tbc^{\pi}_t$. The constant $\kappa_2=\max\left\{ \frac{6\bar{\alpha}\bar{d}^{3} m^{1/2}\gamma}{\underline{\alpha}\underline{\beta}}, \frac{u_{\max}}{\alpha\cdot p_{\min}} \right\}$ where $\bar{d}=\max_{j\in[n]}\{\|\ba_j\|_2\}$ and $\gamma=\max_{i\in[m], j\in[n]:a_{j,i}>0}\frac{u_j}{a_{j,i}}$, $u_{\max}=\max_j\{u_j\}$ and $p_{\min}=\min_j\{p_j\}$, and the constant $\kappa_3=\max\{1, 3\bar{\alpha}\}$.
Moreover, from \Cref{lem:dualbound} and the definition of $\hat{M}_{\tbc^{\pi}_t,\tba_t}$ in \Cref{alg:MLog}, we know
\[\begin{aligned}
&\mathbb{E}_{\tba_t}\left[\bI_{\{\tbc^{\pi}_t\geq \tba_t\}}\cdot\left(\hat{M}_{\tbc^{\pi}_t,\tba_t}-\mathbb{E}_{I_{t+1}}[M_{\tbc^{\pi}_t,\tba_t}(I_{t+1})]\right)^2\right]\leq \mathbb{E}_{\tba_t}\left[ \bI_{\{\tbc^{\pi}_t\geq \tba_t\}}\cdot\left((\hbmu_1-\mathbb{E}[\tbmu_1])^2+(\hbmu_1-\mathbb{E}[\tbmu_2])^2\right) \right]\\
&\leq \mathbb{E}[(\hbmu_1-\tbmu_1)^2]+2\cdot\mathbb{E}[(\hbmu_1-\hbmu_2)^2]+2\cdot\mathbb{E}[\bI_{\{\tbc^{\pi}_t\geq \tba_t\}}\cdot(\hbmu_2-\tbmu_2)^2].
\end{aligned}\]
We apply \Cref{claim:100801} to bound the term $\mathbb{E}[(\hbmu_1-\hbmu_2)^2]$ and we get
\[
\mathbb{E}[(\hbmu_1-\hbmu_2)^2]\leq \frac{2}{(T-t)\underline{\alpha}\underline{\beta}}\cdot\bar{d}^{3}\cdot m^{1/2}\cdot\gamma.
\]
We apply the ``dual convergence'' established in \Cref{lem:boundIandII} to bound $\mathbb{E}[(\hbmu_1-\tbmu_1)^2]$ and $\mathbb{E}[\bI_{\{\tbc^{\pi}_t\geq \tba_t\}}\cdot(\hbmu_2-\tbmu_2)^2]$ at the order of $O(\frac{1}{T-t})$, as long as $T-t$ is large enough. To be specific, we have
\[
\mathbb{E}[(\ba_t^\top\tbmu_1-\ba_t^\top\hbmu_1)^2]\leq\frac{8\bar{d}^2}{\underline{\alpha}^2\underline{\beta}^2\cdot(T-t)}+\frac{1}{9\bar{\alpha}\bar{d}^2\cdot(T-t)}+ \frac{2}{T-t}
\]
and
\[
\mathbb{E}[(\ba_t^\top\tbmu_2-\ba_t^\top\hbmu_2)^2]\leq\frac{8\bar{d}^2}{\underline{\alpha}^2\underline{\beta}^2\cdot(T-t)}+\frac{1}{9\bar{\alpha}\bar{d}^2\cdot(T-t)}+ \frac{2}{T-t}
\]
as long as $T-t\geq t_0$, where $t_0$ is the constant specified in \Cref{lem:boundIandII} and is determined polynomially by the problem parameters $\underline{\alpha}, \underline{\beta}, \bar{\alpha}, \bar{d}, m$. 
Therefore, we have
\[
\text{Myopic}_t(\pi,\tbc^{\pi}_t)\leq \frac{C_1}{T-t}
\]
as long as $T-t\geq t_0$. Here, the constant $C_1$ equals the following
\begin{equation}\label{eqn:pfC1}
C_1= \frac{(24\bar{\alpha}+4\alpha)\bar{d}^{3} m^{1/2}\gamma}{\underline{\alpha}\underline{\beta}}+\kappa_2+(64\bar{\alpha}+2\bar{\alpha}\kappa_3+6\bar{\alpha}^2)\cdot\left( \frac{8\bar{d}^2}{\underline{\alpha}^2\underline{\beta}^2}+\frac{1}{9\bar{\alpha}\bar{d}^2}+2 \right).
\end{equation}
The constant $\kappa_2=\max\left\{ \frac{6\bar{\alpha}\bar{d}^{3} m^{1/2}\gamma}{\underline{\alpha}\underline{\beta}}, \frac{u_{\max}}{\alpha\cdot p_{\min}} \right\}$ where $\bar{d}=\max_{j\in[n]}\{\|\ba_j\|_2\}$ and $\gamma=\max_{i\in[m], j\in[n]:a_{j,i}>0}\frac{u_j}{a_{j,i}}$, $u_{\max}=\max_j\{u_j\}$ and $p_{\min}=\min_j\{p_j\}$, and the constant $\kappa_3=\max\{1, 3\bar{\alpha}\}$. As a result, the final regret bound can be bounded as
\[
\text{Regret}(\pi) \leq C_1\cdot \log T+C_2
\]
where the constant $C_1$ is defined in \eqref{eqn:pfC1} and the constant $C_2$ equals
\begin{equation}\label{eqn:pfC2}
C_2=t_0=O\left( \left(mM\cdot\frac{\bar{\alpha}^3\bar{d}^7}{\underline{\alpha}^2\underline{\beta}^2}\right)^{\frac{25}{2}}\cdot\frac{1}{q} \right),
\end{equation}
which is determined polynomially by the problem parameters $\underline{\alpha}, \underline{\beta}, \bar{\alpha}, \bar{d}, m$.
\end{myproof}

\noindent\textbf{Remarks.} Note that the constant terms $C_1$ and $C_2$ in the regret bound of \Cref{thm:LogRegret} depend polynomially on the problem parameters specified in \Cref{assump:finitesize} and \Cref{assump:contiknown2}. In contrast, the constant terms in the regret bound of \Cref{thm:Regretthm} depend exponentially on the number of customer types, which is $n$, based on the choice of the parameter $\kappa_1$ in \eqref{def:kappa}, and depend polynomially on all the other parameters. Our results show that, for the general setting where only \Cref{assump:finitesize} is satisfied, we can apply \Cref{alg:Mestimator} with the estimator $\hat{M}$ given in \Cref{alg:M1} to achieve a regret bound $O(\log^2 T)$, where the constant term depends exponentially on the number of customer types $n$ and polynomially on all the other parameters. When the additional \Cref{assump:contiknown2} is also satisfied, we can apply \Cref{alg:Mestimator} with the estimator $\hat{M}$ given in \Cref{alg:MLog} and obtain an improved $O(\log T)$, where the constant term depends polynomially on all the problem parameters including $n$. The last result matches the logarithmic regret established in a series of papers (e.g. \cite{li2021online, bray2022logarithmic}), but without the non-degeneracy assumption.

\ACKNOWLEDGMENT{
The authors thank Rob Bray for motivations for removing the degeneracy assumption, and He Wang for bringing up the connection with price-based NRM. The authors also would like to thank Sid Banerjee for bringing references on regret lower bound to our attention. The authors thank the editors and reviewers of the journal Operations Research for the valuable comments that greatly improve our paper.
}

\bibliographystyle{abbrvnat}
\bibliography{bibliography}

\clearpage

%
%
%

\begin{APPENDICES}
\crefalias{section}{appendix}

\section{Numerical Performances of Our Algorithms}
In this section, we conduct numerical experiments to test the empirical performances of our algorithms. To be specific, we consider an NRM problem with $m$ resources and $n$ customer types. For each customer type $j\in[n]$, we randomly generate its size $\ba_j$ by drawing the value of $a_{j,i}$ uniformly from the interval $[0,1]$ for each $i$. We let the reward for type $j$ customer follow a uniform distribution $[l_j, u_j]$, with $l_j$ uniformly drawn from $[0,1]$ and $u_j = l_j + \epsilon_j$ with $\epsilon_j$ being uniformly drawn the interval $[0,1]$, for each $j$. We set the initial capacity $C_i=\alpha_i\cdot T$ for each resource $i\in[m]$, with $\alpha_i$ being a fixed parameter, uniformly drawn from $[0,1]$. We test the performances of our \Cref{alg:M1}, \Cref{alg:MLog}, and compare their performances with previous algorithms that have been developed in the literature. Note that all our algorithms resolve the ex-ante relaxation $\bVF$ (though our benchmark is the semi-fluid relaxation). We compare the algorithm that does not resolve the ex-ante relaxation. For example, the classic fixed bid price control heuristics (FBP) proposed in \cite{talluri1998analysis}, which solves the ex-ante relaxation once to obtain the optimal dual variable $\mu^*$ and then accepts customer $t$ if and only if there are enough remaining capacities and $r_t\geq \ba_t^\top\mu^*$. We also compare against the dual-based policy that has been developed in a stream of literature under various setting (e.g. \cite{balseiro2022best, li2020simple, jiang2020online}), which uses an online learning algorithm to update the dual variable $\mu_t$ at period $t$ and accepts customer $t$ if and only if there are enough remaining capacities and $r_t\geq \ba_t^\top\mu_t$.  

We refer to $\ALG_2$ as the total expected reward collected by \Cref{alg:M1} and refer to $\ALG_3$ as the total expected reward collected by \Cref{alg:MLog}. We then denote by $\ALG_{\mathsf{FB}}$ as the total expected reward collected by the classic fixed bid price control heuristics, and denote by $\ALG_{\mathsf{DU}}$ as the total expected reward collected by the dual update policy. For each instance, we repeat for $K=100$ times and use their average to approximate the expected reward of a policy. In \Cref{numericalfigure1} (a), we show how the expected reward of each policy grows with the horizon $T$, for a fixed problem instance. As we can see, both \Cref{alg:M1} and \Cref{alg:MLog} performs better than the FBP $\ALG_{\mathsf{FB}}$ and the dual update policy $\ALG_{\mathsf{DU}}$, especially when $T$ becomes larger and larger. We also test how the performances of the policies depend on other problem parameters, for example the number of resources $M$ and the number of customer types $N$. To be specific, we fix $T=100$ and plot the ratios of $\ALG_{\mathsf{DU}}/\ALG_{\mathsf{FB}}$, $\ALG_2/\ALG_{\mathsf{FB}}$, and $\ALG_3/\ALG_{\mathsf{FB}}$. For the dependency over the number of resources $M$, as shown in \Cref{numericalfigure1} (b), the three policies, dual update policy, \Cref{alg:M1}, and \Cref{alg:MLog}, perform similar to each other, though all perform much better than the fixed bid price policy. For the dependency over the number of customer types $N$, as shown in \Cref{numericalfigure1} (c), \Cref{alg:M1} and \Cref{alg:MLog} perform better than the dual update policy when $N$ becomes larger and larger. Therefore, we conclude that in all instances, dual update policy, \Cref{alg:M1} and \Cref{alg:MLog} perform better than the fixed bid price policy. When the number of periods $T$ becomes larger or the number of customer types $N$ becomes larger, \Cref{alg:M1} and \Cref{alg:MLog} perform better than the dual update policy. 

\begin{figure*}[ht!]
    \centering
    \begin{subfigure}[h]{0.5\textwidth}
        \centering
        \includegraphics[width=0.9\textwidth]{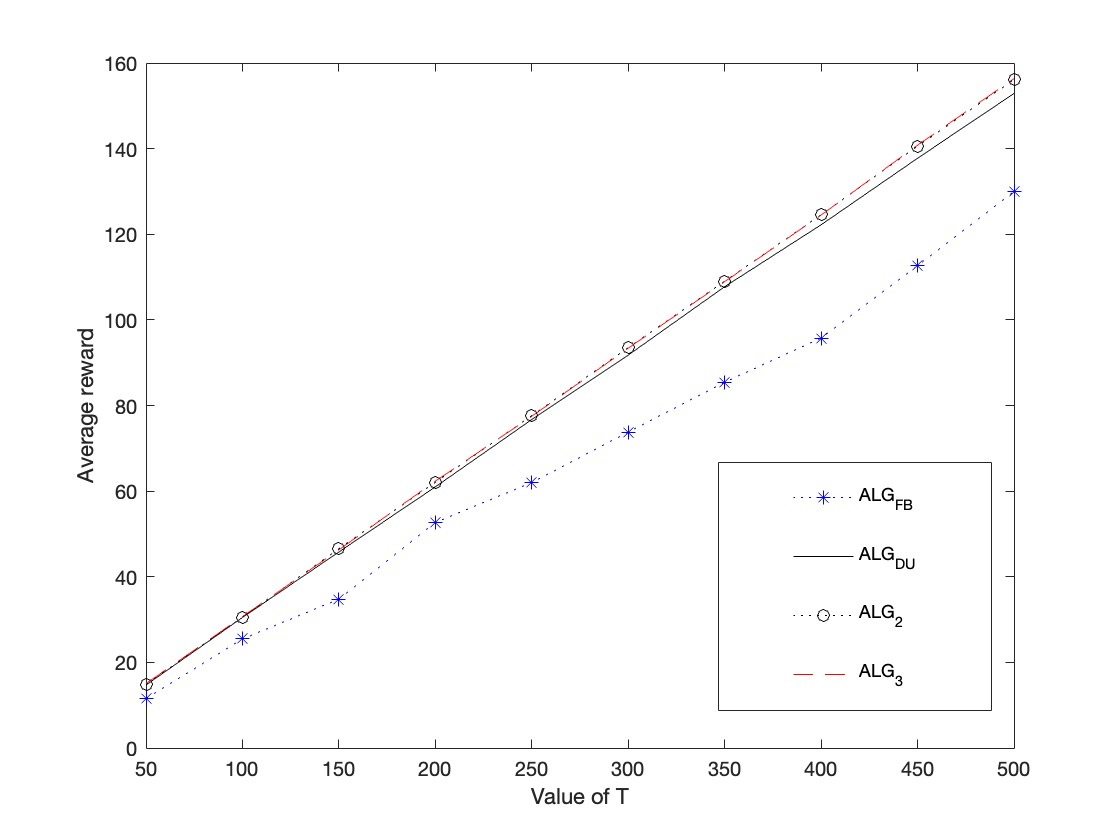}
        \caption{The dependency of the policy rewards on $T$. X-axis denotes the value of $T$ while the y-axis denotes the average reward of each policy.}
    \end{subfigure}%
    ~
    \begin{subfigure}[h]{0.5\textwidth}
        \centering
        \includegraphics[width=0.9\textwidth]{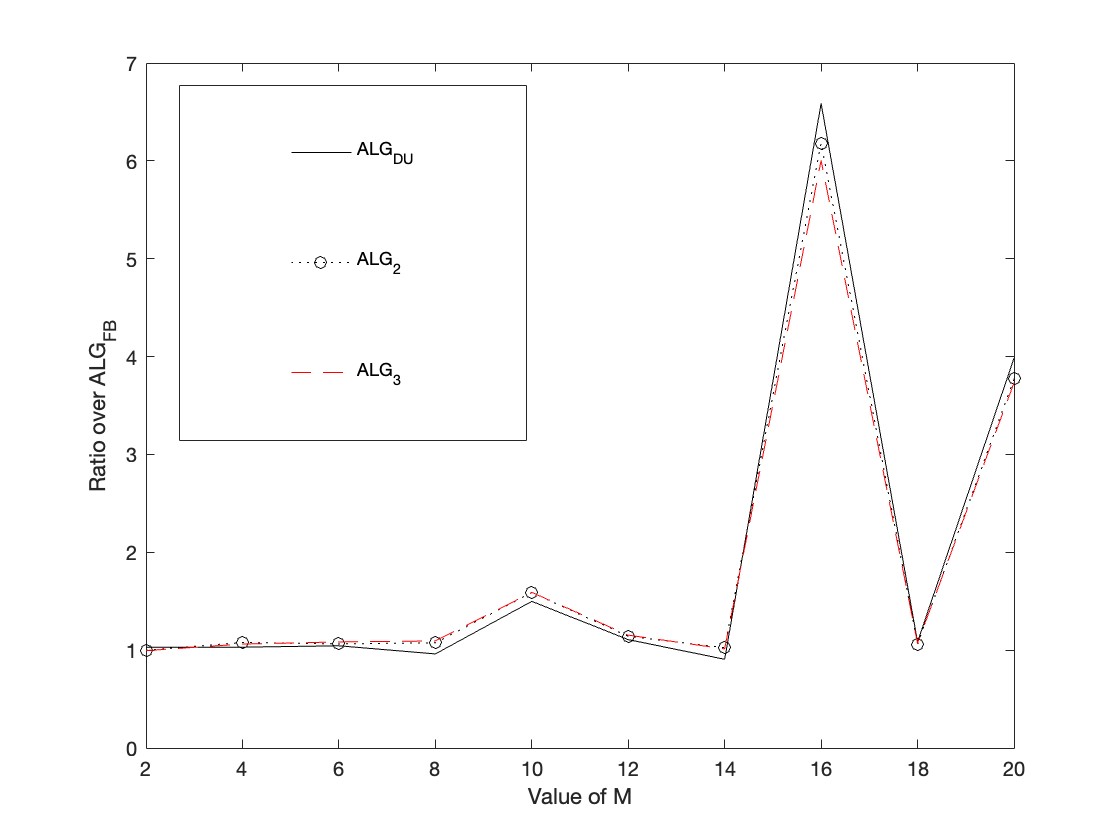}
        \caption{The dependency of the policy rewards on the number of resources $M$. X-axis denotes the value of $M$ while the y-axis denotes the ratio of the average reward of each policy over the fixed bid price policy.}
    \end{subfigure}
    \begin{subfigure}[h]{0.5\textwidth}
        \centering
        \includegraphics[width=0.9\textwidth]{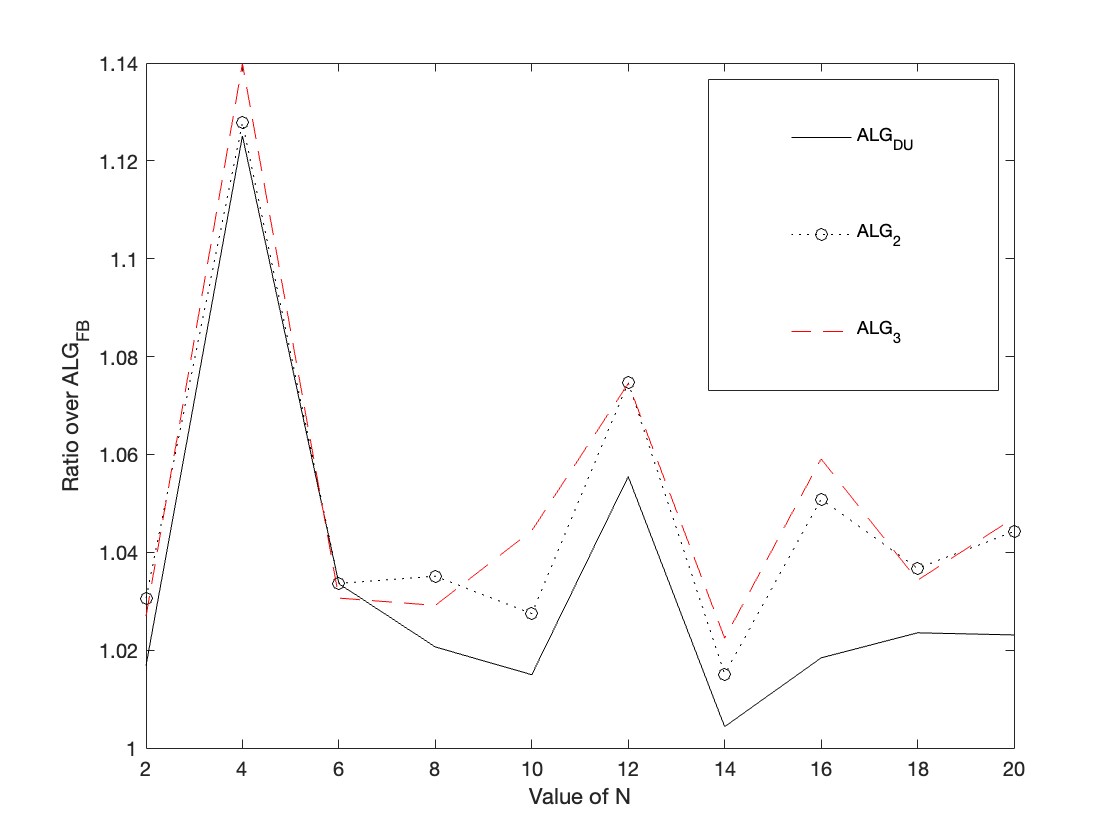}
        \caption{The dependency of the policy rewards on the number of customer types $N$. X-axis denotes the value of $N$ while the y-axis denotes the ratio of the average reward of each policy over the fixed bid price policy.}
    \end{subfigure}
    \caption{The comparison between the four policies, fixed bid price policy, dual update policy, \Cref{alg:M1}, and \Cref{alg:MLog}. }\label{numericalfigure1}
\end{figure*}

\section{Detailed Comparison to Assumptions in Existing Literature}\label{sec:conclusion}
We conclude by making comparisons between our assumptions and the assumptions made in the existing literature, which can be summarized into the following two conditions:\\
(i). The non-degeneracy assumption over the ex-ante relaxation $\bVF$, which requires the optimal solution to $\bVF$ to be unique and strict complementary slackness condition being satisfied.\\
(ii). The second-order growth condition over the dual function $\LF$.\\
We first summarize the assumptions made in \cite{li2021online} in the language of our paper as follows:
\begin{assumption}[Assumptions made in \cite{li2021online}]\label{assumptionLi}
The following conditions have to be satisfied:\\
(i). $\tr_t$ and $\|\tba_t\|_2$ are always bounded for each $t\in[T]$.\\
(ii). $\mathbf{C}$ scales linearly in $T$ and $C_{i}/T\in(\underline{d}, \bar{d})$ with $\bar{d}\geq\underline{d}>0$, for each $i\in[m]$. \\
(iii). The matrix $\mathbb{E}_{\tba}[\tba\tba^\top]$ is positive definite.\\
(iv). There exists a set $\Omega_{\bmu}$ containing all possible optimal dual variable to the offline optimum $\bVL$ \eqref{lp:relaxoffline} such that for any $\bmu\in\Omega_{\bmu}$ and any $\bc\in\Omega_{\bc}=[\underline{d}\cdot T, \bar{d}\cdot T]^m$, it holds that
\[
\underline{\alpha}\cdot |\ba^\top\bmu-\ba^\top\bmu^*(\bc)|\leq |F(\ba^\top\bmu|\ba)-F(\ba^\top\bmu^*(\bc)|\ba)|\leq\bar{\alpha}\cdot |\ba^\top\bmu-\ba^\top\bmu^*(\bc)|
\]
for any $\ba\in\mathcal{A}$,
where $\bmu^*(\bc)$ is the optimal dual solution to the ex-ante relaxation $\bVF_{1,\bc}$ \eqref{lp:exante1}.\\
(v). For any $\bc\in\Omega$, the optimal dual solution $\bmu^*(\bc)$ to the ex-ante relaxation $\bVF_{1,\bc}$ \eqref{lp:exante1} satisfies the strict complementary slackness condition.
\end{assumption}
It is shown in Proposition 2 of \cite{li2021online} that condition (iv) in \Cref{assumptionLi} implies the second-order growth condition of the dual function, while condition (iii), (iv) and (v) all together imply the non-degeneracy assumption. 
We summarize the assumptions made in \cite{bray2022logarithmic} as follows:
\begin{assumption}[Assumptions made in \cite{bray2022logarithmic}]\label{assumptionBray}
The following conditions have to be satisfied:\\
(i). We have $f(r|\ba)\leq\bar{\alpha}$ for any $\ba\in\mathcal{A}$ and any $r$ in the support.\\
(ii). For any $\ba\in\mathcal{A}$, $\mathbb{E}_{(\tr,\tba)\sim F}[\tr|\tba=\ba]\leq \beta$ for a constant $\beta>0$.\\
(iii). $\|\ba\|_2\leq\bar{d}$ for any $\ba\in\mathcal{A}$.\\
(iv). The optimal dual solution to the Lagrangian problem $\min_{\bmu\geq0}\LF_{\bm{c},1}$ of the ex-ante relaxation $\bVF_{1,\bc}$ \eqref{lp:exante1} is unique and the strictly complementary slackness condition is satisfied, when $\bc$ belongs to a neighborhood of $\bm{C}$, which scales linearly in $T$. \\
(v). The Hessian matrix of $\LF_{\bm{C},1}(\bmu^*)$ over $\bmu^*$, where $\bmu^*=\text{argmin}_{\bmu\geq0}\LF_{\bm{C},1}$ is full rank (equivalently, positive definite).\\
(vi). The Hessian matrix of $\LF_{\bm{C},1}(\bmu)$ over $\bmu$ is Lipschitz continuous when $\bmu$ belongs to a neighborhood of $\bmu^*=\text{argmin}_{\bmu\geq0}\LF_{\bm{C},1}$.
\end{assumption}
Note that condition (v) and (vi) in \Cref{assumptionBray} together imply the second-order growth condition over the dual function, while the non-degeneracy assumption is stated in condition (iv) in \Cref{assumptionBray}.
In particular, \cite{balseiro2021survey} has summarized the assumptions in to the following two conditions:
\begin{assumption}[Assumption 2 in \cite{balseiro2021survey}]\label{assumpBalseiro}
The following conditions have to be satisfied:\\
(i). The binding constraints for $\bVF_{1,\bc}$ remains the same as the binding constraints for $\bVF_{1,\bm{C}}$, as long as $\bc$ belongs to a neighborhood of the initial capacity $\bm{C}$, where we denote by $\mathcal{J}$ the set of resource constraints that are binding.\\
(ii). There exists a constant $\kappa>0$ such that
\[
\bVF_{1,\bc}-\bVF_{1,\bm{C}}\geq \left(\nabla_{\bm{C}}\bVF_{1,\bm{C}}\right)^\top \left(\frac{\bm{c}}{T}-\frac{\bm{C}}{T}\right)-\kappa\cdot \left(\frac{\bm{c}_{\mathcal{J}}}{T}-\frac{\bm{C}_{\mathcal{J}}}{T}\right)^2
\]
for all $\bc$ belonging to a neighborhood of the initial capacity $\bm{C}$,
where $\bm{c}_{\mathcal{J}}$ denotes $\mathcal{J}$ part of the vector $\bc$ and $\bm{C}_{\mathcal{J}}$ denotes $\mathcal{J}$ part of the vector $\bm{C}$.  
\end{assumption}
It has been shown that a sufficient condition to guarantee condition (i) in \Cref{assumpBalseiro} is that strict complementary slackness condition is satisfied by $\bVF_{1,\bm{C}}$ (SC 8 in \citet{balseiro2021survey}), and a sufficient condition to guarantee condition (ii) in \Cref{assumpBalseiro} is that the second-order growth condition is satisfied by dual function $\LF$ (SC 7 in \citet{balseiro2021survey}).  

In \Cref{sec:Logsection}, we need the second-order growth condition, but without the non-degeneracy condition, and we derive a $O(\log T)$ regret bound following our myopic regret framework. In contrast, in \Cref{sec:knowndistribution}, we get rid of both the second-order growth condition and the non-degeneracy condition and consider our problem under the most general setting. Our main result is a $O(\log^2 T)$ regret bound. Both regret bounds are new in the literature.

\textbf{Perturbation attempts to overcome degeneracy.} Note that in the traditional LP literature, one prevalent way to overcome degeneracy is to perturb the right-hand side of the constraints of $\bVF_{1,\bm{C}}$ by $\delta\cdot T$ with $\delta$ being a constant satisfying some conditions \citep{megiddo1989varepsilon}. However, such a perturbation way will lead to a $\Omega(\sqrt{T})$ regret for the NRM problem. To be specific, denote by $\bVF_{1,\bm{C}}(\delta)$ the fluid relaxation after perturbation. Then, it holds that $|\bVF_{1,\bm{C}}(\delta)-\bVF_{1,\bm{C}}|=O(\delta\cdot T)$. It has been shown that the regret of their policies with respect to $\bVF_{1,\bm{C}}(\delta)$ scales with $\Omega(1/\delta)$ (e.g. see Theorem 1 of \cite{balseiro2021survey}). Therefore, the regret of the (certainty-equivalent) policies developed in the previous literature \citep{li2021online, bray2022logarithmic, balseiro2021survey} with respect to $\bVF_{1,\bm{C}}$ scales with $O( \delta\cdot T+ 1/\delta)$. From the above discussion, we conclude that the perturbation attempts to overcome degeneracy can only lead to a $O(\sqrt{T})$ regret bound for the NRM problem.

\section{Capturing the Price-based NRM Problem} \label{sec:pricebased}

In this \namecref{sec:pricebased} we explain how our regret results for NRM with accept/reject decisions (often called the \textit{quantity-based} NRM problem) extend to the \textit{price-based} NRM problem, and why our model (with a finite number of possible demand vectors but an infinite number of possible reward values) is quite natural for the pricing problem.

In the prototypical price-based NRM problem, a firm starts with initial resource vector $\bm{C}\in\bR^m$.
The firm is selling a finite set of products, indexed $j=1,\ldots,n$, with each product $j$ needing to consume resource vector $\ba_j\in\bR^m$ in order to be sold.  At each time $t=1,\ldots,T$, the firm must first post a price $P^t_j$ for each product $j$ ($P^t_j$ can be $\infty$ if there are insufficient resources for product $j$ at time $t$).  Afterward, customer $t$ arrives, wanting a random product $\tj_t$ drawn independently according to a known probability vector $(p_1,\ldots,p_n)$.  Conditional on the customer wanting product $j$, their valuation $\tv_t$ is drawn independently from a known, $j$-specific distribution $H_j$ that is continuous over an interval $[l_j,u_j]$ with PDF $h_j$ satisfying $h_j(v)>0$.
Customer $t$ makes a purchase if and only if $\tv_t\ge P^t_{\tj_t}$, in which case the firm collects revenue $P^t_{\tj_t}$ and consumes resources $\ba_{\tj_t}$.
The firm's objective is to maximize the expected total revenue collected.

\begin{definition}
For all $j=1,\ldots,n$, the \textit{virtual valuation} corresponding to a valuation $v$ drawn from $H_j$ is $\phi_j(v):=v-\frac{1-H_j(v)}{h_j(v)}$.
The \textit{expected virtual surplus} is defined to be
\begin{align*}
\bE\left[\max_{S\subseteq[T]:\sum_{t\in S}\ba_{\tj_t}\le\bm{C}}\phi_{\tj_t}(\tv_t)\right].
\end{align*}
For all $j$, let $F_j$ denote the CDF of the random variable $\phi_j(\tv)$ when $\tv$ is drawn from $H_j$.
\end{definition}

The virtual valuation $\phi_j(v)$ is well-defined for all $v\in[l_j,u_j]$, by the assumption that $h_j(v)>0$.
We now make the standard \textit{regularity} assumption that the virtual valuation function $\phi_j$ is monotonic, along with an assumption on $F_j$ having a lower-bounded PDF which will allow us to apply our results by considering accept/reject decisions on rewards drawn from the distribution $F_1,\ldots,F_n$.

\begin{assumption} \label{ass:regular}
For all $j=1,\ldots,n$, valuation distribution $H_j$ is such that $\phi_j$ is a non-decreasing function over $[l_j,u_j]$.
Moreover, the virtual valuation distribution $F_j$ has a PDF $f_j$ satisfying $f_j(r)\ge\alpha$ for all values $r\ge0$ lying in the support of $f_j$, where $\alpha>0$ is a constant.
\end{assumption}

We note that even under the first part of \Cref{ass:regular} (regularity), virtual valuations can be negative.  The second part of \Cref{ass:regular} imposes that the non-negative part of the support of virtual valuations for any $j=1,\ldots,n$ has a PDF that is lower-bounded by $\alpha$.

\begin{theorem}[\citet{myerson1981optimal,chawla2010multi}] \label{thm:reduction}
Under \Cref{ass:regular}, the expected revenue of the optimal Bayesian incentive-compatible and individually-rational mechanism, which is an upper bound on the revenue of any online pricing policy, is equal to the expected virtual surplus.
Moreover, consider the quantity-based NRM problem on rewards drawn from the virtual valuation distributions $F_1,\ldots,F_n$.
Any online accept/reject policy for this problem can be converted into an online pricing policy for the price-based NRM problem, such that
the expected revenue earned in the price-based NRM problem
equals
the expected reward collected in the quantity-based NRM problem.
\end{theorem}

\Cref{thm:reduction} shows that regret guarantees for quantity-based NRM imply the same guarantees for price-based NRM.  We note that ``regret'' in the price-based NRM setting is defined against the optimal mechanism benchmark, which is an upper bound on the revenue of any online pricing policy because these are special cases of sequential mechanisms.

When applying the accept/reject policy from the virtual valuation space on the original price-based NRM problem, for any $j=1,\ldots,n$, one takes the minimum acceptable virtual valuation threshold $Q_j$ and maps it to a price $P_j\in[l_j,u_j]$ with the same probability of sale (i.e.\ $1-F_j(Q_j)=1-H_j(P_j)$) and the same immediate expected reward/revenue (i.e.\ $\int_{F_j(Q_j)}^1F^{-1}_j(q)dq=P_j(1-H_j(P_j))$).
Since virtual valuations can be negative, the accept/reject problem can face negative rewards, something not captured by our initial quantity-based NRM model.
However, this is easily assuaged by splitting each ``type'' $j$ into two, one of which has a reward that is deterministically 0 (such rewards would never be accepted, so it is without loss of generality to convert negative virtual valuations into 0 rewards).

We now show that \Cref{ass:regular} is satisfied for many valuation distributions commonly used in the pricing literature, leading to the following corollary.

\begin{corollary} \label{cor:priceConvert}
If all valuation distributions satisfy \Cref{ass:regular}, then logarithmic-level regret can be achieved for the price-based NRM problem.
In particular, the following classes of valuation distributions satisfy \Cref{ass:regular}:
\begin{enumerate}
\item Uniform valuations, i.e.\ $h_j(v)=\frac{1}{u_j-l_j}$ for all $v\in[l_j,u_j]$;
\item Truncated normal valuations, i.e.\ $h_j(v)=\alpha_j\cdot\exp(-v^2/\sigma_j)$ for all $v\in[l_j, u_j]$;
\item Truncated exponential valuations, i.e.\ $h_j(v)=\alpha_j\cdot\exp(-\lambda_j\cdot v)$ for all $v\in[l_j,u_j]$;
\end{enumerate}
\end{corollary}

\begin{myproof}
\begin{enumerate}
\item Virtual valuations, defined by $\phi_j(v)=2v-u_j$ for all $v\in[l_j,u_j]$, are uniformly distributed over $[2l_j-u_j,u_j]$, i.e.\ $f_j(r)=\frac{1}{2(u_j-l_j)}$ for all $r\in[2l_j-u_j,u_j]$;
\item Virtual valuations of the truncated normal valuations are known to be monotonic \citep{hartline2013mechanism}. We only prove the bounds for the pdf $f_j(r)$. For a $v_1, v_2\in[l_j, u_j]$, we denote by
\[
r_1=\phi_j(v_1)=v_1-\frac{1-H_j(v_1)}{h_j(v_1)}\text{~and~}r_2=\phi_j(v_2)=v_2-\frac{1-H_j(v_2)}{h_j(v_2)}.
\]
Then, it holds that
\begin{equation}\label{eqn:112901}
\begin{aligned}
f_j(r_1)&=\lim_{r_2\rightarrow r_1}\frac{F_j(r_2)-F_j(r_1)}{r_2-r_1}=\lim_{v_2\rightarrow v_1}\frac{H_j(v_2)-H_j(v_1)}{\phi'_j(v_1)\cdot(v_2-v_1)}=\frac{h_j(v_1)}{\phi'_j(v_1)}\\
&=\frac{h^3_j(v_1)}{2h^2_j(v_1)+h'_j(v_1)-H_j(v_1)\cdot h'_j(v_1)}.
\end{aligned}
\end{equation}
By noting that $h_j'(v_1)\leq0$, it is easy to see that
\[
\frac{h_j(v_1)}{2}\leq f_j(r_1).
\]
Clearly, $\frac{h_j(v_1)}{2}$ is lower bounded by a positive constant on the interval $v_1\in[l_j, u_j]$, which implies that the pdf of the virtual valuation has a positive lower bound.
\item Virtual valuations of the truncated exponential valuations are known to be monotonic \citep{hartline2013mechanism}. We only prove the bounds for the pdf $f_j(r)$. Following \eqref{eqn:112901} and $h'_j(v_1)\leq0$, we have
\[
\frac{h_j(v_1)}{2}\leq f_j(r_1).
\]
Therefore, $f_j(r_1)$ is lower bounded by a positive constant for any $r_1$ by noting that $h_j(v_1)$ is lower bounded by a positive constant for any $v_1\in[l_j, u_j]$.
\end{enumerate}
\end{myproof}

\section{Useful Known Results}\label{sec:usefulknown}
We proceed now to establish Lipschitz continuity of solutions of linear systems with respect to right-hand side perturbations.
\begin{lemma}[Theorem 2.2 of \cite{mangasarian1987lipschitz}]\label{lem:Lipschitz}
Consider two linear systems
\begin{equation}\label{system:1}
\hat{A}\bm{x}\leq\bm{b}^1
\end{equation}
and
\begin{equation}\label{system:2}
\hat{A}\bm{x}\leq\bm{b}^2.
\end{equation}
For any solution $\bm{x}^1$ that satisfies linear system \eqref{system:1}, there exists a solution $\bm{x}^2$ that satisfies linear system \eqref{system:2} such that
\[
\|\bm{x}^1-\bm{x}^2\|_{\infty}\leq\mu\cdot\|\bm{b}^1-\bm{b}^2\|_{\infty}
\]
where
\[
\mu=\sup_{\bm{v}}\left\{\left\|
\bm{v}
\right\|_{1} \left|\begin{aligned}
&\|\bm{v}^\top \hat{A}\|_1=1,~\bm{v}\geq0\\
&\text{Rows of }\hat{A}
\text{ corresponding non-zero elements}\\
&\text{of }\bm{v}
\text{ are linear independent.}
\end{aligned}
\right.\right\}.
\]
\end{lemma}
We state the Hadamard's inequality on matrix determinant in the following lemma.
\begin{lemma}[\cite{hadamard1893resolution}]\label{lem:hadamard}
For any matrix $A$, let $K$ denotes the number of columns in matrix $A$ and let $\{\bm{a}_k\}_{k=1}^K$ denotes all the columns of matrix $A$. Then, the determinant of matrix $A$, denoted by $\text{det}(A)$, satisfies the following inequality
\[
\text{det}(A)\leq\prod_{k=1}^K\|\bm{a}_k\|_2.
\]
\end{lemma}
We state the well-known Bernstein's inequality in the following lemma.
\begin{lemma}[Bernstein's Inequality]\label{lem:Berstein}
Let $X_1,\dots,X_K$ be independent zero-mean random variables. Suppose that $|X_k|\leq M$ almost surely for all $k\in[K]$. Then, for all positive $\epsilon>0$, it holds that
\[
P\left(\frac{1}{K}\cdot\sum_{i=1}^K X_k\geq\epsilon\right)\leq\exp\left(-\frac{\frac{1}{2}\cdot K^2\epsilon^2}{\sum_{k=1}^K\text{Var}(X_k)+\frac{1}{3}\cdot MK\epsilon}\right)
\]
\end{lemma}
We also state the well-known Hoeffding's inequality in the following lemma.
\begin{lemma}[Hoeffding's Inequality]\label{lem:Hoeffding}
Let $X_1,\dots,X_K$ be independent random variables such that $a_k\leq X_k\leq b_k$ almost surely, for each $k\in[K]$. Denote by $S_K=\frac{1}{K}\cdot\sum_{k=1}^K X_k$. Then, for any $\epsilon>0$, it holds that
\[
P\left( |S_k-\mathbb{E}[S_k]|\geq \epsilon \right)\leq 2\exp(-\frac{2K^2\epsilon^2}{\sum_{k=1}^K(b_k-a_k)^2}).
\]
\end{lemma}
We then state the results from \cite{huber1967under} under our notations.
\begin{assumption}\label{huberassumption}
Suppose the following conditions hold:\\
(N-1). For each fixed $\bmu$, the function $\psi((r,\ba), \bmu)$ is separable.\\
(N-2). Denote $\lambda(\bmu)=\mathbb{E}_{(r,\ba)}[\psi((r,\ba),\bmu)]$, then we have $\lambda(\bmu^*)=0$.\\
Denote $$u((r,\ba),\bmu,d)=\sup_{\|\bmu'-\bmu\|_2\leq d}|\psi((r,\ba),\bmu')-\psi((r,\ba),\bmu)|.$$
(N-3). There are strictly positive numbers $a, b, c_1, c_2, d_0$ such that\\
(i). $|\lambda(\bmu)|\geq a\cdot\|\bmu-\bmu^*\|$ for $\|\bmu-\bmu^*\|\leq d_0$.\\
(ii). $\mathbb{E}_{(\tr,\tba)}[u((\tr,\tba),\bmu,d)]\leq b\cdot d$ for $\|\bmu-\bmu^*\|+d\leq d_0$.\\
(iii). $\mathbb{E}_{(\tr,\tba)}[u((\tr,\tba),\bmu,d)^2]\leq c_1\cdot d$ for $\|\bmu-\bmu^*\|+d\leq d_0$.\\
(iv). $u((r,\ba),\bmu,d)\leq c_2$ for any $(r,\ba)$.
\end{assumption}
Denote by
\[
Z_n(\bmu',\bmu'')=\frac{\sum_{j=1}^n[\psi((r_j,\ba_j), \bmu')-\psi((r_j,\ba_j), \bmu'')+\lambda(\bmu')-\lambda(\bmu'')]  }{\sqrt{n}+n|\lambda(\bmu')|}.
\]
Then we have the following result from \citet{huber1967under}. Note that the original statement in \citet{huber1967under} only concerns the convergence of $\sup_{\|\bmu-\bmu^*\|\leq d_0} Z_n(\bmu,\bmu^*)$ to $0$ as $n\rightarrow\infty$. We now specify the constant terms in their bound and characterize the convergence rate, which will be helpful in our other proofs. The proof simply follows the proof in \citet{huber1967under}, except that we make specific the constant terms, and we include here for completeness.
\begin{lemma}[Lemma 3 in \cite{huber1967under}]\label{lem:huber}
The conditions in \Cref{huberassumption} imply that 
\[
\sup_{\|\bmu-\bmu^*\|\leq d_0} Z_n(\bmu,\bmu^*)\rightarrow0
\]
in probability as $n\rightarrow\infty$. Moreover, for any $\epsilon>0$, it holds that
\[
P\left( \sup_{\|\bmu-\bmu^*\|\leq d_0} Z_n(\bmu,\bmu^*)\geq2\epsilon \right)\leq c_1\epsilon^{-2} n^{-\gamma'}+2\exp\left(-\frac{\min\{b^2q, \eps^2a^2\}\cdot(1-q)^{2}n^{1-\gamma'}}{c_1(3-q) +2c_2(b+\eps a)(1-q)/3 }\right)\cdot \left(\frac{\gamma'\cdot(\log n+\log d_0)}{|\log(1-q)|}+1\right)\cdot (2M)^m
\]
as long as $n\geq n_0$, where $n_0$ satisfying $n_0^{\gamma'-\frac{1}{2}}=\frac{2b}{\epsilon}$ and $\gamma'\in(\frac{1}{2},1)$ is an arbitrary number. Moreover, we set $M\geq(3b)/(\epsilon a)$ and $q=1/M$.
\end{lemma}
\begin{myproof}[Proof of \Cref{lem:huber}]
For the sake of simplicity, and without loss of generality, we choose the coordinate system such that $\bmu^*=0$. We also use $\bx_j$ to denote $(r_j,\ba_j)$ for each $j\in[n]$. The idea of the proof is to divide the cube $\|\bmu\|\leq d_0$ into a slowly increasing number of smaller cubes and to bound $Z_n(\bmu,0)$ in probability on each of those smaller cubes.

Put $q=1/M$, where $M\geq2$ is an integer to be chosen later, and consider the concentric cubes
\[
C_k=\{\bmu: \|\bmu\|\leq(1-q)^k\cdot d_0\},~~k=0,1,\dots,k_0.
\]
Subdivide the difference $C_{k-1}\backslash C_k$ into smaller cubes with edges of length $2d=(1-q)^{k-1}q\cdot d_0$ such that the coordinates of their centers $\bm{\xi}$ are odd multiples of $d$, and
\[
|\bm{\xi}|=(1-q)^{k-1}(1-\frac{q}{2})\cdot d_0.
\]
For each value of $k$ there are less than $(2M)^m$ such smalle cubes, so there are $N<k_0\cdot(2M)^m$ cubes contained in $C_0\backslash C_{k_0}$; number them $C_{(1)},\dots, C_{(N)}$.

Now let $\epsilon>0$ be given. We shall show that for a proper choice of $M$ and of $k_0=k_0(n)$, the right-hand side of 
\begin{equation}\label{eqn:lem1201}
    P\left( \sup_{\bmu\in C_0}Z_n(\bmu,0)\geq2\epsilon \right)\leq P\left( \sup_{\bmu\in C_{k_0}}Z_n(\bmu,0)\geq2\epsilon \right)+\sum_{l=1}^N P\left( \sup_{\mu\in C_{(l)}}Z_n(\mu,0)\geq2\epsilon \right)
\end{equation}
tends to $0$ with increasing $n$, which establishes the final result.

Actually, we shall choose 
\begin{equation}\label{eqn:lem1202}
    M\geq(3b)/(\epsilon a), \text{~which~implies~}q\leq(\epsilon a)/(3b),
\end{equation}
and $k_0=k_0(n)$ is defined by
\begin{equation}\label{eqn:lem1203}
    (1-q)^{k_0}\cdot d_0\leq n^{-\gamma'}<(1-q)^{k_0-1}\cdot d_0
\end{equation}
where $\frac{1}{2}<\gamma'<1$ is an arbitrary fixed number. Thus, we have
\begin{equation}\label{eqn:lem1204}
    k_0(n)-1<\frac{\gamma'\cdot(\log n+\log d_0)}{|\log(1-q)|}\leq k_0(n),
\end{equation}
hence
\begin{equation}\label{eqn:lem1205}
    N\leq (\frac{\gamma'\cdot(\log n+\log d_0)}{|\log(1-q)|}+1)\cdot (2M)^m.
\end{equation}

Now take any of the cubes $C_{(l)}$, with center $\bm{\xi}$ and edges of length $2d$ according to $2d=(1-q)^{k-1}q\cdot d_0$ and $|\bm{\xi}|=(1-q)^{k-1}(1-\frac{q}{2})\cdot d_0$. For $\bmu\in C_{(l)}$, we have then by (N-3),
\begin{equation}\label{eqn:lem1206}
    |\lambda(\bmu)|\geq a\cdot\|\bmu\|\geq a\cdot(1-q)^k\cdot d_0
\end{equation}
and 
\begin{equation}\label{eqn:lem1207}
    |\lambda(\bmu)-\lambda(\bm{\xi})|\leq\mathbb{E}_{\tilde{\bx}}[u(\tilde{\bx},\bm{\xi},d)]\leq bd\leq b(1-q)^kq\cdot d_0.
\end{equation}
We have
\begin{equation}\label{eqn:lem1208}
Z_n(\bmu,0)\leq Z_n(\bmu,\bm{\xi})+\frac{\left| \sum_{j=1}^n[\psi(\bx_j,\bm{\xi})-\psi(\bx_j,0)-\lambda(\bm{\xi})] \right|}{\sqrt{n}+n|\lambda(\bmu)|},
\end{equation}
hence
\begin{equation}\label{eqn:lem1209}
    \sup_{\bmu\in C_{(l)}}Z_n(\bmu,0)\leq U_n+V_n
\end{equation}
with 
\begin{equation}\label{eqn:lem1210}
    U_n=\frac{\sum_{j=1}^n[u(\bx_j,\bm{\xi},d)+\mathbb{E}_{\tilde{\bx}}[u(\tilde{\bx},\bm{\xi},d)]]}{na(1-q)^k\cdot d_0},
\end{equation}
and
\begin{equation}\label{eqn:lem1211}
    V_n=\frac{\sum_{j=1}^n[\psi(\bx_j,\bm{\xi})-\psi(\bx_j,0)-\lambda(\bm{\xi})]}{na(1-q)^k\cdot d_0}.
\end{equation}
Thus,
\begin{equation}\label{eqn:lem1212}
    P(U_n\geq\epsilon)=P\left( \sum_{j=1}^n[u(\bx_j,\bm{\xi},d)-\mathbb{E}_{\tilde{\bx}}[u(\tilde{\bx},\bm{\xi},d)]]\geq\epsilon na(1-q)^k\cdot d_0-2n\mathbb{E}_{\tilde{\bx}}[u(\tilde{\bx},\bm{\xi},d)] \right).
\end{equation}
In view of \eqref{eqn:lem1207} and \eqref{eqn:lem1202}, 
\begin{equation}\label{eqn:lem1213}
    \epsilon a(1-q)^k\cdot d_0-2\mathbb{E}_{\tilde{\bx}}[u(\tilde{\bx},\bm{\xi},d)]\geq\epsilon a(1-q)^k\cdot d_0-2bq(1-q)^k\cdot d_0\geq bq(1-q)^k\cdot d_0.
\end{equation}
Then, we know that
\begin{equation}\label{eqn:112701}
P(U_n\geq\epsilon) \leq P\left( \sum_{j=1}^n[u(\bx_j,\bm{\xi},d)-\mathbb{E}_{\tilde{\bx}}[u(\tilde{\bx},\bm{\xi},d)]]\geq nbq(1-q)^k\cdot d_0 \right).
\end{equation}
We now apply Bernstein's inequality (\Cref{lem:Berstein}) to bound the right hand side of \eqref{eqn:112701}. From condition (N-3) (iii) we know that $\mathbb{E}_{\tilde{\bx}}[u(\tilde{\bx},\bmu,d)^2]\leq c_1\cdot d= c_1(1-q)^{k-1}q d_0/2$. From condition (N-3) (iv) we know that $|u(\tilde{\bx},\bmu,d)|\leq c_2$ almost surely. Therefore, we have the following bound as the results of Bernstein's inequality (\Cref{lem:Berstein}).
\begin{equation}\label{eqn:lem1214}
\begin{aligned}
    P(U_n\geq\epsilon)&\leq P\left( \sum_{j=1}^n[u(\bx_j,\bm{\xi},d)-\mathbb{E}_{\tilde{\bx}}[u(\tilde{\bx},\bm{\xi},d)]]\geq nbq(1-q)^k\cdot d_0 \right)\\
    &\leq \exp\left(-\frac{b^2q^2(1-q)^{2k}d_0^2n^2}{c_1n(1-q)^{k-1}q d_0+2c_2nbq(1-q)^kd_0/3 }\right)=\exp\left(-\frac{b^2q^2(1-q)^{k+1}d_0n}{c_1q +2c_2bq(1-q)/3 }\right)\\
    &\leq\exp\left(-\frac{b^2q^2(1-q)^{k_0+1}d_0n}{c_1q +2c_2bq(1-q)/3 }\right)\leq \exp\left(-\frac{b^2q(1-q)^{2}n^{1-\gamma'}}{c_1 +2c_2b(1-q)/3 }\right),
\end{aligned}
\end{equation}
where the last inequality follows from \eqref{eqn:lem1203} that $d_0(1-q)^{k_0+1}\geq n^{-\gamma'}\cdot(1-q)^2$.

In a similar way, we apply the Bernstein inequality to bound $P(V_n\geq\eps)$. We note that 
\[
\mathbb{E}_{\tilde{\bx}}[(\psi(\tilde{\bx},\bm{\xi})-\psi(\tilde{\bx},0))^2]\leq \mathbb{E}_{\tilde{x}}[(u(\tilde{\bx}, \bm{\xi}, |\bm{\xi}|))^2]\leq c_1\cdot |\bm{\xi}|=c_1 (1-q)^{k-1}(1-\frac{q}{2})\cdot d_0,
\]
where the last inequality follows from the condition (N-3) (iii). Also, from the condition (N-3) (iv) we have that
\[
|\psi(\tilde{\bx},\bm{\xi})-\psi(\tilde{\bx},0)|\leq u(\tilde{\bx}, \bm{\xi}, |\bm{\xi}|) \leq c_2.
\]
Therefore, from Bernstein's inequality (\Cref{lem:Berstein}),  
we have
\begin{equation}\label{eqn:lem1215}
\begin{aligned}
    P(V_n\geq\epsilon)&\leq \exp\left(-\frac{\eps^2a^2n(1-q)^{k+1}d_0}{c_1(2-q)+2c_2\eps a (1-q)/3} \right) \leq \exp\left(-\frac{\eps^2a^2n(1-q)^{k_0+1}d_0}{c_1(2-q)+2c_2\eps a (1-q)/3} \right)\\
    &\leq \exp\left(-\frac{\eps^2a^2(1-q)^{2}n^{1-\gamma'}}{c_1(2-q)+2c_2\eps a (1-q)/3} \right),
\end{aligned}
\end{equation}
where the last inequality follows from \eqref{eqn:lem1203} that $d_0(1-q)^{k_0+1}\geq n^{-\gamma'}\cdot(1-q)^2$.
Hence, we obtain from \eqref{eqn:lem1203}, \eqref{eqn:lem1209}, \eqref{eqn:lem1214} and \eqref{eqn:lem1215} that
\begin{equation}\label{eqn:lem1216}
\begin{aligned}
    P\left(\sup_{\bmu\in C_{(j)}}Z_n(\bmu,0)\geq2\epsilon\right)&\leq \exp\left(-\frac{b^2q(1-q)^{2}n^{1-\gamma'}}{c_1 +2c_2b(1-q)/3 }\right)+\exp\left(-\frac{\eps^2a^2(1-q)^{2}n^{1-\gamma'}}{c_1(2-q)+2c_2\eps a (1-q)/3} \right)\\
    &\leq 2\exp\left(-\frac{\min\{b^2q, \eps^2a^2\}\cdot(1-q)^{2}n^{1-\gamma'}}{c_1(3-q) +2c_2(b+\eps a)(1-q)/3 }\right).
\end{aligned}
\end{equation}
Furthermore, 
\begin{equation}\label{eqn:lem1217}
    \sup_{\bmu\in C_{k_0}}Z_n(\bmu,0)\leq\frac{\sum_{j=1}^n[u(\bx_j,0,d)+\mathbb{E}_{\tilde{\bx}}[u(\tilde{\bx},0,d)]]}{\sqrt{n}}
\end{equation}
with $d=(1-q)^{k_0}\cdot d_0\leq n^{-\gamma'}$. Hence,
\begin{equation}\label{eqn:lem1218}
    P\left( \sup_{\bmu\in C_{k_0}}Z_n(\bmu,0)\geq2\epsilon \right)\leq P\left( \sum_{j=1}^n[u(\bx_j,0,d)-\mathbb{E}_{\tilde{\bx}}[u(\tilde{\bx},0,d)]]\geq2\sqrt{n}\epsilon-2n\mathbb{E}_{\tilde{\bx}}[u(\tilde{\bx},0,d)] \right).
\end{equation}
Since $\mathbb{E}_{\tilde{\bx}}[u(\tilde{\bx},0,d)]\leq bd\leq bn^{-\gamma'}$, set $n_0$ such that $n_0^{\gamma'-\frac{1}{2}}=\frac{2b}{\epsilon}$. Then, for $n\geq n_0$, we have
\[
\mathbb{E}_{\tilde{\bx}}[u(\tilde{\bx},0,d)]\leq\frac{\epsilon}{2\sqrt{n}}\Rightarrow 2\sqrt{n}\epsilon-2n\mathbb{E}_{\tilde{\bx}}[u(\tilde{\bx},0,d)]\geq\sqrt{n}\epsilon;
\]
thus, by Chebyshev's inequality,
\begin{equation}\label{eqn:lem1219}
    P\left( \sup_{\bmu\in C_{k_0}}Z_n(\bmu,0)\geq2\epsilon \right)\leq c_1\cdot\epsilon^{-2}\cdot n^{-\gamma'}.
\end{equation}
Now, putting \eqref{eqn:lem1201}, \eqref{eqn:lem1205}, \eqref{eqn:lem1216}, and \eqref{eqn:lem1219} together, we obtain
\[
P\left( \sup_{\bmu\in C_0} Z_n(\bmu,0)\geq2\epsilon \right)\leq c_1\epsilon^{-2} n^{-\gamma'}+2\exp\left(-\frac{\min\{b^2q, \eps^2a^2\}\cdot(1-q)^{2}n^{1-\gamma'}}{c_1(3-q) +2c_2(b+\eps a)(1-q)/3 }\right)\cdot \left(\frac{\gamma'\cdot(\log n+\log d_0)}{|\log(1-q)|}+1\right)\cdot (2M)^m
\]
as long as $n\geq n_0$ where $n_0$ satisfying $n_0^{\gamma'-\frac{1}{2}}=\frac{2b}{\epsilon}$, 
which completes our proof of the lemma.
\end{myproof}

\section{Missing Proofs for \Cref{sec:knowndistribution}}\label{sec:missingpf}
\begin{myproof}[Proof of \Cref{lem:threshold}]
We now fix an arbitrary $j\in[n]$. Suppose that there exists two points $r_1\geq r_2$ such that $x^*_j(r_1)<1$ while $x^*_j(r_2)>0$. Then, denote by $\delta=\min\{ 1-x^*_j(r_1), x^*_j(r_2) \}$ and we define a new set of solution
\[
\hat{x}^*_j(r_1)=x^*_j(r_1)+\delta,~\hat{x}^*_j(r_2)=x^*_j(r_2)-\delta,~\text{and~}\hat{x}^*_{j'}(r')=x^*_{j'}(r'),~\forall r'\neq r_1, r_2, \forall j'\neq j. 
\]
It is easy to see that $\{\hat{x}^*_j(r), \forall j\in[n], \forall r\}$ is still a feasible solution to \eqref{lp:Prophrelax}. However, the objective value under the solution $\{\hat{x}^*_j(r), \forall j\in[n], \forall r\}$ can only become larger in that
\[
\sum_{j=1}^n d_{j}\cdot\mathbb{E}_{r\sim F_j}[r\cdot \hat{x}^*_j(r)]\geq \sum_{j=1}^n d_{j}\cdot\mathbb{E}_{r\sim F_j}[r\cdot x^*_j(r)].
\]
Therefore, we conclude that $\{\hat{x}^*_j(r), \forall j\in[n], \forall r\}$ is still an optimal solution to \eqref{lp:Prophrelax}. Keep operating as above, we can transfer any optimal solution that does not possess the threshold property into an optimal solution that enjoys the threshold property, as described in the statement of \Cref{lem:threshold}. Our proof is thus completed.
\end{myproof}

\begin{myproof}[Proof of \Cref{lem:perturbationLP}]
Our proof can be classifed into three steps. We fix $\{\hat{q}^*_j\}_{j=1}^n$ as an optimal solution to \eqref{lp:exante}. In the first step,
we discretize both the convex optimization problem \eqref{lp:newProphrelax} and \eqref{lp:exante} into two LPs with a granularity $K$ such that $\{\hat{q}^*_j\}_{j=1}^n$ is an optimal solution to the discretized LP of \eqref{lp:exante}. In the second step, we show that we can select one optimal solution to the discretized LP of \eqref{lp:newProphrelax}, such that the gap between the selected optimal solution and $\{\hat{q}^*_j\}_{j=1}^n$ can be bounded by a constant independent of the granularity $K$. In the final step, we show that as the granularity $K$ grows to infinity, there exists a subsequence of $K$ such that the selected optimal solution to the discretized LP of \eqref{lp:newProphrelax} will converge to an optimal solution of \eqref{lp:newProphrelax}, which completes our proof.

We now do the first step to discretize the convex optimization problem \eqref{lp:newProphrelax} and \eqref{lp:exante}. For each $j\in[n]$, we denote by a function
\[
G_j(q)=\int_{q'=1-q}^1 F^{-1}_j(q')dq'.
\]

For any integer $K\geq n+1$, we denote by a set $\{0,\frac{1}{K-n}, \frac{2}{K-n},\dots, 1\}\cup\{\hat{q}^*_j\}_{j=1}^n$ and let $q^K_k$ to be the $k$-th smallest element in this set, for $k=1,2,\dots,K+1$. For each $j\in[n]$, we denote by $\hat{G}^K_j(\cdot)$ the piece-wise linear interpolation of $G_j$ based on the values at points $\{q^K_k\}_{k=1}^{K+1}$. From the concavity of the function $G_j(\cdot)$, it is clear that we have
\begin{equation}\label{eqn:Constpiecewise}
\hat{G}^K_j(q)\leq G_j(q)\leq \hat{G}^K_j(q)+\frac{\max_{j\in[n]}\{u_j\}}{K-n},~\forall q\in[0,1],~\text{and~}\hat{G}^K_j(\hat{q}^*_j)=G_j(\hat{q}^*_j).
\end{equation}
Then, we know that $\{\hat{q}^*_j\}_{j=1}^n$ is an optimal solution to the following optimization problem
\begin{eqnarray}
\hV^K_{t,\bc}=&\max &\sum_{j=1}^n p_j\cdot s\cdot \hat{G}^K_j(q_j)  \label{lp:0Discreteexante}\\
&\mbox{s.t.} & \sum_{j=1}^n p_j\cdot s\cdot  a_{j,i}\cdot q_j\leq c_i,~~\forall i\in[m]\nonumber\\
&& q_j\in[0,1],~~\forall j\in[n] \nonumber
\end{eqnarray}
{because optimization problem~\eqref{lp:0Discreteexante} differs from~\eqref{lp:exante} only by having a pointwise-dominated objective function, and $\{\hat{q}^*_j\}_{j=1}^n$ attains the optimal objective value from~\eqref{lp:exante} in the dominated problem~\eqref{lp:0Discreteexante}.}
Without loss of generality, we assume that for each $j\in[n]$, the piece-wise linear functions $\hat{G}^K_j(\cdot)$ share the same set of end points of the piece-wise linear intervals, and we denote by $\{q^K_k\}_{k=1}^{\hK+1}$ the set of end points, with $q^K_1=0$ and $q^K_{\hK+1}=1$. Then, we have the following linear programming as a re-formulation of the discretization $\hV^K_{t,\bc}$ of the convex problem $\bVF_{t,\bc}$ \eqref{lp:exante}: 
\begin{eqnarray}
\hV^K_{t,\bc}=&\max &\sum_{j=1}^n\sum_{k=1}^{\hK} \beta_{j,k}\cdot x^K_{k,j}  \label{lp:Discreteexante}\\
&\mbox{s.t.} & \sum_{j=1}^n\sum_{k=1}^{\hK}  a_{j,i}\cdot x^K_{k,j}\leq c_i,~~\forall i\in[m]\nonumber\\
&& 0\leq x^K_{k,j}\leq p_j\cdot s\cdot (q^K_{k+1}-q^K_k),~~\forall j\in[n], \forall k\in[\hK] \nonumber
\end{eqnarray}
where $\beta_{j,k}$ is the coefficient that is inherited from the piece-wise linear function $\hat{G}^K_j$.
Here, the variable $x^K_{k,j}$ can be interpreted as the number of queries,with type $j$ and reward realization quantile lying in the interval $[q^K_k, q^K_{k+1}]$, being served in the relaxation $\bVF_{t,\bc}$ \eqref{lp:exante1}. We also denote by the following linear programming as a discretization of the convex problem $\bVH_{\bc}(\bd)$ \eqref{lp:newProphrelax}, for each $\bd$:
\begin{eqnarray}
\bV^K_{\bc}(\bd)=&\max &\sum_{j=1}^n\sum_{k=1}^{\hK} \beta_{j,k}\cdot x^K_{k,j}  \label{lp:DiscretenewProphrelax}\\
&\mbox{s.t.} & \sum_{j=1}^n\sum_{k=1}^{\hK}  a_{j,i}\cdot x^K_{k,j}\leq c_i,~~\forall i\in[m]\nonumber\\
&& 0\leq x^K_{k,j}\leq d_j\cdot(q^K_{k+1}-q^K_k),~~\forall j\in[n], \forall k\in[\hK]. \nonumber
\end{eqnarray}

We now do the second step and compare one optimal solution of $\hV^K_{t,\bc}$ to another optimal solution of $\bV^K_{\bc}(\bd)$. Our result is formalized in the following claim. Our analysis follows the analysis in \citet{mangasarian1987lipschitz} over the Lipschitz continuity of solutions of linear programming, and we further show an equivalence between two linear systems to obtain a bound that is independent of the granularity $K$.
\begin{claim}\label{claim:Lipschitz}
There exists a constant $\mu$ such that for any $K$, for any optimal solution $\{\hat{x}^K_{k,j}\}$ of $\hV^K_{t,\bc}$ \eqref{lp:Discreteexante}, we can select one optimal solution $\{\bar{x}^K_{k,j}\}$ of $\bV_{\bc}^K(\bd)$ \eqref{lp:DiscretenewProphrelax} such that
\begin{equation}\label{eqn:Lipschitz}
\left|\sum_{k=1}^{\hK}\hat{x}^K_{k,j}-\bar{x}^K_{k,j}\right|\leq \mu\cdot\max_{j'\in[n]}\{ (p_{j'}\cdot s-d_{j'})\},~~\forall j\in[n].
\end{equation}
where 
\begin{equation}\label{const:Lipschitz}
\mu=\sup_{\bm{v}^1\in\mathbb{R}^m, \bm{v}^2, \bm{v}^3\in\mathbb{R}^n}\left\{\left\|
\begin{aligned}
&\bm{v}^1\\
&\bm{v}^2\\
&\bm{v}^3
\end{aligned}
\right\|_{1} \left|\begin{aligned}
&\|(\bm{v}^1)^\top A+(\bm{v}^2)^\top-(\bm{v}^3)^\top\|_1=1,\\
&\text{Rows of }\begin{bmatrix}
&A\\
&I_{n}\\
&-I_{n}
\end{bmatrix}
\text{ corresponding to non-zero elements}\\
&\text{of }\begin{pmatrix}
&\bm{v}^1\\
&\bm{v}^2\\
&\bm{v}^3
\end{pmatrix}
\text{ are linear independent.}
\end{aligned}
\right.\right\}
\end{equation}
$A=(a_{i,j})_{\forall i\in[m], j\in[n]}\in\mathbb{R}^{m\times n}$ and $I_{n}$ is an identity matrix with a size $n\times n$.
\end{claim}
We do the third step to complete our proof. For any granularity $K$, we let $\{\hat{x}^K_{k,j}\}$ be a solution of $\hV^K_{t,\bc}$ \eqref{lp:Discreteexante} satisfying
\[
\hat{x}^{K}_{k,j}=\left\{\begin{aligned}
&p_j\cdot s\cdot(q^K_{k+1}-q^K_k), &&\text{if~}q^K_k\geq 1-\hat{q}^*_j\\
&\left(q^K_{k+1}-(1-\hat{q}^*_j)\right)\cdot p_j\cdot s, &&\text{if~}q^K_k< 1-\hat{q}^*_j\leq q^K_{k+1}\\
&0, &&\text{if~}q^K_{k+1}<1-\hat{q}^*_j.
\end{aligned}\right.
\]
Since $\{\hat{q}^*_j\}_{j=1}^n$ is an optimal solution to $\hV^K_{t,\bc}$ under the formulation \eqref{lp:0Discreteexante}, we must have $\{\hat{x}^K_{k,j}\}$ is an optimal solution of $\hV^K_{t,\bc}$ under the formulation \eqref{lp:Discreteexante}. Then, we denote by
 $\{\bar{x}^K_{k,j}\}$ one optimal solution of $\bV_{\bc}^K(\bd)$ \eqref{lp:DiscretenewProphrelax}, as specified in \Cref{claim:Lipschitz}. We further construct
\begin{equation}\label{eqn:100201}
    \bar{q}^K_j=\frac{\sum_{k=1}^{\hK}\bar{x}^K_{k,j}}{d_j},~~\forall j\in[n].
\end{equation}
We denote by $\hat{\bm{q}}^*=(\hat{q}^*_{1},\dots,\hat{q}^*_{n})$ and $\bar{\bm{q}}^K=(\bar{q}^K_{1},\dots,\bar{q}^K_{n})$. From definition, we know that $\hat{\bm{q}}^*, \bar{\bm{q}}^K\in[0,1]^K$. Therefore, there exists a point $\bar{\bm{q}}'$ and a sequence of integers $\{K_1,\dots,K_w,\dots\}_{w=1,2,\dots}$ such that
\begin{equation}\label{eqn:100202}
    \bar{\bm{q}}'=\lim_{w\rightarrow\infty}\bar{\bm{q}}^{K_w}.
\end{equation}
We show in the following claim that $\bar{\bm{q}}'$ is an optimal solution to $\bVH_{\bc}(\bd)$ \eqref{lp:newProphrelax}.
\begin{claim}\label{claim:Optimal}
Let $\bar{\bm{q}}'$ be constructed in \eqref{eqn:100202}. Then, $\bar{\bm{q}}'$ is an optimal solution to $\bVH_{\bc}(\bd)$ \eqref{lp:newProphrelax}.
\end{claim}
For any integer $w$, from \Cref{claim:Lipschitz}, we know that
\[\begin{aligned}
\|\hat{\bm{q}}^*-\bar{\bm{q}}^{K_w}\|_{\infty}&=\left\|\frac{\sum_{k=1}^{\hK_w}\hat{x}^{K_w}_{k,j}}{p_j\cdot s}-\frac{\sum_{k=1}^{\hK_w}\bar{x}^{K_w}_{k,j}}{d_j}\right\|_{\infty}\leq \left\|\frac{\sum_{k=1}^{\hK_w}\hat{x}^{K_w}_{k,j}}{p_j\cdot s}-\frac{\sum_{k=1}^{\hK_w}\bar{x}^{K_w}_{k,j}}{p_j\cdot s}\right\|_{\infty}+\left\|\frac{\sum_{k=1}^{\hK_w}\bar{x}^{K_w}_{k,j}}{p_j\cdot s}-\frac{\sum_{k=1}^{\hK_w}\bar{x}^{K_w}_{k,j}}{d_j}\right\|_{\infty}\\
&\leq \max_{j\in[n]}\left\{\frac{\mu}{p_j}\right\}\cdot \max_{j'\in[n]}\{ (p_{j'}-d_{j'}/s)\}+\max_{j\in[n]}\left\{\left|\frac{d_j}{p_j\cdot s}-1\right|\right\}.
\end{aligned}\]
Therefore, from \eqref{eqn:100202} and \Cref{claim:Optimal}, there exists an optimal solution $\bar{\bm{q}}'$ of $\bVH_{\bc}(\bd)$ \eqref{lp:newProphrelax} such that
\[
\|\hat{\bm{q}}^*-\bar{\bm{q}}'\|_{\infty}\leq \max_{j\in[n]}\left\{\frac{\mu}{p_j}\right\}\cdot \max_{j'\in[n]}\{ (p_{j'}-d_{j'}/s)\}+\max_{j\in[n]}\left\{\left|\frac{d_j}{p_j\cdot s}-1\right|\right\}.
\]
Our proof of \eqref{eqn:102601} is thus completed. We now bound the constant $\mu$ and thus bound the constant $\kappa_1$.
From Theorem 2.2 of \cite{mangasarian1987lipschitz}, we know that $\mu$ in \eqref{const:Lipschitz} is finite. We now derive an upper bound of it.

From Proposition 2.6 of \cite{mangasarian1987lipschitz}, we know that 
\begin{equation}\label{eqn:112301}
    \mu\leq(m+2n)\cdot \Delta A
\end{equation}
where $\Delta A$ denotes the maximum of the absolute values of the determinants of the square submatrices of $\begin{bmatrix}
&A\\
&I_{n}\\
&-I_{n}
\end{bmatrix}$.

We denote by $A'$ to be a square submatrix of $\begin{bmatrix}
&A\\
&I_{n}\\
&-I_{n}
\end{bmatrix}$. Following the Hadamard's inequality on matrix determinant (\Cref{lem:hadamard}), we have that
\[
|\text{det}(A')|=\prod_{i\in[n']}\|\bm{a}'_i\|_2
\]
where $n'$ denotes the size of matrix $A'$ and $\bm{a}'_i$ denotes the $i$-th column of matrix $A'$. Therefore, it is clear to see that
\[
|\text{det}(A')|=\prod_{i\in[n']}\|\bm{a}'_i\|_2\leq \bar{a}^{n'}\leq \bar{a}^n
\]
for any submatrix $A'$, which completes our proof.
\end{myproof}
\begin{myproof}[Proof of \Cref{claim:Lipschitz}]
We denote the linear programming $\hV^K_{t,\bc}$ \eqref{lp:Discreteexante} by
\begin{eqnarray}
\hV^K_{t,\bc}=&\max &\bm{h}^\top\bm{x}\label{lp:abstract1}\\
&~\mbox{s.t.} &\bm{l}'\leq\mathcal{A}\bm{x}\leq \bm{u}'\nonumber
\end{eqnarray}
and denote the linear programming $\bV^K_{\bc}(\bd)$ \eqref{lp:DiscretenewProphrelax} by
\begin{eqnarray}
\bV^K_{\bc}(\bd)=&\max &\bm{h}^\top\bm{x}\label{lp:abstract2}\\
&~\mbox{s.t.} &\bm{l}''\leq\mathcal{A}\bm{x}\leq \bm{u}''.\nonumber
\end{eqnarray}
Fix an optimal solution $\hat{\bm{x}}^K$ of $\hV^K_{t,\bc}$. We denote by $J_1 ,J_2$ and $J_3$ three row index sets of $\mathcal{A}$ such that
\[
\mathcal{A}_{J_1}\hat{\bm{x}}^K=\bm{u}'_{J_1}, \mathcal{A}_{J_2}\hat{\bm{x}}^K=\bm{l}'_{J_2}, \text{~and~}\bm{u}'_{J_3}\mathcal{A}_{J_3}\hat{\bm{x}}^K<\bm{b}'_{J_3}.
\]
We further fix an optimal solution $\bar{\bm{x}}^{K'}$ of $\bV^K_{\bc}(\bd)$ and denote by $J_1=J_{1,1}\cup J_{1,2}$, $J_2=J_{2,1}\cup J_{2,2}$ such that
\[
\mathcal{A}_{J_{1,1}}\bar{\bm{x}}^{K'}=\bm{u}''_{J_{1,1}},~ \mathcal{A}_{J_{1,2}}\bar{\bm{x}}^{K'}<\bm{u}''_{J_{1,2}},~ \mathcal{A}_{J_{2,1}}\bar{\bm{x}}^{K'}=\bm{l}''_{J_{2,1}}\text{~and~}\mathcal{A}_{J_{2,2}}\bar{\bm{x}}^{K'}>\bm{l}''_{J_{2,2}}.
\]
Then, we denote by a set of linear equalities and linear inequalities
\begin{eqnarray}
&\mathcal{A}_{J_{1,1}}{\bm{x}}=\bm{u}''_{J_{1,1}} & \label{linearsystem1}\\
&\mathcal{A}_{J_{1,2}}\bar{\bm{x}}^{K'}\leq \mathcal{A}_{J_{1,2}}{\bm{x}}, & \mathcal{A}_{J_{1,2}}{\bm{x}}\leq \bm{u}''_{J_{1,2}} \nonumber\\
&\mathcal{A}_{J_{2,1}}{\bm{x}}=\bm{l}''_{J_{2,1}} & \nonumber\\
&\mathcal{A}_{J_{2,2}}\bar{\bm{x}}^{K'}\geq \mathcal{A}_{J_{2,2}}{\bm{x}}, & \mathcal{A}_{J_{2,2}}{\bm{x}}\geq \bm{l}''_{J_{1,2}} \nonumber\\
&\bm{l}''_{J_3}\leq\mathcal{A}_{J_3}{\bm{x}}\leq\bm{u}''_{J_3}. & \nonumber
\end{eqnarray}
It is clear that $\bar{\bm{x}}^{K'}$ satisfies the linear system \eqref{linearsystem1}. On the other hand, $\hat{\bm{x}}^K$ satisfies the following linear system:
\begin{eqnarray}
&\mathcal{A}_{J_{1,1}}{\bm{x}}=\bm{u}'_{J_{1,1}} & \label{linearsystem2}\\
&\bm{u}'_{J_{1,2}}-\bm{u}''_{J_{1,2}}+\mathcal{A}_{J_{1,2}}\bar{\bm{x}}^{K'}\leq \mathcal{A}_{J_{1,2}}{\bm{x}}, & \mathcal{A}_{J_{1,2}}{\bm{x}}\leq \bm{u}'_{J_{1,2}} \nonumber\\
&\mathcal{A}_{J_{2,1}}{\bm{x}}=\bm{l}'_{J_{2,1}} & \nonumber\\
&\bm{l}'_{J_{2,2}}-\bm{l}''_{J_{2,2}}+\mathcal{A}_{J_{2,2}}\bar{\bm{x}}^{K'}\geq \mathcal{A}_{J_{2,2}}{\bm{x}}, & \mathcal{A}_{J_{2,2}}{\bm{x}}\geq \bm{l}'_{J_{2,2}} \nonumber\\
&\bm{l}'_{J_3}\leq \mathcal{A}_{J_3}{\bm{x}}\leq\bm{u}'_{J_3}. & \nonumber
\end{eqnarray}
Our remaining analysis can be classified into two steps. For the first step, we show that any variable $\bm{x}$ that satisfies the linear system \eqref{linearsystem1} turns out to be an optimal solution of $\bV^K_{\bc}(\bd)$ \eqref{lp:abstract2}. For the second step, we show that for $\hat{\bm{x}}^K$ that satisfies the linear system \eqref{linearsystem2}, we can find a variable $\bar{\bm{x}}^K$ satisfying the linear system \eqref{linearsystem1} such that \eqref{eqn:Lipschitz} holds. 

We now prove the first step. Since $\hat{\bm{x}}^K$ is an optimal solution of $\hV^K_{t,\bc}$ \eqref{lp:abstract1}, from the KKT optimality condition, we know that there exists dual variables $\bmu_{J_1}\geq0$, $\bm{v}_{J_2}\leq 0$ such that 
\[
\mathcal{A}_{J_1}^\top\bmu_{J_1}+\mathcal{A}_{J_2}^\top\bm{v}_{J_2}=\bm{h}.
\]
Then, for any $\bm{x}$ that satisfies the linear system \eqref{linearsystem1}, we have
\[\begin{aligned}
\bm{h}^\top\bm{x}&=\bmu^\top_{J_1}\mathcal{A}_{J_1}\bm{x}+\bm{v}^\top_{J_2}\mathcal{A}_{J_2}\bm{x}
=\bmu^\top_{J_{1,1}}\mathcal{A}_{J_{1,1}}\bm{x}+\bmu^\top_{J_{1,2}}\mathcal{A}_{J_{1,2}}\bm{x}+\bm{v}^\top_{J_{2,1}}\mathcal{A}_{J_{2,1}}\bm{x}+\bm{v}^\top_{J_{2,2}}\mathcal{A}_{J_{2,2}}\bm{x}
\\
&\geq\bmu^\top_{J_{1,1}}\mathcal{A}_{J_{1,1}}\bar{\bm{x}}^{K'}+\bmu^\top_{J_{1,2}}\mathcal{A}_{J_{1,2}}\bar{\bm{x}}^{K'}+\bm{v}^\top_{J_{2,1}}\mathcal{A}_{J_{2,1}}\bar{\bm{x}}^{K'}+\bm{v}^\top_{J_{2,2}}\mathcal{A}_{J_{2,2}}\bar{\bm{x}}^{K'}=\bmu^\top_{J_1}\mathcal{A}_{J_1}\bar{\bm{x}}^{K'}+\bm{v}^\top_{J_{2}}\mathcal{A}_{J_{2}}\bar{\bm{x}}^{K'}\\
&=\bm{h}^\top\bar{\bm{x}}^{K'}=\bV^K_{t,\bc}(\bd)
\end{aligned}\]
where the first inequality follows from both $\bm{x}$, $\bar{\bm{x}}^{K'}$ satisfies the linear system \eqref{linearsystem1} and thus $\mathcal{A}_{J_{1,1}}\bm{x}=\mathcal{A}_{J_{1,1}}\bar{\bm{x}}^{K'}=\bm{u}''_{J_{1,1}}$, $\mathcal{A}_{J_{2,1}}\bm{x}=\mathcal{A}_{J_{2,1}}\bar{\bm{x}}^{K'}=\bm{l}''_{J_{2,1}}$, $\mathcal{A}_{J_{1,2}}\bar{\bm{x}}^{K'}\leq \mathcal{A}_{J_{1,2}}{\bm{x}}$ and $\mathcal{A}_{J_{2,2}}\bar{\bm{x}}^{K'}\geq \mathcal{A}_{J_{2,2}}{\bm{x}}$.

It only remains to show the second step. We prove by exploiting the special structure of the linear systems \eqref{linearsystem1} and \eqref{linearsystem2}. Note that under our setting, we have
\[
\mathcal{A}=\begin{bmatrix}
& A & \dots & A\\
& &I_{n\hK} &
\end{bmatrix}
\]
where $I_{n\hK}$ denotes an identity matrix with size $n\hK\times n\hK$ and $A=(a_{i,j})_{i\in[m], j\in[n]}\in\mathbb{R}^{m\times n}$. Also, the linear system \eqref{linearsystem1} can be rewritten as
\begin{equation}\label{newsystem1}
\bm{l}^1\leq \mathcal{A}\bm{x}\leq \bm{u}^1
\end{equation}
and the linear system \eqref{linearsystem2} can be rewritten as
\begin{equation}\label{newsystem2}
\bm{l}^2\leq \mathcal{A}\bm{x}\leq \bm{u}^2.
\end{equation}
Then, for the solution $\hat{\bm{x}}^K$ of the linear system \eqref{newsystem2}, we construct $\hat{\bm{y}}\in\mathbb{R}^n$ with
\[
\hat{y}_j=\sum_{k=1}^{\hK} \hat{x}^K_{k,j},~~\forall j\in[n].
\]
It is clear that $\hat{\bm{y}}$ is a solution to the following linear system
\begin{equation}\label{newAbstract1}
\hat{\bm{l}}^2\leq \begin{bmatrix}
&A\\
&I_n
\end{bmatrix} \bm{y} \leq \hat{\bm{u}}^2
\end{equation}
where $\hat{\bm{l}}^2\in\mathbb{R}^{n+m}$ satisfying
\[
\hat{l}^2_i=l^2_i,~\forall i\in[m]\text{~and~}\hat{l}^2_{n+j}=\sum_{k=1}^{\hK}l^2_{j+kn},~\forall j\in[n]
\]
and $\hat{\bm{u}}^2\in\mathbb{R}^{n+m}$ satisfying
\[
\hat{u}^2_i=u^2_i,~\forall i\in[m]\text{~and~}\hat{u}^2_{n+j}=\sum_{k=1}^{\hK}u^2_{j+kn},~\forall j\in[n].
\]
In the same way, we denote by $\hat{\bm{l}}^1\in\mathbb{R}^{n+m}$ satisfying
\[
\hat{l}^1_i=l^1_i,~\forall i\in[m]\text{~and~}\hat{l}^1_{n+j}=\sum_{k=1}^{\hK}l^1_{j+kn},~\forall j\in[n]
\]
and $\hat{\bm{u}}^1\in\mathbb{R}^{n+m}$ satisfying
\[
\hat{u}^1_i=u^1_i,~\forall i\in[m]\text{~and~}\hat{u}^1_{n+j}=\sum_{k=1}^{\hK}u^1_{j+kn},~\forall j\in[n].
\]
We consider the linear system
\begin{equation}\label{newAbstract2}
\hat{\bm{l}}^1\leq \begin{bmatrix}
&A\\
&I_n
\end{bmatrix} \bm{y} \leq \hat{\bm{u}}^1.
\end{equation}
From \Cref{lem:Lipschitz}, we know that for the solution $\hat{\bm{y}}$ that satisfies linear system \eqref{newAbstract1}, there exists a solution $\bar{\bm{y}}$ that satisfies linear system \eqref{newAbstract2} and there also exists a constant $\mu$ such that
\begin{equation}\label{eqn:100301}
\|\hat{\bm{y}}-\bar{\bm{y}}\|_{\infty}\leq \mu\cdot \|(\hat{\bm{l}}^1, \hat{\bm{u}}^1)- (\hat{\bm{l}}^2, \hat{\bm{u}}^2)\|_{\infty}\leq\mu\cdot \max_{j'\in[n]}\{ (p_{j'}\cdot s-d_{j'})\}. 
\end{equation}
Note that here the constant $\mu$ depends solely on $A$ and $I_n$ and is independent of the granularity $K$. We now construct a solution to the linear system \eqref{newsystem1} from $\bar{\bm{y}}$ to complete our proof. For each $j\in[n]$ and each $k\in[\hK]$, we define
\[
\bar{x}^K_{k,j}=l^1_{j+kn}+(\bar{y}_j-\hat{l}^1_{n+j})\cdot\frac{u^1_{j+kn}-l^1_{j+kn}}{\hat{u}^1_{n+j}-\hat{l}^1_{n+j}}.
\]
It is clear to see that $\bar{\bm{x}}^K$ satisfies linear system \eqref{newsystem1} and satisfies
\begin{equation}\label{eqn:100302}
\sum_{k=1}^{\hK} \bar{x}^K_{k,j}=\bar{y}_j,~~\forall j\in[n].
\end{equation}
Our proof of the second step is completed from \eqref{eqn:100301}, \eqref{eqn:100202} and the fact that $\bar{\bm{x}}^K$ satisfies linear system \eqref{newsystem1}. From the conclusion of the first step, we know that $\bar{\bm{x}}^K$ is an optimal solution to $\bar{V}^K_{\bc}(\bd)$ and our proof of \Cref{claim:Lipschitz} is thus completed.
\end{myproof}
\begin{myproof}[Proof of \Cref{claim:Optimal}]
For any integer $w$, it is clear to see that $\bar{\bm{q}}^{K_w}$ is a feasible solution to $\bVH_{\bc}(\bd)$ \eqref{lp:newProphrelax}, which implies that {the limiting point} $\bar{\bm{q}}'$ is a feasible solution to $\bVH_{\bc}(\bd)$ \eqref{lp:newProphrelax} {since the feasible set is closed}. We now prove optimality.

It is direct to see
from the construction of the piece-wise linear function $\hat{G}^K_j(\cdot)$ that
\[
\bV^{K_w}_{\bc}(\bd)=\sum_{j=1}^n\sum_{k=1}^{\hK_w} \beta^{K_w}_{j,k}\cdot \bar{x}^{K_w}_{k,j}\leq\sum_{j=1}^n d_j\cdot\int^1_{q=1-\bar{q}^{K_w}_j}F^{-1}_j(q)dq\leq\bVH_{\bc}(\bd)
\]
where the first inequality follows from the definition $\bar{q}^{K_w}_{j}=\frac{\sum_{k=1}^{K_w}\bar{x}^{K_w}_{k,j}}{p_j\cdot s}$. 

We now show that $\bVH_{\bc}(\bd)\leq \bV^{K_w}_{\bc}(\bd)+\frac{(\sum_{j=1}^nd_j)\cdot\max_{j\in[n]}\{u_j\}}{K_w}$. For an optimal solution $\{\bar{q}^*_{j}\}_{j=1}^n$ of $\bVH_{\bc}(\bd)$. We construct for any $k\in[\hK_w]$ and $j\in[n]$
\[
x^{K_w}_{k,j}=\left\{\begin{aligned}
&p_j\cdot s\cdot(q^{K_w}_{k+1}-q^{K_w}_k), &&\text{if~}q^{K_w}_k\geq 1-\bar{q}^*_j\\
&\left(q^{K_w}_{k+1}-(1-\bar{q}^*_j)\right)\cdot p_j\cdot s, &&\text{if~}q^{K_w}_k< 1-\bar{q}^*_j\leq q^{K_w}_{k+1}\\
&0, &&\text{if~}q^{K_w}_{k+1}<1-\bar{q}^*_j.
\end{aligned}\right.
\]
It is clear to see that $\{x^{K_w}_{k,j}\}$ is a feasible solution to $\bV^{K_w}_{\bc}(\bd)$ \eqref{lp:DiscretenewProphrelax} and  it holds that
\[\begin{aligned}
\bVH_{\bc}(\bd)&=\sum_{j=1}^nd_j\cdot G_j(\bar{q}^*_j)\leq \sum_{j=1}^nd_j\cdot\hat{G}^{K_w}_j(\bar{q}^*_j)+\frac{(\sum_{j=1}^nd_j)\cdot\max_{j\in[n]}\{u_j\}}{K_w-n}\\
&= \sum_{j=1}^n\sum_{k=1}^{\hK_w}d_j\cdot \beta^{K_w}_{j,k}\cdot x^{K_w}_{k,j}+\frac{(\sum_{j=1}^nd_j)\cdot\max_{j\in[n]}\{u_j\}}{K_w-n}\\
&\leq \bV^{K_w}_{\bc}(\bd)+\frac{(\sum_{j=1}^nd_j)\cdot\max_{j\in[n]}\{u_j\}}{K_w-n}
\end{aligned}\]
where the first inequality follows from \eqref{eqn:Constpiecewise}. Therefore, we conclude that
\[
\bV^{K_w}_{\bc}(\bd)\leq \sum_{j=1}^np_j\cdot s\cdot\int^1_{q=1-\bar{q}^{K_w}_j}F^{-1}_j(q)dq\leq \bVH_{\bc}(\bd)\leq \bV^{K_w}_{\bc}(\bd)+\frac{(\sum_{j=1}^nd_j)\cdot\max_{j\in[n]}\{u_j\}}{K_w-n}.
\]
Note that
\[
\sum_{j=1}^nd_j\cdot\int^1_{q=1-\bar{q}'_j}F^{-1}_j(q)dq=\lim_{w\rightarrow\infty}\sum_{j=1}^nd_j\cdot\int^1_{q=1-\bar{q}^{K_w}_j}F^{-1}_j(q)dq.
\]
We have
\[\begin{aligned}
\left|\sum_{j=1}^nd_j\cdot\int^1_{q=1-\bar{q}'_j}F^{-1}_j(q)dq-\bVH_{\bc}(\bd)\right|&=\lim_{w\rightarrow\infty}\left|\sum_{j=1}^nd_j\cdot\int^1_{q=1-\bar{q}^{K_w}_j}F^{-1}_j(q)dq-\bVH_{\bc}(\bd)\right|\\
&\leq \lim_{w\rightarrow\infty} \frac{(\sum_{j=1}^nd_j)\cdot\max_{j\in[n]}\{u_j\}}{K_w-n}=0
\end{aligned}\]
which implies that $\bar{\bm{q}}'$ is an optimal solution to $\bVH_{\bc}(\bd)$ \eqref{lp:newProphrelax}. Our proof is therefore completed.
\end{myproof}

\begin{myproof}[Proof of \Cref{lem:proof1}]
Under Case (i) when $\hat{q}^*_{j_t,t}\geq 1-2\kappa_1\cdot\sqrt{\frac{\log s}{s}}$,    
we know that 
\[
\bc\geq p_{j_t}\cdot s\cdot\hat{q}^*_{j_t,t}\cdot \ba_{j_t}\geq\frac{p_{j_t}\cdot s\cdot \ba_{j_t}}{2}\geq \ba_{j_t}
\]
when $s\geq s_0$ for a constant $s_0\geq 0$. Therefore, we always have enough remaining capacity to serve query $t$ with type $j_t$.

Since we have $q^{\pi}_{j_t,t}=1$, we only need to construct a feasible solution to $\bVH_{\bc-\ba_{j_t}}(\tilde{\bd}_{t+1})$ by noting that the term $\bVH_{\bc-\ba_{j_t}}(\tilde{\bd}_{t+1})$ contributes negatively to $\text{Myopic}_t(\pi,\bc,\mathcal{G})$ as shown in \eqref{eqn:231801}. 
From the feasibility of $\{\tilde{q}^*_{j,t}\}$, we know that
\begin{equation}\label{eqn:Feasibility}
    \sum_{j=1}^n \td_{j,t+1}\cdot a_{j,i}\cdot\tilde{q}^*_{j,t}+a_{j_t,i}\cdot\tilde{q}^*_{j_t,t}\leq c_{i},~~\forall i\in[m].
\end{equation}
Note that conditioning on the event $\mathcal{G}$, following \eqref{eqn:101201}, we have
\[
|\tilde{q}^*_{j_t,t}-\hat{q}^*_{j_t,t}|\leq \kappa_1\cdot \sqrt{\frac{\log(s-1)}{s-1}}
\]
which implies $\tilde{q}^*_{j_t,t}\geq 1-3\kappa_1\cdot\sqrt{\frac{\log s}{s}}\geq \frac{1}{2}$ when $s\geq s_0$ for a constant $s_0$ satisfying $\frac{s_0}{\log s_0}\geq36\kappa_1^2$.
We construct the following solution $\{\tilde{q}'_{j,t}\}$ for $\bVH_{\bc-\ba_{j_t}}(\tilde{\bd}_{t+1})$ satisfying
\begin{equation}\label{sol:feasible1}
\tilde{q}'_{j,t}=\tilde{q}^*_{j,t}, \forall j\neq j_t\text{~~and~~}\tilde{q}'_{j_t,t}=\tilde{q}^*_{j_t,t}+\frac{\tilde{q}^*_{j_t,t}-1}{\td_{j_t,t+1}}.
\end{equation}
Since $\sum_{j=1}^n \td_{j,t+1}\cdot a_{j,i}\cdot\tilde{q}'_{j,t}\leq c_i-a_{j_t,i}$ for each $i\in[m]$, we know that $\{\tilde{q}'_{j,t}\}$ is a feasible solution to $\bVH_{\bc-\ba_{j_t}}(\tilde{\bd}_{t+1})$, where $\tilde{q}'_{j_t,t}\geq0$ follows from $\tilde{q}^*_{j_t, t}\geq\frac{1}{2}$ and $\tilde{d}_{j_t, t+1}\geq1$ conditioning on the event $\mathcal{G}$.
Therefore, we have that
\begin{equation}\label{eqn:091801}
\begin{aligned}
\text{Myopic}_t(\pi,\bc,\mathcal{G})\leq&\mathbb{E}_{j_t, I_{t+1}}\left[ 
\sum_{j=1}^n \td_{j,t+1}\cdot \int_{q=1-\tilde{q}^*_{j,t}}^{1}F^{-1}_j(q)dq+\int_{q=1-\tilde{q}^*_{j_t,t}}^{ 1-{q}^{\pi}_{j_t,t} }F^{-1}_{j_t}(q)dq \right.\\
&\left. -\sum_{j=1}^n \td_{j,t+1}\cdot \int_{q=1-\tilde{q}'_{j,t}}^{1}F^{-1}_{j}(q)dq\mid\mathcal{G}
\right]\cdot P(\mathcal{G}) \\
=&\mathbb{E}_{j_t, I_{t+1}}\left[ 
\td_{j_t,t+1}\cdot \int_{q=1-\tilde{q}^*_{j_t,t}}^{1-\tilde{q}'_{j_t,t}}F^{-1}_{j_t}(q)dq+\int_{q=1-\tilde{q}^*_{j_t,t}}^{1-{q}^{\pi}_{j_t,t} }F^{-1}_{j_t}(q)dq\mid\mathcal{G}
\right]\cdot P(\mathcal{G}).
\end{aligned}
\end{equation}
We make the following claim.
\begin{claim}\label{claim:Rdeviation}
For any $q_1, q_2\in[0,1]$, it holds that
\[
\int_{q=q_1}^{q_2}F^{-1}_j(q)dq\leq F_j^{-1}(q_1)\cdot(q_2-q_1)+\frac{(q_2-q_1)^2}{\alpha}
\]
for any $j\in[n]$, where $\alpha$ is the lower bound of the density function $f(\cdot|\ba_j)$ specified in \Cref{assump:finitesize}.
\end{claim}
The proof of \Cref{claim:Rdeviation} is relegated to \Cref{sec:missingpf}. Therefore, applying \Cref{claim:Rdeviation}, we have
\begin{equation}\label{eqn:091802}
 \int_{q=1-\tilde{q}^*_{j_t,t}}^{1-\tilde{q}'_{j_t,t}}F^{-1}_{j_t}(q)dq\leq F^{-1}_{j_t}(1-\tilde{q}^*_{j_t,t})\cdot\frac{1-\tilde{q}^*_{j_t,t}}{\td_{j_t,t+1}}+\frac{(1-\tilde{q}^*_{j_t,t})^2}{\alpha\cdot \td^2_{j_t, t+1}}
\end{equation} 
and
\begin{equation}\label{eqn:091803}
\int_{q=1-\tilde{q}^*_{j_t,t}}^{ 1-{q}^{\pi}_{j_t,t} }F^{-1}_{j_t}(q)dq\leq F^{-1}_{j_t}(1-\tilde{q}^*_{j_t,t})\cdot(\tilde{q}^*_{j_t,t}-1)+  \frac{(1-\tilde{q}^*_{j_t,t})^2}{\alpha} 
\end{equation}
by noting that $q^{\pi}_{j_t, t}=1$.
Plugging \eqref{eqn:091802} and \eqref{eqn:091803} into \eqref{eqn:091801}, we have
\begin{equation}\label{eqn:new091804}
\begin{aligned}
\text{Myopic}_t(\pi,\bc,\mathcal{G})&\leq\mathbb{E}_{j_t, I_{t+1}}\left[ \frac{(1-\tilde{q}^*_{j_t,t})^2}{\alpha}+\frac{(1-\tilde{q}^*_{j_t,t})^2}{\alpha\cdot \td_{j_t,t+1}}\mid\mathcal{G} \right]\cdot P(\mathcal{G})\\
&\leq 2\mathbb{E}_{j_t, I_{t+1}}\left[\frac{(1-\hat{q}^*_{j_t,t})^2}{\alpha}+\frac{(\tilde{q}^*_{j_t,t}-\hat{q}^*_{j_t,t})^2}{\alpha}+\frac{1}{\alpha\cdot \td_{j_t, t+1}}\mid\mathcal{G}\right]\cdot P(\mathcal{G})\\
&\leq \frac{2\log s}{\alpha\cdot s}+\frac{2}{\alpha}\cdot\mathbb{E}_{j_t, I_{t+1}}[(\tilde{q}^*_{j_t,t}-\hat{q}^*_{j_t,t})^2]+\frac{2}{\alpha}\cdot\mathbb{E}_{j_t, I_{t+1}}\left[\frac{1}{\td_{j_t, I_{t+1}}}\mid\mathcal{G}\right]
\end{aligned}
\end{equation}
where the second inequality follows from $(1-\tilde{q}^*_{j_t,t})^2\leq 2\left((1-\hat{q}^*_{j_t,t})^2+(\tilde{q}^*_{j_t,t}-\hat{q}^*_{j_t,t})^2\right)$, and $(1-\tilde{q}^*_{j_t,t})^2\leq1$, and the third inequality follows from the condition that $\hat{q}^*_{j_t,t}\geq 1-2\kappa_1\cdot\sqrt{\frac{\log s}{s}}$. Therefore, our proof is completed.
\end{myproof}

\begin{myproof}[Proof of \Cref{claim:Rdeviation}]
Denote by functions
\[
G_j(q):=\int_{q'=q}^1 F^{-1}_j(q')dq',~~\forall j\in[n].
\]
It is clear that
\[
G_j'(q)=-F^{-1}_j(q)\text{~and~}G_j''(q)=-\frac{1}{f(q|\ba_j)}\in[-\frac{1}{\alpha},0].
\]
Therefore, from the concavity of $G_j(\cdot)$, we have
\[
\int_{q=q_1}^{q_2}F^{-1}_j(q)dq=G_j(q_1)-G_j(q_2)\leq F^{-1}_j(q_1)\cdot(q_2-q_1)+\frac{(q_2-q_1)^2}{\alpha}
\]
which completes our proof.
\end{myproof}

\begin{myproof}[Proof of \Cref{lem:proof2}]
In this case,
since we have ${q}^{\pi}_{j_t,t}=0$, we only need to construct feasible solution to $\bVH_{\bc}(\tilde{\bd}_{t+1})$. Note that conditioning on the event $\mathcal{G}$, following \eqref{eqn:101201}, we have
\[
|\tilde{q}^*_{j_t,t}-\hat{q}^*_{j_t,t}|\leq \kappa_1\cdot \sqrt{\frac{\log(s-1)}{s-1}}
\]
which implies $\tilde{q}^*_{j_t, t}\leq 3\kappa_1\cdot\sqrt{\frac{\log s}{s}}\leq\frac{1}{2}$ as long as $s\geq s_0$ for a constant $s_0$ satisfying $\frac{s_0}{\log s_0}\geq36\kappa_1^2$.
From the feasibility of $\{\tilde{q}^*_{j,t}\}$ demonstrated in \eqref{eqn:Feasibility}, we construct the following solution $\{\tilde{q}''_{j,t}\}$ for $\bVH_{\bc}(\tilde{\bd}_{t+1})$ satisfying
\begin{equation}\label{sol:feasible2}
\tilde{q}''_{j,t}=\tilde{q}^*_{j,t}, \forall j\neq j_t\text{~~and~~}\tilde{q}''_{j_t,t}=\tilde{q}^*_{j_t,t}\cdot\frac{\td_{j_t,t+1}+1}{\td_{j_t, t+1}}.
\end{equation}
Since $\sum_{j=1}^n \td_{j,t+1}\cdot a_{j,i}\cdot \tilde{q}''_{j,t}\leq c_i$ for each $i\in[m]$, we know that $\{\tilde{q}''_{j,t}\}$ is a feasible solution to $\bVH_{\bc}(\tilde{\bd}_{t+1})$, where $\tilde{q}''_{j_t, t}\leq1$ follows from $\tilde{q}^*_{j_t, t}\leq\frac{1}{2}$ and $\tilde{d}_{j_t, t+1}\geq1$ conditioning on the event $\mathcal{G}$.
Therefore, we have that
\begin{equation}\label{eqn:091805}
\begin{aligned}
&\text{Myopic}_t(\pi,\bc,\mathcal{G})\\
\leq&\mathbb{E}_{j_t, I_{t+1}}\left[ 
\sum_{j=1}^n \td_{j,t+1}\cdot \int_{q=1-\tilde{q}^*_{j,t}}^{1}F^{-1}_j(q)dq+\int_{q=1-\tilde{q}^*_{j_t,t}}^{ 1-{q}^{\pi}_{j_t,t} }F^{-1}_{j_t}(q)dq -\sum_{j=1}^n \td_{j,t+1}\cdot \int_{q=1-\tilde{q}''_{j,t}}^{1}F^{-1}_j(q)dq\mid\mathcal{G}
\right]\cdot P(\mathcal{G}) \\
=&\mathbb{E}_{j_t, I_{t+1}}\left[ 
\td_{j_t,t+1}\cdot \int_{q=1-\tilde{q}^*_{j_t,t}}^{1-\tilde{q}''_{j_t,t}}F^{-1}_{j_t}(q)dq+\int_{q=1-\tilde{q}^*_{j_t,t}}^{1-{q}^{\pi}_{j_t,t} }F^{-1}_{j_t}(q)dq\mid\mathcal{G}
\right]\cdot P(\mathcal{G}).
\end{aligned}
\end{equation}
Applying \Cref{claim:Rdeviation}, we have that
\begin{equation}\label{eqn:091806}
 \int_{q=1-\tilde{q}^*_{j_t,t}}^{1-\tilde{q}''_{j_t,t}}F^{-1}_{j_t}(q)dq\leq- F^{-1}_{j_t}(1-\tilde{q}^*_{j_t,t})\cdot\frac{\tilde{q}^*_{j_t,t}}{\td_{j_t,t+1}}+\frac{(\tilde{q}^*_{j_t,t})^2}{\alpha\cdot \td^2_{j_t, t+1}}
\end{equation} 
and
\begin{equation}\label{eqn:091807}
\int_{q=1-\tilde{q}^*_{j_t,t}}^{1-{q}^{\pi}_{j_t,t} }F^{-1}_{j_t}(q)dq\leq F^{-1}_{j_t}(1-\tilde{q}^*_{j_t,t})\cdot\tilde{q}^*_{j_t,t}+  \frac{(\tilde{q}^*_{j_t,t})^2}{\alpha}. 
\end{equation}
Plugging \eqref{eqn:091806} and \eqref{eqn:091807} into \eqref{eqn:091805}, we get
\begin{equation}\label{eqn:new091808}
\begin{aligned}
\text{Myopic}_t(\pi,\bc,\mathcal{G})&\leq\mathbb{E}_{j_t, I_{t+1}}\left[ \frac{(\tilde{q}^*_{j_t,t})^2}{\alpha}+\frac{(\tilde{q}^*_{j_t,t})^2}{\alpha\cdot \td_{j_t,t+1}}\mid\mathcal{G} \right]\cdot P(\mathcal{G})\\
&\leq 2\mathbb{E}_{j_t, I_{t+1}}\left[
\frac{(\hat{q}^*_{j_t,t})^2}{\alpha}+\frac{(\hat{q}^*_{j_t,t}-\tilde{q}^*_{j_t,t})^2}{\alpha}+\frac{1}{\alpha\cdot \td_{j_t, t+1}}\mid\mathcal{G} \right]\cdot P(\mathcal{G})\\
&\leq \frac{2\log s}{\alpha\cdot s}+\frac{2}{\alpha}\cdot\mathbb{E}_{j_t, I_{t+1}}[(\tilde{q}^*_{j_t,t}-\hat{q}^*_{j_t,t})^2]+\frac{2}{\alpha}\cdot\mathbb{E}_{j_t, I_{t+1}}\left[\frac{1}{\td_{j_t, t+1}}\mid\mathcal{G}\right]
\end{aligned}\end{equation}
where the second inequality follows from $(\tilde{q}^*_{j_t,t})^2\leq2\left( (\hat{q}^*_{j_t,t})^2+(\hat{q}^*_{j_t,t}-\tilde{q}^*_{j_t,t})^2 \right)$ and $(\tilde{q}^*_{j_t,t})^2\leq1$, and the third inequality follows from the condition that $\hat{q}^*_{j_t,t}\leq2\kappa_1\cdot\sqrt{\frac{\log s}{s}}$. Our proof is thus completed.
\end{myproof}

\begin{myproof}[Proof of \Cref{lem:proof3}]  
In Case (iii) when $2\kappa_1\cdot\sqrt{\frac{\log s}{s}}\leq \hat{q}^*_{j_t,t}\leq 1-2\kappa_1\cdot\sqrt{\frac{\log s}{s}}$, we know that 
\[
\bc\geq p_{j_t}\cdot s\cdot\hat{q}^*_{j_t,t}\cdot \ba_{j_t}\geq p_{j_t}\cdot 2\kappa_1\cdot\sqrt{s\log s}\cdot \ba_{j_t}\geq \ba_{j_t}
\]
when $s\geq s_0$ for a constant $s_0$ such that $s_0\log s_0\geq \frac{1}{4\kappa_1^2}\cdot\max_{j\in[n]}\{\frac{1}{p_j^2}\}$. Therefore, we always have enough remaining capacity to serve query $t$ with type $j_t$.

Note that conditioning on the event $\mathcal{G}$, following \eqref{eqn:101201}, we have
\[
|\tilde{q}^*_{j_t,t}-\hat{q}^*_{j_t,t}|\leq \kappa_1\cdot \sqrt{\frac{\log(s-1)}{s-1}}
\]
which implies $\kappa_1\cdot\sqrt{\frac{\log s}{s}}\leq \tilde{q}^*_{j_t,t}\leq1-\kappa_1\cdot\sqrt{\frac{\log s}{s}}$.

We construct feasible solution $\{\tilde{q}'_{j,t}\}$ for $\bVH_{\bc-\ba_{j_t}}(\tilde{\bd}_{t+1})$ following the definition in \eqref{sol:feasible1}. Then, $\tilde{q}'_{j_t,t}\geq0$ follows from the fact that 
\[
\tilde{q}^*_{j_t,t}\geq \kappa_1\cdot\sqrt{\frac{\log s}{s}}\geq \frac{2}{p_{j_t}\cdot(s-1)}\geq \frac{1}{\td_{j_t, t+1}}
\]
conditioning on the event $\mathcal{G}$, as long as $s\geq s_0$ for a constant $s_0$ satisfying $s_0\log s_0\geq \frac{4}{\kappa_1^2}\cdot\max_{j\in[n]}\{\frac{1}{p_j^2}\}$ and $\frac{s_0}{\log s_0}\geq\max_{j\in[n]}\{\frac{4}{p_j^2}\}$. 

We construct a feasible solution $\{\tilde{q}''_{j,t}\}$ for $\bVH_{\bc}(\tilde{\bd}_{t+1})$ following the definition in \eqref{sol:feasible2}. Then, $\tilde{q}''_{j_t,t}\leq1$ follows from
\[
\tilde{q}''_{j_t,t}\leq \tilde{q}^*_{j_t,t}+\frac{1}{\td_{j_t,t+1}}\leq 1-\kappa_1\cdot\sqrt{\frac{\log s}{s}}+\frac{3}{2p_{j_t}(s-1)}\leq1
\]
conditioning on the event $\mathcal{G}$, as long as $s\geq s_0$ for a constant $s_0$ such that $s_0\log s_0\geq \frac{9}{4\kappa_1^2}\cdot\max_{j\in[n]}\{\frac{1}{p_j^2}\}$.

Therefore, we have that
\begin{equation}\label{eqn:091809}
\begin{aligned}
\text{Myopic}_t(\pi,\bc,\mathcal{G})\leq&\mathbb{E}_{j_t, I_{t+1}}\left[ 
\sum_{j=1}^n \td_{j,t+1}\cdot \int_{q=1-\tilde{q}^*_{j,t}}^{1}F^{-1}_j(q)dq+\int_{q=1-\tilde{q}^*_{j_t,t}}^{ 1-{q}^{\pi}_{j_t,t} }F^{-1}_{j_t}(q)dq \right.\\
&\left. -q^{\pi}_{j_t,t}\cdot\sum_{j=1}^n \td_{j,t+1}\cdot \int_{q=1-\tilde{q}'_{j,t}}^{1}n F^{-1}_{j}(q)dq-(1-q^{\pi}_{j_t,t})\cdot\sum_{j=1}^n \td_{j,t+1}\cdot \int_{q=1-\tilde{q}''_{j,t}}^{1}F^{-1}_j(q)dq
\mid\mathcal{G}\right]\cdot P(\mathcal{G}) \\
=&\mathbb{E}_{j_t, I_{t+1}}\left[ 
q^{\pi}_{j_t,t}\cdot \td_{j_t,t+1}\cdot \int_{q=1-\tilde{q}^*_{j_t,t}}^{1-\tilde{q}'_{j_t,t}}F^{-1}_{j_t}(q)dq+(1-q^{\pi}_{j_t,t})\cdot \td_{j_t,t+1}\cdot \int_{q=1-\tilde{q}^*_{j_t,t}}^{1-\tilde{q}''_{j_t,t}}F^{-1}_{j_t}(q)dq  \right.\\
&\left.+\int_{q=1-\tilde{q}^*_{j_t,t}}^{1-{q}^{\pi}_{j_t,t} }F^{-1}_{j_t}(q)dq\mid\mathcal{G}\right]\cdot P(\mathcal{G}).
\end{aligned}
\end{equation}
Applying \Cref{claim:Rdeviation}, we get \eqref{eqn:091802}, \eqref{eqn:091806}, and
\begin{equation}\label{eqn:091810}
   \int_{q=1-\tilde{q}^*_{j_t,t}}^{1-{q}^{\pi}_{j_t,t} }F^{-1}_{j_t}(q)dq=F^{-1}_{j_t}(1-\tilde{q}^*_{j_t,t})\cdot(\tilde{q}^*_{j_t,t}-q^{\pi}_{j_t,t})+\frac{(\tilde{q}^*_{j_t,t}-q^{\pi}_{j_t,t})^2}{\alpha}.
\end{equation}
Therefore, we have
\begin{equation}\label{eqn:new091811}
\begin{aligned}
\text{Myopic}_t(\pi,\bc,\mathcal{G})&\leq\mathbb{E}_{j_t, I_{t+1}}\left[
q^{\pi}_{j_t,t}\cdot\frac{(1-\tilde{q}^*_{j_t,t})^2}{\alpha\cdot \td_{j_t,t+1}}+(1-q^{\pi}_{j_t,t})\cdot\frac{(\tilde{q}^*_{j_t,t})^2}{\alpha\cdot \td_{j_t, t+1}}+\frac{(\tilde{q}^*_{j_t,t}-q^{\pi}_{j_t,t})^2}{\alpha}\mid\mathcal{G}\right]\cdot P(\mathcal{G})\\
&\leq\mathbb{E}_{j_t, I_{t+1}}\left[ \frac{1}{\alpha\cdot d_{j_t, t+1}}+\frac{(\tilde{q}^*_{j_t,t}-\hat{q}^*_{j_t,t})^2}{\alpha}\mid\mathcal{G} \right]\cdot P(\mathcal{G}).
\end{aligned}
\end{equation}
Our proof is thus completed.
\end{myproof}

\section{Missing Proofs for \Cref{sec:Logsection}}\label{sec:missingpf2}

\begin{myproof}[Proof of \Cref{lem:Example}]
We first show that it is optimal to restrict the range of the dual variable $\tbmu$ of the relaxed offline optimum $\bV^{\text{Off}}_{\bc}(I_t)$ \eqref{lp:relaxoffline} into the region $\Omega=\{ \bmu\geq0: \mu_1+\mu_2+\mu_3\leq 2 \}$.

The dual problem of $\bV^{\text{Off}}_{\bc}(I_t)$ \eqref{lp:relaxoffline} can be written as follows
\[
\min_{\bmu\geq0} \sum_{i\in[m]}\mu_i\cdot c_i+\sum_{\tau=t}^T \max_{x_{\tau}\in[0,1]}\left\{\tr_{\tau}\cdot x_{\tau}-\sum_{i\in[m]} \mu_i\cdot \ta_{\tau, i}\cdot x_{\tau}  \right\}.
\]
We prove that any optimal solution to the above dual problem must belong to the region $\Omega=\{ \bmu\geq0: \mu_1+\mu_2+\mu_3\leq 2 \}$ by showing contradiction. Suppose that there is a dual variable $\hat{\bmu}$ optimal to the dual problem above and it holds that $\hat{\mu}_1+\hat{\mu}_2+\hat{\mu}_3> 2$. Clearly, all type 4 queries will be rejected. We classify it into two scenarios. 

\noindent Scenario 1: suppose that there exists a type $i=1,2,3$, say type $i=1$, that will not be reject, i.e., $\hat{\mu}_2+\hat{\mu}_3\leq 1$. Then, we know that $\hat{\mu}_1>1$, which implies that type $2$ and type $3$ queries will all be rejected. Note that $\hat{\mu}_2+\hat{\mu}_3\leq 1$ implies $\hat{\mu}_1-1\geq\hat{\mu}_1+\hat{\mu}_2+\hat{\mu}_3-2$.
We can obtain a new dual variable that $\hat{\mu}_2'=\hat{\mu}_2$, $\hat{\mu}'_3=\hat{\mu}_3$, and $\hat{\mu}'_1=\hat{\mu}_1-\delta$, for a positive constant $\delta= \hat{\mu}_1+\hat{\mu}_2+\hat{\mu}_3- 2$ satisfying $\delta\leq \hat{\mu}_1-1$. It is clear to see that $\hat{\mu}'_1+\hat{\mu}'_2+\hat{\mu}'_3\geq 2$ and $\hat{\mu}'_1\geq1$, and thus the primal solution and $\hat{\bmu}'$ still satisfy the saddle-point condition and thus $\hat{\bmu}'$ is optimal. 
Therefore, we know that it is optimal to select $\hat{\bmu}'$ as the optimal dual variable and it holds that $\hat{\bmu}'\in\Omega$.

\noindent Scenario 2: suppose that it holds $\hat{\mu}_1+\hat{\mu}_2>1$, $\hat{\mu}_2+\hat{\mu}_3>1$, and $\hat{\mu}_1+\hat{\mu}_3>1$. Then we set $\delta'=\min\{ \hat{\mu}_1+\hat{\mu}_2, \hat{\mu}_1+\hat{\mu}_3, \hat{\mu}_2+\hat{\mu}_3, \frac{\hat{\mu}_1+\hat{\mu}_2+\hat{\mu}_3}{2} \}$ and we obtain a new dual variable $\hat{\bmu}'=\delta'\cdot\bmu$. It is clear to see that $\hat{\bmu}'$ satisfies $\hat{\mu}'_1+\hat{\mu}'_2\geq1$, $\hat{\mu}'_2+\hat{\mu}'_3\geq1$, $\hat{\mu}'_1+\hat{\mu}'_3\geq1$ and $\hat{\mu}_1'+\hat{\mu}_2'+\hat{\mu}_3'\geq2$. Thus the primal solution and $\hat{\bmu}'$ still satisfy the saddle-point condition and thus $\hat{\bmu}'$ is optimal. Now $\hat{\bmu}'$ either satisfies the condition in scenario 1 or $\hat{\bmu}'\in\Omega$. In either case, we can finally show that it is optimal to restrict the range of the dual variable to the region $\Omega=\{ \bmu\geq0: \mu_1+\mu_2+\mu_3\leq 2 \}$.

We now show that \Cref{assump:contiknown2} is satisfied. For any $\bmu', \bmu''\in\Omega$, we know that 
\[
F(\ba_4^\top\bmu'|\ba_4)-F(\ba_4^\top\bmu''|\ba_4)=\frac{1}{2}\cdot (\ba_4^\top\bmu'-\ba_4^\top\bmu'').
\]
Therefore, it holds that 
\[
\mathbb{E}_{\tba}\left[ \left(F(\tba^\top\bmu'|\tba)-F(\tba^\top\bmu''|\tba)\right)\cdot (\tba^\top\bmu'-\tba^\top\bmu'')\right]\geq \frac{1}{2}\cdot (\ba_4^\top\bmu'-\ba_4^\top\bmu'')^2\geq 
\frac{p_4}{2}\cdot\mathbb{E}_{\tba}\left[(\tba^\top\bmu'-\tba^\top\bmu'')^2\right]
\]
where the last inequality holds by noting that
\[
(\ba_4^\top\bmu'-\ba_4^\top\bmu'')^2=\|\bmu'-\bmu''\|_2^2\geq (\ba_i^\top\bmu'-\ba_i^\top\bmu'')^2
\]
for all $i=1,2,3$. On the other hand, it is clear to see that 
\[
\mathbb{E}_{\tba}\left[ \left(F(\tba^\top\bmu'|\tba)-F(\tba^\top\bmu''|\tba)\right)\cdot (\tba^\top\bmu'-\tba^\top\bmu'')\right]\leq \mathbb{E}_{\tba}\left[(\tba^\top\bmu'-\tba^\top\bmu'')^2\right]
\]
since the densities are all upper bound by 1. Therefore, we know that \Cref{assump:contiknown2} is satisfied. 

We finally show that the strict complementary slackness condition is violated by \eqref{lp:Example2}. Denote by $\mu_i$ the dual variable for constraint for resource $i$. Then, it is clear to see that the primal-dual pair $\mu_1^*=\mu^*_3=\frac{1-\eps}{2}$, $\mu^*_2=0$ and $q^*_1=q^*_3=q^*_4=\frac{1+\eps}{2}$, $q^*_2=\eps$ is optimal to \eqref{lp:Example2} by checking that the saddle point condition is satisfied, for any $\eps>0$. To be specific, it holds that
\[\begin{aligned}
&\mu^*_1\cdot \left( p_2q^*_2+p_3q^*_3+p_4q^*_4- \frac{5\eps(1+\eps)}{2(1+6\eps)} \right)=0\\
&\mu^*_2\cdot \left(p_1q^*_1+p_3q^*_3+p_4q^*_4- \frac{5\eps(1+\eps)}{2(1+6\eps)} \right) =0\\
&\mu^*_3\cdot \left(p_1q^*_1+ p_2q^*_2+p_4q^*_4- \frac{5\eps(1+\eps)}{2(1+6\eps)} \right)=0
\end{aligned}\]
and
\[\begin{aligned}
&q^*_1=1-\mu^*_2-\mu^*_3, ~~~q^*_2=1-\mu^*_1-\mu^*_3\\
&q^*_3=1-\mu^*_1-\mu^*_2, ~~~q^*_4=\frac{1}{2}\cdot(2-\mu^*_1-\mu^*_2-\mu^*_3).
\end{aligned}\]
Therefore, the saddle point conditions are satisfied and the primal-dual pair $\mu_1^*=\mu^*_3=\frac{1-\eps}{2}$, $\mu^*_2=0$ and $q^*_1=q^*_3=q^*_4=\frac{1+\eps}{2}$, $q^*_2=\eps$ is optimal to \eqref{lp:Example2}.
However, while the resource constraint is binding for every $i=1,2,3$, the optimal dual variable $\mu^*_2=0$, which shows that the strict complementary slackness condition is not satisfied for every $\eps>0$. Our proof is thus completed.
\end{myproof}

\begin{myproof}[Proof of \Cref{lem:decompose}]
By plugging \eqref{eqn:roundViterate} into \eqref{def:myopicregret}, we have that
\begin{align}
\text{Myopic}_t(\pi,\tbc^{\pi}_t)&=\mathbb{E}_{(\tr_t,\tba_t)}\left[ \mathbb{E}_{I_{t+1}}[ \bVL_{\tbc^{\pi}_t}(I_{t})]-\bVL_{\tbc^{\pi}_t-\tba_t\cdot \tx^{\pi}_t}(I_{t+1})]]-\tr_t\cdot \tx^{\pi}_t \right] \label{eqn:derivation01} \\
&=\mathbb{E}_{(\tr_t,\tba_t)}\left[ \mathbb{E}_{I_{t+1}}[(\tx^{\text{round}}_t-\tx^{\pi}_t)\cdot(\tr_t+\bVL_{\tbc_t^{\pi}-\tba_t}(I_{t+1}))+ (\tx^{\pi}_t-\tx^{\text{round}}_t)\cdot \bVL_{\tbc_t^{\pi}}(I_{t+1})+G_{\tbc^{\pi}_t}(I_t)]\nonumber \right]\\
&=\mathbb{E}_{I_{t+1}}\left[\mathbb{E}_{(\tr_t,\tba_t)}[(\tx^{\text{round}}_t-\tx^{\pi}_t)\cdot(\tr_t-M_{\tbc_t^{\pi},\tba_t}(I_{t+1}))]\right]+\mathbb{E}[G_{\tbc_t^{\pi}}(I_t)].   \nonumber
\end{align}
From the definition of $\tx^{\text{round}}_t$ and $\tx^{\pi}_t$, we know that $\tx^{\text{round}}_t-\tx^{\pi}_t\in\{-1,1\}$ if and only if $\hat{M}_{\tbc^{\pi}_t,\tba_t}\leq \tr_t\leq M_{\tbc^{\pi}_t,\tba_t}(I_{t+1})$ and $\tbc^{\pi}_t\geq \tba_t$, or $M_{\tbc^{\pi}_t,\tba_t}(I_{t+1})\leq \tr_t\leq \hat{M}_{\tbc^{\pi}_t,\tba_t}$ and $\tbc^{\pi}_t\geq \tba_t$. Thus, we have that 
\begin{align}
&\mathbb{E}_{I_{t+1}}\left[\mathbb{E}_{(\tr_t,\tba_t)}[(\tx^{\text{round}}_t-\tx^{\pi}_t)\cdot(\tr_t-M_{\tbc_t^{\pi},\tba_t}(I_{t+1}))]\right] \label{eqn:derivation02}\\
\leq&\mathbb{E}_{I_{t+1}}\left[\mathbb{E}_{\tba_t}\left[\mathbb{E}_{\tr_t\sim \Ftat}[\bI_{\{\hat{M}_{\tbc^{\pi}_t,\tba_t}\leq \tr_t\leq M_{\tbc^{\pi}_t,\tba_t}(I_{t+1})\}}\cdot(M_{\tbc^{\pi}_t,\tba_t}(I_{t+1})-\hat{M}_{\tbc^{\pi}_t,\tba_t})] \right]\right] \cdot\bI_{\{\tbc^{\pi}_t\geq \tba_t\}}  \nonumber\\
&+\mathbb{E}_{I_{t+1}}\left[\mathbb{E}_{\tba_t}\left[\mathbb{E}_{\tr_t\sim \Ftat}[\bI_{\{\hat{M}_{\tbc^{\pi}_t,\tba_t}\geq \tr_t\geq M_{\tbc^{\pi}_t,\tba_t}(I_{t+1})\}}\cdot(\hat{M}_{\tbc^{\pi}_t,\tba_t}-M_{\tbc^{\pi}_t,\tba_t}(I_{t+1}))] \right]\right] \cdot\bI_{\{\tbc^{\pi}_t\geq \tba_t\}} \nonumber\\
\leq& \bar{\alpha}\cdot \mathbb{E}_{I_{t+1}}\left[\mathbb{E}_{\tba_t}\left[(\hat{M}_{\tbc^{\pi}_t,\tba_t}-M_{\tbc^{\pi}_t,\tba_t}(I_{t+1}))^2\right]\right] \cdot\bI_{\{\tbc^{\pi}_t\geq \tba_t\}} \nonumber\\
\leq& 2\bar{\alpha}\cdot\bI_{\{\tbc^{\pi}_t\geq \tba_t\}}\cdot\mathbb{E}_{\tba_t}\left[(\hat{M}_{\tbc^{\pi}_t,\tba_t}-\mathbb{E}_{I_{t+1}}[M_{\tbc^{\pi}_t,\tba_t}(I_{t+1})])^2\right]\nonumber\\
&+2\bar{\alpha}\cdot\bI_{\{\tbc^{\pi}_t\geq \tba_t\}}\cdot\mathbb{E}_{\tba_t}\left[\mathbb{E}_{I_{t+1}}\left[(\mathbb{E}_{I_{t+1}}[M_{\tbc^{\pi}_t,\tba_t}(I_{t+1})]-M_{\tbc^{\pi}_t,\tba_t}(I_{t+1}))^2\right]\right]\nonumber
\end{align}
where the second inequality follows from \Cref{assump:contiknown2} which implies that the density of $\tr_t$ is upper bounded by $\bar{\alpha}$ and the third inequality follows from $(a+b)^2\leq 2a^2+2b^2$ for any $a, b$. Plugging \eqref{eqn:derivation02} into \eqref{eqn:derivation01}, we complete our proof of the lemma.
\end{myproof}

\begin{myproof}[Proof of \Cref{lem:secondorder}]
Note that we have
\[
\LF_{\bc,t+1}(\bmu)=\frac{\bc^\top\bmu}{s-1}+\mathbb{E}_{\tba}\left[\mathbb{E}_{\tr\sim F_{\tba}}[\tr-\tba^\top\bmu]^+ |\tba\right]=\frac{\bc^\top\bmu}{s-1}+\mathbb{E}_{\tba}\left[\int_{\tba^\top\bmu}^{\infty} (\tr-\tba^\top\bmu)f(r|\tba)dr \right]
\]
which implies that
\[
\frac{\partial\LF_{\bc,t+1}(\bmu)}{\partial \mu_i}=\frac{c_i}{s-1}-\mathbb{E}_{\tba}[\tilde{a}_i\cdot(1-F(\tba^\top\bmu|\tba))],~~\forall i\in[m]
\]
and
\[
\frac{\partial^2\LF_{\bc,t+1}(\bmu)}{\partial \mu_i\partial \mu_{i'}}=\mathbb{E}_{\tba}[f(\tba^\top\bmu|\tba)\cdot \tilde{a}_i\cdot \tilde{a}_{i'}], ~~\forall i, i'\in[m]
\]
where we use $\tilde{a}_i$ to denote the $i$-th element of vector $\tba$.
Moreover, note that for any $\bmu\geq0$, from Taylor's theorem, there exists a $\bmu'$ that lies on the intersection between $\bmu$ and $\bmu^*$ such that
\[
\LF_{\bc,t+1}(\bmu)-\LF_{\bc}(\mu^*)=(\bmu-\bmu^*)^\top\frac{\partial\LF_{\bc,t+1}(\bmu^*)}{\partial \bmu}+\frac{1}{2}\cdot (\bmu-\bmu^*)^\top\cdot \frac{\partial^2 \LF_{\bc,t+1}(\bmu')}{\partial \bmu^2}\cdot(\bmu-\bmu^*)
\]
where $\frac{\partial\LF_{\bc,t+1}(\bmu^*)}{\partial \bmu}=(\frac{\partial\LF_{\bc,t+1}(\bmu^*)}{\partial \mu_1},\ldots, \frac{\partial\LF_{\bc,t+1}(\bmu^*)}{\partial \mu_m})$, and $\frac{\partial^2 \LF_{\bc,t+1}(\bmu')}{\partial \bmu^2}$ is the Hessian matrix that equals $\mathbb{E}_{\tba}[f(\tba^\top\bmu'|\tba)\cdot \tba\cdot \tba^\top]$. Note that $\Omega$ is assumed to be a convex set in \Cref{assump:contiknown2} and both $\bmu, \bmu^*\in\Omega$. Then, we have $\bmu'\in\Omega$ and \Cref{assump:contiknown2} implies that $\mathbb{E}_{\tba}[f(\tba^\top\bmu'|\tba)]\geq\underline{\alpha}$. The smallest positive eigenvalue of $\frac{\partial^2 \LF_{\bc,t+1}(\bmu')}{\partial \bmu^2}$ is therefore lower bounded by $\underline{\alpha}\cdot\underline{\beta}$. Also, from the optimality of $\bmu^*$, we must have
\[
(\bmu-\bmu^*)^\top\frac{\partial\LF_{\bc,t+1}(\bmu^*)}{\partial \bmu}\geq0
\]
for any $\bmu\geq0$. Thus, it holds that
\[
\frac{\underline{\alpha}\underline{\beta}}{2}\cdot\|\mathcal{P}_S(\bmu-\bmu^*)\|_2^2\leq \frac{1}{2}\cdot (\bmu-\bmu^*)^\top\cdot \frac{\partial^2 \LF_{\bc,t+1}(\bmu')}{\partial \bmu^2}\cdot(\bmu-\bmu^*)\leq \LF_{\bc,t+1}(\bmu)-\LF_{\bc,t+1}(\bmu^*)
\]
which completes our proof.
\end{myproof}

\begin{myproof}[Proof of \Cref{lem:Decomposevariation}]
From \eqref{eqn:dualboundvariance}, we know that 
\[\begin{aligned}
&M_{\bc,\ba_t}(I_{t+1})-\mathbb{E}[M_{\bc,\ba_t}(I_{t+1})]\leq \ba^\top_t\tbmu_2-\mathbb{E}[\ba^\top_t\tbmu_1], &&\text{if~}M_{\bc,\ba_t}(I_{t+1})\geq\mathbb{E}[M_{\bc,\ba_t}(I_{t+1})]\\
&\mathbb{E}[M_{\bc,\ba_t}(I_{t+1})]-M_{\bc,\ba_t}(I_{t+1})\leq\mathbb{E}[\ba^\top_t\tbmu_2]-\ba^\top_t\tbmu_1, &&\text{if~}M_{\bc,\ba_t}(I_{t+1})\leq\mathbb{E}[M_{\bc,\ba_t}(I_{t+1})]
\end{aligned}\]
which implies that
\[
\Var(M_{\bc,\ba_t}(I_{t+1}))\leq 2\Var(\ba_t^\top\tbmu_1)+2\Var(\ba_t^\top\tbmu_2)+2(\mathbb{E}[\ba_t^\top\tbmu_1]-\mathbb{E}[\ba_t^\top\tbmu_2])^2.
\]
Note that we have
\[
\Var(\ba_t^\top\tbmu_1)=\mathbb{E}[(\ba_t^\top\tbmu_1-\ba_t^\top\hbmu_1+\ba_t^\top\hbmu_1-\ba_t^\top\mathbb{E}[\tbmu_1])^2]\leq 2\mathbb{E}[(\ba_t^\top\tbmu_1-\ba_t^\top\hbmu_1)^2]+2(\ba_t^\top\hbmu_1-\ba_t^\top\mathbb{E}[\tbmu_1])^2.
\]
Similarly, we have
\[
\Var(\ba_t^\top\tbmu_2)=\mathbb{E}[(\ba_t^\top\tbmu_2-\ba_t^\top\hbmu_2+\ba_t^\top\hbmu_2-\ba_t^\top\mathbb{E}[\tbmu_2])^2]\leq 2\mathbb{E}[(\ba_t^\top\tbmu_2-\ba_t^\top\hbmu_2)^2]+2(\ba_t^\top\hbmu_2-\ba_t^\top\mathbb{E}[\tbmu_2])^2.
\]
Also, we have
\[\begin{aligned}
(\mathbb{E}[\ba_t^\top\tbmu_2]-\mathbb{E}[\ba_t^\top\tbmu_1])^2&=(\ba_t^\top\mathbb{E}[\tbmu_2]-\ba_t^\top\hbmu_2+\ba_t^\top\hbmu_2-\ba_t^\top\hbmu_1+\ba_t^\top\hbmu_1-\ba_t^\top\mathbb{E}[\tbmu_1])^2\\
&\leq 3(\ba_t^\top\mathbb{E}[\tbmu_2]-\ba_t^\top\hbmu_2)^2+3(\ba_t^\top\hbmu_2-\ba_t^\top\hbmu_1)^2+3(\ba_t^\top\hbmu_1-\ba_t^\top\mathbb{E}[\tbmu_1])^2.
\end{aligned}\]
Thus, we have that
\begin{equation}\label{eqn:100901}
\begin{aligned}
\Var(M_{\bc,\ba_t}(I_{t+1}))&\leq 10(\ba_t^\top\mathbb{E}[\tbmu_1]-\ba_t^\top\hbmu_1)^2+10(\ba_t^\top\mathbb{E}[\tbmu_2]-\ba_t^\top\hbmu_2)^2+6(\ba_t^\top\hbmu_2-\ba_t^\top\hbmu_1)^2\\
&~~~+4\mathbb{E}[(\ba_t^\top\tbmu_1-\ba_t^\top\hbmu_1)^2]+4\mathbb{E}[(\ba_t^\top\tbmu_2-\ba_t^\top\hbmu_2)^2]\\
&\leq 14\mathbb{E}[(\ba_t^\top\tbmu_1-\ba_t^\top\hbmu_1)^2]+14\mathbb{E}[(\ba_t^\top\tbmu_2-\ba_t^\top\hbmu_2)^2]+6(\ba_t^\top\hbmu_2-\ba_t^\top\hbmu_1)^2.
\end{aligned}
\end{equation}
It only remains to bound the term $(\ba_t^\top\hbmu_2-\ba_t^\top\hbmu_1)^2$. We have the following claim, where the proof is relegated to the end of this proof.
\begin{claim}\label{claim:100801}
It holds that
\[
(\ba_t^\top(\hbmu_1-\hbmu_2))^2\leq \frac{2}{(s-1)\underline{\alpha}\underline{\beta}}\cdot\bar{d}^{3}\cdot m^{1/2}\cdot\gamma
\]
where $\bar{d}=\max_{j\in[n]}\{\|\ba_j\|_2\}$ and $\gamma=\max_{i\in[m], j\in[n]:a_{j,i}>0}\frac{u_j}{a_{j,i}}$.
\end{claim}
Therefore, our proof is completed by combining \eqref{eqn:100901} and \Cref{claim:100801}.
\end{myproof}

\begin{myproof}[Proof of \Cref{claim:100801}]
From \Cref{lem:secondorder}, by substituting $\hbmu_1$ into $\bmu^*$, and substituting $\hbmu_2$ into $\bmu$, we have that
\[\begin{aligned}
\frac{\underline{\alpha}\underline{\beta}}{2}\cdot\|\mathcal{P}_S(\hbmu_2-\hbmu_1)\|_2^2&\leq \LF_{\bc,t+1}(\hbmu_2)-\LF_{\bc,t+1}(\hbmu_1)=\LF_{\bc,t+1}(\hbmu_2)-\LF_{\bc-\ba_t,t+1}(\hbmu_2)+\LF_{\bc-\ba_t,t+1}(\hbmu_2)-\LF_{\bc,t+1}(\hbmu_1)\\
&\leq \LF_{\bc,t+1}(\hbmu_2)-\LF_{\bc-\ba_t,t+1}(\hbmu_2)=\frac{1}{s-1}\cdot \ba_t^\top\hbmu_2
\end{aligned}\]
where the second inequality follows from $\LF_{\bc-\ba_t,t+1}(\hbmu_2)\leq \LF_{\bc-\ba_t,t+1}(\hbmu_1)\leq \LF_{\bc,t+1}(\hbmu_1)$ by noting that $\hbmu_2\in\argmin_{\bmu\geq0}\LF_{\bc-\ba_t,t+1}(\bmu)$.
Thus, we have
\[
(\ba_t^\top(\hbmu_2-\hbmu_1))^2\leq \bar{d}^2\cdot \|\mathcal{P}_S(\hbmu_2-\hbmu_1)\|^2_2\leq \frac{2}{(s-1)\underline{\alpha}\underline{\beta}}\cdot (\ba_t^\top\hbmu_2)\cdot\bar{d}^2\leq \frac{2}{(s-1)\underline{\alpha}\underline{\beta}}\cdot \|\ba_t\|_2\cdot\|\hbmu_2\|_2\cdot\bar{d}^2.
\]
Clearly, we must have $\hat{\mu}_{2,i}\leq\gamma$ for each $i\in[m]$ with $\gamma=\max_{i\in[m], j\in[n]:a_{j,i}>0}\frac{u_j}{a_{j,i}}$, which implies that $\|\hbmu_2\|_2\leq \sqrt{m}\gamma$. Thus, it holds that
\[
(\ba_t^\top(\hbmu_1-\hbmu_2))^2\leq \frac{2}{(s-1)\underline{\alpha}\underline{\beta}}\cdot\bar{d}^{3}\cdot m^{1/2}\cdot\gamma
\]
which completes our proof.
\end{myproof}

\begin{myproof}[Proof of \Cref{lem:firstvariation}]
We first consider the setting where $\bc\geq\tba_t$ and thus both $\tbmu_1$ and $\tbmu_2$ are well-defined in \eqref{eqn:dualsolution}.
We show that $\tx^*_t\neq\tx^{\text{round}}_t$ happens only if $\ba^\top_t\tbmu_1\leq \tr_t\leq \tba^\top_t\tbmu_2$. Note that the value of $\tx^*_t$ can be determined in the following way:
\begin{equation}\label{pf:lem701}
    \tx^*_t=\argmax_{\phi\in[0,1]}\phi\cdot \tr_t+\bVL_{\bc-\phi\cdot \tba_t}(I_{t+1})-\bVL_{\bc}(I_{t+1}).
\end{equation}
Note that for any $\phi\in[0,1]$, we must have
\[
-\phi\cdot \tba^\top_t\tbmu_2\leq\bVL_{\bc-\phi\cdot \tba_t}(I_{t+1})-\bVL_{\bc}(I_{t+1})\leq-\phi\cdot \tba^\top_t\tbmu_1
\]
which can be proved following the same intuition of \Cref{lem:dualbound}.
Therefore, when $\tr_t\leq \tba^\top_t\tbmu_1$, we must have $\tx^*_t=\tx^{\text{round}}_t=0$, and when $\tr_t\geq \tba^\top_t\tbmu_2$, we must have $\tx^*_t=\tx^{\text{round}}_t=1$. We conclude that $G_{\bc}(I_t)=0$ when $\tr_t\leq \tba^\top_t\tbmu_1$ or $\tr_t\geq \tba^\top_t\tbmu_2$. We now assume that $\tba^\top_t\tbmu_1\leq \tr_t\leq \tba^\top_t\tbmu_2$ and we further consider two cases as follows:

\noindent Case 1: If $\tx^{\text{round}}_t=0$, then we have
\[
G_{\bc}(I_t)=\tr_t\cdot \tx^*_t+\bVL_{\bc-\tba_t\cdot \tx^*_t}(I_{t+1})-\bVL_{\bc}(I_{t+1}).
\]
Note that we have 
\[
\bVL_{\bc-\tba_t\cdot \tx^*_t}(I_{t+1})-\bVL_{\bc}(I_{t+1})\leq-\tx^*_t\cdot \tba^\top_t\tbmu_1.
\]
By noting $\tr_t\leq \tba^\top_t\tbmu_2$ and $\tx^*_t\leq1$, we have that
\[
G_{\bc}(I_t)\leq \tx^*_t\cdot (\tba^\top_t\tbmu_2-\tba^\top_t\tbmu_1)\leq \tba^\top_t\tbmu_2-\tba^\top_t\tbmu_1
\]
which gives us an upper bound for the first case.

\noindent Case 2: If $\tx^{\text{round}}_t=1$, then we have
\[
G_{\bc}(I_t)=\tr_t\cdot (\tx^*_t-1)+\bVL_{\bc-\tba_t\cdot \tx^*_t}(I_{t+1})-\bVL_{\bc-\tba_t}(I_{t+1}).
\]
Note that we have
\[
\bVL_{\bc-\tba_t\cdot \tx^*_t}(I_{t+1})-\bVL_{\bc-\tba_t}(I_{t+1})\leq(1-\tx^*_t)\cdot \tba^\top_t\tbmu_2.
\]
By noting $\tr_t\geq \tba^\top_t\tbmu_1$ and $0\leq1-\tx^*_t\leq1$, we have that
\[
G_{\bc}(I_t)\leq (1-\tx^*_t)\cdot (\tba^\top_t\tbmu_2-\tba^\top_t\tbmu_1)\leq \tba^\top_t\tbmu_2-\tba^\top_t\tbmu_1 
\]
which gives us an upper bound for the second case. Therefore, on both cases, we show that if $\tba^\top_t\tbmu_1\leq \tr_t\leq \tba^\top_t\tbmu_2$, it holds that
\begin{equation}\label{eqn:100801}
G_{\bc}(I_t)\leq \tba^\top_t\tbmu_2-\tba^\top_t\tbmu_1.
\end{equation}
which implies that
\[\begin{aligned}
\mathbb{E}_{\tba_t}\left[\mathbb{E}_{I_{t+1}}\left[\mathbb{E}_{\tr_t\sim \Ftat}[G_{\bc}(I_t)]\right]\right]
&\leq \mathbb{E}_{\tba_t}\left[\mathbb{E}_{I_{t+1}}\left[\mathbb{E}_{\tr_t\sim \Ftat}[ \bI_{\tba^\top_t\tbmu_1\leq \tr_t\leq \tba^\top_t\tbmu_2}\cdot(\tba^\top_t\tbmu_2-\tba^\top_t\tbmu_1) ]\right]\right]\\
&\leq\bar{\alpha}\cdot\mathbb{E}_{\tba_t}\left[ \mathbb{E}_{I_{t+1}}\left[(\tba^\top_t\tbmu_2-\tba^\top_t\tbmu_1)^2 \right]\right].
\end{aligned}\]
We further note that
\[\begin{aligned}
(\tba^\top_t\tbmu_2-\tba^\top_t\tbmu_1)^2&=(\tba^\top_t\tbmu_2-\tba^\top_t\hbmu_2+\tba^\top_t\hbmu_2-\tba^\top_t\hbmu_1+\tba^\top_t\hbmu_1-\tba^\top_t\tbmu_1)^2\\
&\leq 3(\tba^\top_t\tbmu_2-\tba^\top_t\hbmu_2)^2+3(\tba^\top_t\hbmu_2-\tba^\top_t\hbmu_1)^2+3(\tba^\top_t\hbmu_1-\tba^\top_t\tbmu_1)^2.
\end{aligned}\]
We use \Cref{claim:100801} to bound the term $(\tba^\top_t\hbmu_2-\tba^\top_t\hbmu_1)^2$. Then, we have 
\begin{equation}\label{eqn:100803}
\mathbb{E}_{I_{t}}\left[G_{\bc}(I_t)\right]\leq \frac{6\bar{\alpha}}{(s-1)\underline{\alpha}\underline{\beta}}\bar{d}^{3} m^{1/2}\gamma+3\bar{\alpha}\cdot \mathbb{E}_{\tba_t}\left[\mathbb{E}_{I_{t+1}}[(\tba^\top_t\tbmu_1-\tba^\top_t\hbmu_1)^2]\right]+3\bar{\alpha}\cdot\mathbb{E}_{\tba_t}\left[ \mathbb{E}_{I_{t+1}}[(\tba^\top_t\tbmu_2-\tba^\top_t\hbmu_2)^2]\right]
\end{equation}
where $\bar{d}=\max_{j\in[n]}\{\|\ba_j\|_2\}$ and $\gamma=\max_{i\in[m], j\in[n]:a_{j,i}>0}\frac{u_j}{a_{j,i}}$.

We then consider the setting where $\bc\geq\tba_t$ does not hold and only $\tbmu_1$ is well-defined in \eqref{eqn:dualsolution}. Still, when $\tr_t\leq \tba^\top_t\tbmu_1$, we must have $\tx^*_t=\tx^{\text{round}}_t=0$, which implies $G_{\bc}(I)=0$. We now assume that $\tr_t\geq\tba_t^\top\tbmu_1$. Then, since
\[
-\phi\cdot \tba^\top_t\tbmu_2\leq\bVL_{\bc-\phi\cdot \tba_t}(I_{t+1})-\bVL_{\bc}(I_{t+1})\leq-\phi\cdot \tba^\top_t\tbmu_1
\]
for any $\phi\in[0,1]$, we have
\[
G_{\bc}(I_t)\leq \tx^*_t\cdot(\tr_t-\tba_t^\top\tbmu_1)\leq \tr_t-\tba_t^\top\tbmu_1
\]
by noting $\tx^*_t\in[0,1]$. Denote by $j_t$ as the type of query $t$. It holds that 
\[
\mathbb{E}_{\tr_{t}\sim\Ftat}[G_{\bc}(I_t)]\leq \frac{1}{2}\cdot(u_{j_t}-\ba_{j_t}^\top\tbmu_1)^2\leq (u_{j_t}-\ba_{j_t}^\top\hbmu_1)^2+(\ba_{j_t}^\top\hbmu_1-\ba_{j_t}^\top\tbmu_1)^2\leq u_{j_t}\cdot(u_{j_t}-\ba_{j_t}^\top\hbmu_1)+(\ba_{j_t}^\top\hbmu_1-\ba_{j_t}^\top\tbmu_1)^2.
\]
It is clear to see that $\min_{\mu\in\Omega} \LF_{\bc,t+1}(\mu)$ is the dual problem of  $\max_{\bm{x}} \bVF_{t+1,\bc}$. Denote by $\bar{\bm{x}}=\{\bar{x}_j(r)\}\in\text{argmax}_{\bm{x}} \bVF_{t+1,\bc}$ such that $\bar{\bm{x}}$ and $\hbmu_1$ is an optimal primal-dual pair.
Then, when $c_i< a_{j_t, i}$ for some $i\in[m]$, we must have
\[
\mathbb{E}_{r\sim F(\cdot|\ba_{j_t})}[\bar{x}_{j_t}(r)]\leq \frac{1}{p_{j_t}\cdot(s-1)}
\]
and as a result we have
\[
\alpha\cdot(u_{j_t}-\ba_{j_t}^\top\tbmu_1)\leq P(\ba_{j_t}^\top\tbmu_1\leq r\leq u_{j_t}|r\sim F_{j_t})=\mathbb{E}_{r\sim F(\cdot|\ba_{j_t})}[\bar{x}_{j_t}(r)]\leq \frac{1}{p_{j_t}\cdot(s-1)}
\]
where $\alpha>0$ is a lower bound on the density function specified in \Cref{assump:finitesize}. Then we have
\[
    \mathbb{E}_{I_{t+1}}\left[\mathbb{E}_{\tr_t\sim\Ftat}[G_{\bc}(I_t)]\right]\leq \frac{u_{j_t}}{\alpha\cdot p_{j_t}\cdot(s-1)}+\mathbb{E}_{I_{t+1}}[(\tba^\top_t\tbmu_1-\tba^\top_t\hbmu_1)^2]. 
\]
Therefore, we have
\begin{equation}\label{eqn:100807}
\mathbb{E}_{I_t}[G_{\bc}(I_t)]\leq \frac{u_{j_t}}{\alpha\cdot p_{j_t}\cdot(s-1)}+ \mathbb{E}_{\tba_t}\left[\mathbb{E}_{I_{t+1}}[(\tba^\top_t\tbmu_1-\tba^\top_t\hbmu_1)^2]\right].
\end{equation}
Our proof is completed by combining \eqref{eqn:100803} and \eqref{eqn:100807}.
\end{myproof}

\begin{myproof}[Proof of \Cref{lem:boundIandII}]
From \eqref{def:subspace}, $\mathcal{S}$ refers to the space spanned by $\mathcal{A}$, which is a subspace of $\mathbb{R}^m$. We further denote by $\mathcal{S}^{\bot}$ the orthogonal complement subspace of $S$ in the Euclidean space $\mathbb{R}^m$, i.e., $\mathbb{R}^m=\mathcal{S}+\mathcal{S}^{\bot}$ and $\bm{u}^\top\bm{v}=0$ for any $\bm{u}\in\mathcal{S}$ and any $\bm{v}\in\mathcal{S}^{\bot}$. 

Following basics in linear algebra, any vector $\bmu\in\mathbb{R}^m$ can be decomposed uniquely as $\bmu=\bmu_S+\bmu_{S^{\bot}}$, where $\bmu_S\in \mathcal{S}$ and $\bmu_{S^{\bot}}\in \mathcal{S}^{\bot}$. Then the projection of $\bmu$ to the subspace $\mathcal{S}$ can be given as $\mathcal{P}_S(\bmu)=\bmu_S$. Following this decomposition, for any $\bc\geq0$, we have that
\[\begin{aligned}
\min_{\bmu\geq\Omega}\LF_{\bc,t+1}(\bmu)&=\min_{\bmu_S+\bmu_{S^{\bot}}\in\Omega}\left(\frac{\bc}{T-t}\right)^\top\bmu_S+\left(\frac{\bc}{T-t}\right)^\top\bmu_{S^{\bot}}+\mathbb{E}_{(\tr,\tba)}[\tr-\tba^\top\bmu_S]^+\\
&=\min_{\bmu_S\in \mathcal{S}}\left(\frac{\bc}{T-t}\right)^\top\bmu_S+h(\bmu_S)+\mathbb{E}_{(\tr,\tba)}[\tr-\tba^\top\bmu_S]^+
\end{aligned}\]
where we note that $\ba^\top\bmu_{S^{\bot}}=0$ for any $\ba\in\mathcal{A}$ and
\[
h(\bmu_S)=\min_{\bmu_{S^{\bot}}\in S^{\bot}}\left(\frac{\bc}{T-t}\right)^\top\bmu_{S^{\bot}}\text{~s.t.~}\bmu_{S^{\bot}}+\bmu_S\in\Omega.
\]
Note that if the optimization problem that defines $h(\bmu_S)$ is infeasible for some $\bmu_S$, then we simply set $h(\bmu_S)=+\infty$. Denote by $\hat{\mathcal{S}}\subset\mathcal{S}\cap\Omega$ the convex set such that $h(\bmu)$ is finite for all $\bmu\in\hat{\mathcal{S}}$. Therefore, the dual problem $\min_{\bmu\in\Omega}\LF_{\bc,t+1}(\bmu)$ can be transferred into a minimization problem over the set $\hat{\mathcal{S}}$, i.e.,
\[
\min_{\bmu\in\Omega}\LF_{\bc,t+1}(\bmu)=\min_{\bmu\in \hat{\mathcal{S}}} \hat{G}_{\bc}(\bmu):=\left(\frac{\bm{c}}{T-t}\right)^\top\bmu+h(\bmu)+\mathbb{E}_{(\tr,\tba)}[\tr-\tba^\top\bmu]^+.
\]
Following the same way, the minimization of the sample average dual function $\LL_{\bc, I_{t+1}}(\bmu)$ in \eqref{eqn:Lagrangian} can be formulated as
\[
\min_{\bmu\in\Omega}\LL_{\bc,I_{t+1}}(\bmu)=\min_{\bmu\in \hat{\mathcal{S}}} \hat{G}_{\bc, I_{t+1}}(\bmu):=\left(\frac{\bm{c}}{T-t}\right)^\top\bmu+h(\bmu)+\frac{1}{T-t}\cdot\sum_{\tau=t+1}^T[\tr_{\tau}-\tba^\top_{\tau}\bmu]^+.
\]
We denote by
\begin{equation}\label{eqn:optimalG}
    \bmu^*(\bm{c}')\in \argmin_{\mu\in \hat{\mathcal{S}}} \hat{G}_{\bc'}(\bmu)\text{~and~}\tbmu(\bm{c}')\in\argmin_{\bmu\in \hat{\mathcal{S}}}\hat{G}_{\bc',I_{t+1}}(\bmu)
\end{equation}
for any $\bc'\geq0$, 
where $\tbmu(\bm{c}')$ is a random variable whose value depends on $I_{t+1}$. Clearly, we have that
\begin{equation}\label{eqn:pflem01}
\mu^*(\bm{c})=\mathcal{P}_S(\hbmu_1), ~\text{and}
~\tbmu(\bm{c})=\mathcal{P}_S(\tbmu_1),\text{~and~}\tbmu(\bm{c}-\ba_t)=\mathcal{P}_S(\tbmu_2).
\end{equation}
Thus, in order to bound $\mathbb{E}[(\ba_t^\top\tbmu_1-\ba_t^\top\hbmu_1)^2]$, it is sufficient to bound
\begin{equation}\label{eqn:pflem02}
    \mathbb{E}[(\bmu^*(\bm{c}')-\tbmu(\bm{c}'))^2]
\end{equation}
for any $\bm{c}'\geq0$, where the $\mathbb{E}[\cdot]$ is taken over $I_{t+1}$. In what follows, we will consider bounding \eqref{eqn:pflem02}. Since we will assume that $\bm{c}'$ is now fixed, for notation simplicity, we will drop $\bm{c}'$ in the expression of $\bmu^*(\bm{c}')$, $\tbmu(\bm{c}')$, $\hat{G}_{\bc'}(\bmu)$ and $\hat{G}_{\bc',I_{t+1}}(\bmu)$. We simply denote $\bmu^*$, $\tbmu$, $\hat{G}(\bmu)$ and $\hat{G}_{I_{t+1}}(\bmu)$.

We first note that the function $h(\bmu)$ is a convex function over $\bmu\in\hat{\mathcal{S}}$. We also note that the function $\hat{G}$ is simply a re-formulation of the function $\hat{L}$ after projecting the decision variable onto the subspace $\mathcal{S}$. Thus, \Cref{lem:secondorder} implies the following second-order growth condition for the function $\hat{G}$, i.e.,
\begin{equation}\label{eqn:Gsecondorder}
\hat{G}(\bmu)-\hat{G}(\bmu^*)\geq \frac{\underline{\alpha}\underline{\beta}}{2}\cdot\|\bmu-\bmu^*\|_2^2\text{~for~any~}\bmu\in \hat{\mathcal{S}}\subset\mathcal{S}
\end{equation}
which implies that $\bmu^*$ is unique. Thus, from Lemma 2.1 in \cite{shapiro1992perturbation}, we know that
\[
\|\tbmu-\bmu^*\|_2\leq\frac{2}{\underline{\alpha}\underline{\beta}}\cdot\sup_{\bmu\in \hat{\mathcal{S}}\cap B(\bmu^*, r_0)}\left\{ \frac{ \hat{G}_{I_{t+1}}(\bmu)-\hat{G}(\bmu)-\hat{G}_{I_{t+1}}(\bmu^*)+\hat{G}(\bmu^*) }{\|\bmu-\bmu^*\|_2} \right\}
\]
where $r_0=\|\tbmu-\bmu^*\|_2$ and $B(\bmu^*, r_0)=\{\bmu:\|\bmu-\bmu^*\|_2\leq r_0\}$. Therefore, in order to further bound $\|\tbmu-\bmu^*\|_2$, it is sufficient to bound the term
\[
\sup_{\bmu\in \hat{\mathcal{S}}\cap B(\bmu^*, r_0)}\left\{ \frac{ \frac{1}{T-t}\sum_{\tau=t+1}^T[\tr_{\tau}-\tba^\top_{\tau}\bmu]^+-\mathbb{E}_{(\tr,\tba)}[\tr-\tba^\top\bmu]^+-\frac{1}{T-t}\sum_{\tau=t+1}^T[\tr_{\tau}-\tba^\top_{\tau}\bmu^*]^++\mathbb{E}_{(\tr,\tba)}[\tr-\tba^\top\bmu^*]^+ }{\|\bmu-\bmu^*\|_2} \right\}.
\]
We denote by the function
\[
g(\bmu, (r,\ba)):=[r-\ba^\top\bmu]^+ +\bar{\alpha}\bar{d}^2\cdot\|\bmu-\bmu^*\|_2^2.
\]
Moreover, we denote by
\begin{equation}\label{def:deltafunction}
\tdelta(\bmu)=\frac{1}{T-t}\cdot\sum_{\tau=t+1}^Tg(\bmu, (\tr_{\tau}, \tba_{\tau}))-\mathbb{E}_{(\tr,\tba)}[ g(\bmu, (\tr,\tba))]
\end{equation}
which is a random variable that depends on $I_{t+1}$.
Then, the above upper bound over $\|\tbmu-\bmu^*\|_2$ can be re-formulated as
\[
\|\tbmu-\bmu^*\|_2\leq \frac{2}{\underline{\alpha}\underline{\beta}}\cdot \sup_{\bmu\in \hat{\mathcal{S}}\cap B(\bmu^*, r_0)}\left\{ \frac{\tdelta(\bmu)-\tdelta(\bmu^*)}{\|\bmu-\bmu^*\|_2} \right\}.
\]
We denote by $\tilde{E}$ the set of $\bmu$ such that $\tdelta(\bmu)$ is non-differentiable. For simplicity, we denote by $\hat{B}(\bmu^*, r_0)=\hat{\mathcal{S}}\cap B(\bmu^*, r_0)$. Note that $\tdelta(\bmu)$ is Liptschitz continuous.  Then, from the mean value theorem for Liptschitz function (\cite{clarke1990optimization} p41), we have that
\begin{equation}\label{eqn:53001}
\|\tbmu-\bmu^*\|_2^2\leq\frac{4}{\underline{\alpha}^2\underline{\beta}^2}\cdot \sup_{\bmu\in \hat{B}(\bmu^*, r_0)\setminus E}\{\|\nabla \tdelta(\bmu)\|_2^2\}=\frac{4}{\underline{\alpha}^2\underline{\beta}^2}\cdot(\|\nabla \tdelta(\bmu^*)\|_2^2+\sup_{\bmu\in \hat{B}(\bmu^*, r_0)\setminus \tilde{E}}\{\|\nabla \tdelta(\bmu)-\nabla \tdelta(\bmu^*)\|_2^2\}).
\end{equation}
We have the following claim over the term $\|\nabla \tdelta(\bmu^*)\|_2^2$, which is proved at the end of this proof.
\begin{claim}\label{claim:bounddelta1}
It holds that
\[
\mathbb{E}[\|\nabla \tdelta(\bmu^*)\|_2^2]\leq\frac{\bar{d}^2}{T-t}
\]
where the expectation $\mathbb{E}[\cdot]$ takes over $I_{t+1}=\{(\tr_{t+1}, \tba_{t+1}),\dots,(\tr_T,\tba_T)\}$ that defines $\tdelta(\cdot)$ in \eqref{def:deltafunction}.
\end{claim}
Regarding the term $\sup_{\bmu\in \hat{B}(\bmu^*, r_0)\setminus \tilde{E}}\{\|\nabla \tdelta(\bmu)-\nabla \tdelta(\bmu^*)\|_2^2\}$, following the steps in \cite{shapiro1993asymptotic}, we can show that
\[
\sup_{\bmu\in B(\mu^*, r_0)\setminus \tilde{E}}\{\frac{\|\nabla \tdelta(\bmu)-\nabla \tdelta(\bmu^*)\|_2}{\frac{1}{\sqrt{T-t}}+r_0}\}\rightarrow0\text{~in~probability~as~}T-t\rightarrow\infty.
\]
As a result, there exists a constant $t_0$ such that as long as $T-t\geq t_0$, we have that 
\[
\sup_{\bmu\in \hat{B}(\bmu^*, r_0)\setminus \tilde{E}}\{\|\nabla \tdelta(\bmu)-\nabla \tdelta(\bmu^*)\|_2^2\}\leq\frac{1}{2}\cdot(\frac{1}{T-t}+r_0^2)
\]
with a high probability. We formalize the above step in the following claim for completeness, by making explicit the constant term $t_0$, as well as the ``high probability''.
\begin{claim}\label{claim:bounddelta2}
It holds that 
\begin{equation}\label{eqn:newfinalterm}
\begin{aligned}
P\left(\sup_{\|\bmu-\bmu^*\|_2\leq r_0}\frac{\|\nabla\tdelta(\bmu)-\nabla\tdelta(\bmu^*)\|_2}{\frac{1}{\sqrt{T-t}}+3\bar{\alpha}\bar{d}^2\cdot r_0}\geq\frac{\underline{\alpha}\underline{\beta}}{12\bar{\alpha}\bar{d}^2}\right)\leq &2\bar{\alpha}\bar{d}^3\left(\frac{12\bar{\alpha}\bar{d}^2}{\underline{\alpha}\underline{\beta}}\right)^2 (T-t)^{-\frac{13}{25}} \\
&+2\exp\left(-\delta\cdot (T-t)^{12/25}\right) \left(\frac{13(\log (T-t)+\log 2\gamma\sqrt{m})}{25|\log(1-q)|}+1\right) (2M)^m
\end{aligned}
\end{equation}
as long as $T-t\geq \hat{t}$, where $q=1/M=\underline{\alpha}\underline{\beta}/(144\bar{\alpha}^2\bar{d}^2)$, and $\hat{t}, \delta$ are two constants that is independent of $m$ and are determined polynomially by the problem parameters $\underline{\alpha}, \underline{\beta}, \bar{\alpha}, \bar{d}$.
\end{claim}
The proof of \Cref{claim:bounddelta2} is relegated to the end of this proof. Denote by 
\[
\tilde{z}=\sup_{\|\bmu-\bmu^*\|_2\leq r_0}\frac{\|\nabla\tdelta(\bmu)-\nabla\tdelta(\bmu^*)\|_2}{\frac{1}{\sqrt{T-t}}+3\bar{\alpha}\bar{d}^2\cdot r_0}.
\]
We also denote by three events
\[
\mathcal{E}_1=\{ \tilde{z}\leq \frac{\underline{\alpha}\underline{\beta}}{12\bar{\alpha}\bar{d}^2} \}\text{~~and~~}\mathcal{E}_2=\{ \frac{\underline{\alpha}\underline{\beta}}{12\bar{\alpha}\bar{d}^2}<\tilde{z}\text{~and~}r_0\leq\frac{1}{(T-t)^{\frac{7}{25}}} \}~\text{~~and~~}\mathcal{E}_3=\{\frac{\underline{\alpha}\underline{\beta}}{12\bar{\alpha}\bar{d}^2}<\tilde{z}\text{~and~}r_0>\frac{1}{(T-t)^{\frac{7}{25}}}\}.
\]
Clearly, we have
\begin{equation}\label{eqn:61507}
\mathbb{E}[r_0^2]=P(\mathcal{E}_1)\cdot\mathbb{E}[r_0^2|\mathcal{E}_1]+P(\mathcal{E}_2)\cdot\mathbb{E}[r_0^2|\mathcal{E}_2]+P(\mathcal{E}_2)\cdot\mathbb{E}[r_0^2|\mathcal{E}_2].
\end{equation}
Now, from \eqref{eqn:53001}, \Cref{claim:bounddelta1}, and the definition of the event $\mathcal{E}_1$, we have
\begin{equation}\label{eqn:61502}
\begin{aligned}
P(\mathcal{E}_1)\cdot\mathbb{E}[r_0^2|\mathcal{E}_1]&\leq \frac{4}{\underline{\alpha}^2\underline{\beta}^2}\cdot P(\mathcal{E}_1)\cdot \mathbb{E}[\|\nabla \tdelta(\bmu^*)\|_2^2|\mathcal{E}_1]+P(\mathcal{E}_1)\cdot \frac{1}{36\bar{\alpha}^2\bar{d}^4}\cdot\mathbb{E}\left[\left(\frac{1}{\sqrt{T-t}}+3\bar{\alpha}\bar{d}^2\cdot r_0\right)^2\mid\mathcal{E}_1\right]\\
&\leq \frac{4}{\underline{\alpha}^2\underline{\beta}^2}\cdot\mathbb{E}[\|\nabla \tdelta(\bmu^*)\|_2^2]+\frac{1}{2}\cdot P(\mathcal{E}_1)\cdot \mathbb{E}[r_0^2|\mathcal{E}_1]+\frac{1}{18\bar{\alpha}^2\bar{d}^4\cdot(T-t)}
\end{aligned}
\end{equation}
which implies that
\begin{equation}\label{eqn:61503}
P(\mathcal{E}_1)\cdot\mathbb{E}[r_0^2|\mathcal{E}_1]\leq \frac{8\bar{d}^2}{\underline{\alpha}^2\underline{\beta}^2\cdot(T-t)}+\frac{1}{9\bar{\alpha}\bar{d}^2\cdot(T-t)}.
\end{equation}
For the term $P(\mathcal{E}_2)\cdot\mathbb{E}[r_0^2|\mathcal{E}_2]$, from the definition of the event $\mathcal{E}_2$ and \Cref{claim:bounddelta2}, we have
\begin{equation}\label{eqn:61504}
\begin{aligned}
    P(\mathcal{E}_2)\cdot\mathbb{E}[r_0^2|\mathcal{E}_2]\leq P(\mathcal{E}_2)\cdot\frac{1}{(T-t)^{\frac{14}{25}}}\leq &2\bar{\alpha}\bar{d}^3\left(\frac{12\bar{\alpha}\bar{d}^2}{\underline{\alpha}\underline{\beta}}\right)^2 (T-t)^{-\frac{27}{25}}\\
    &+2\exp\left(-\delta\cdot (T-t)^{12/25}\right) \left(\frac{13(\log (T-t)+\log 2\gamma\sqrt{m})}{25|\log(1-q)|}+1\right) (2M)^m\\
    &\leq \frac{1}{T-t}
\end{aligned}
\end{equation}
as long as $T-t>t_1$, where 
\[
t_1=O\left( \left(mM\cdot\frac{\bar{\alpha}^3\bar{d}^7}{\underline{\alpha}^2\underline{\beta}^2}\right)^{\frac{25}{2}} \right).
\]


We now want to bound the probability that event $\mathcal{E}_3$ happens. Note that we have
\begin{equation}\label{eqn:61005}
  \frac{\underline{\alpha}\underline{\beta}}{2}\cdot\|\tbmu-\bmu^*\|_2^2\leq \frac{1}{T-t}\sum_{\tau=t+1}^T[\tr_{\tau}-\tba^\top_{\tau}\tbmu]^+-\mathbb{E}_{(\tr,\tba)}[\tr-\tba^\top\tbmu]^+-\frac{1}{T-t}\sum_{\tau=t+1}^T[\tr_{\tau}-\tba^\top_{\tau}\bmu^*]^++\mathbb{E}_{(\tr,\tba)}[\tr-\tba^\top\bmu^*]^+.  
\end{equation}
By modifying the approach in \citet{li2021online}, we have the following lemma, which follows Proposition 2 and Proposition 3 of \citet{li2021online} and is proved at the end of this proof.
\begin{lemma}\label{lem:61501}
It holds that 
\[
P\left( \|\tbmu-\bmu^*\|_2^2\geq k_1/(\underline{\alpha}^2\underline{\beta}^2)\cdot\eps^2 \right)\leq k_2\cdot \left(\frac{2}{q}\right)^m\cdot \log\left(\frac{1}{\eps}\right)\cdot \exp\left(-\frac{k_3\eps^2(T-t)}{\bar{d}^2} \right)
\]
where $k_1$, $k_2$, $k_3$ are constant numbers independent of problem parameters and $q$ is a constant such that $\bar{\alpha}\cdot\bar{d}^2\cdot\frac{q(1+q)}{(1-q)^2}\leq \frac{\underline{\alpha}\underline{\beta}}{32}$.
\end{lemma}
By setting $\epsilon=\frac{\underline{\alpha}\underline{\beta}}{k_1\cdot(T-t)^{\frac{7}{25}}}$ in \Cref{lem:61501}, we have that
\begin{equation}\label{eqn:61506}
P(\mathcal{E}_3)\leq P\left( \|\tbmu-\bmu^*\|^2_2\geq\left(\frac{1}{(T-t)^{\frac{7}{25}}}\right)^2 \right)\leq O\left((2/q)^m\right)\cdot\log(T-t)\cdot\exp(-(T-t)^{\frac{11}{25}})\leq\frac{1}{T-t}
\end{equation}
as long as $T-t>t_2$ where $t_2=O(\frac{m}{q})$.
Therefore, combining \eqref{eqn:61507}, \eqref{eqn:61503}, \eqref{eqn:61504}, and \eqref{eqn:61506}, we have
\[
\mathbb{E}[\|\tbmu-\bmu^*\|_2^2]=\mathbb{E}[r_0^2]\leq \frac{8\bar{d}^2}{\underline{\alpha}^2\underline{\beta}^2\cdot(T-t)}+\frac{1}{9\bar{\alpha}\bar{d}^2\cdot(T-t)}+ \frac{2}{T-t}
\]
as long as $T-t>t_0$, where 
\[
t_0=O\left( \left(mM\cdot\frac{\bar{\alpha}^3\bar{d}^7}{\underline{\alpha}^2\underline{\beta}^2}\right)^{\frac{25}{2}}\cdot\frac{1}{q} \right).
\]
Therefore, our proof is completed.
\end{myproof}
\begin{myproof}[Proof of \Cref{claim:bounddelta1}]
Note that
\[
\nabla \tdelta(\bmu^*)=-\frac{1}{T-t}\cdot\sum_{\tau=t+1}^T\tba_{\tau} \cdot\bI_{\{\tr_{\tau}\geq \tba_{\tau}^\top\bmu^*\}}+\mathbb{E}_{\tba}[\tba\cdot(1-F(\tba^\top\bmu^*|\tba))].
\]
Clearly, for each $\tau=t+1,\ldots,T$, we have
\[
\mathbb{E}_{(\tr_\tau, \tba_\tau)}[\tba_{\tau} \cdot\bI_{\{\tr_{\tau}\geq \tba_{\tau}^\top\bmu^*\}}]=\mathbb{E}_{\tba}[\tba\cdot(1-F(\tba^\top\bmu^*|\tba))].
\]
Thus, we have
\[
\mathbb{E}[\|\nabla \tdelta(\bmu^*)\|_2^2]=\frac{1}{T-t}\cdot \Var(\tba_{\tau} \cdot\bI_{\{\tr_{\tau}\geq \tba_{\tau}^\top\bmu^*\}})\leq\frac{\bar{d}^2}{T-t}
\]
which completes our proof.
\end{myproof}

\begin{myproof}[Proof of \Cref{claim:bounddelta2}]
We denote by
\[
\psi((r,\ba), \bmu)=\nabla g(\bmu, (r,\ba))-\nabla g(\bmu^*, (r,\ba)).
\]
We also denote by
\[
\lambda(\bmu)=\mathbb{E}_{(r,\ba)}[\psi((r,\ba),\bmu)]
\]
and
\[
u((r,\ba),\bmu,d)=\sup_{\|\bmu-\bmu^*\|_2\leq d}\|\psi((r,\ba), \bmu)-\psi((r,\ba), \bmu^*)\|_2.
\]
Clearly, we have that
\begin{equation}\label{eqn:N2}
\lambda(\bmu^*)=0.
\end{equation}
Thus, condition (N-2) in \cite{huber1967under} is satisfied. Also, for any $\bmu$, from definition of the functions $\psi$ and $\lambda$, we have that
\[
\|\lambda(\bmu)\|_2=\|2\bar{\alpha}\bar{d}^2(\bmu-\bmu^*)+\mathbb{E}_{\tba}[\tba\cdot (F(\tba^\top\bmu|\tba)-F(\tba^\top\bmu^*|\tba))]\|_2.
\]
Note that
\[
F(\tba^\top\bmu|\tba)-F(\tba^\top\bmu^*|\tba)\leq\bar{\alpha}\cdot|\tba^\top\bmu-\tba^\top\bmu^*|\leq\bar{\alpha}\bar{d}\cdot\|\bmu-\bmu^*\|_2.
\]
Thus, we have
\begin{equation}\label{eqn:N31}
\begin{aligned}
\|\lambda(\bmu)\|_2&\geq 2\bar{\alpha}\bar{d}^2\cdot\|\bmu-\bmu^*\|_2-\|\mathbb{E}_{\tba}[\tba\cdot (F(\tba^\top\bmu|\tba)-F(\tba^\top\bmu|\tba))]\|_2\\
&\geq 2\bar{\alpha}\bar{d}^2\cdot\|\bmu-\bmu^*\|_2-\bar{\alpha}\bar{d}^2\cdot\|\bmu-\bmu^*\|_2=\bar{\alpha}\bar{d}^2\cdot\|\bmu-\bmu^*\|_2
\end{aligned}
\end{equation}
for any $d_0\geq0$ and $\|\bmu-\bmu^*\|_2\leq d_0$.
Thus, we verify that condition (N-3) (i) in \cite{huber1967under} holds. Moreover, we have
\begin{equation}\label{eqn:52904}
\|\psi((r,\ba), \bmu)-\psi((r,\ba), \bmu^*)\|_2=\|\ba\cdot \bI_{\{r\geq \ba^\top\bmu^*\}}-\ba\cdot \bI_{\{r\geq \ba^\top\bmu\}}\|\leq \|\ba\cdot (\bI_{\{ \ba^\top\bmu\leq r\leq \ba^\top\bmu^* \}}+\bI_{\{ \ba^\top\bmu^*\leq r\leq \ba^\top\bmu \}})\|\leq\bar{d}
\end{equation}
which implies that
\begin{equation}\label{eqn:52107}
\mathbb{E}_{(\tr,\tba)}[u((\tr,\tba),\bmu,d)]\leq \mathbb{E}_{\tba}[\|\tba\|_2\cdot 2\bar{\alpha}\bar{d}d]\leq 2\bar{\alpha}\bar{d}^2\cdot d.
\end{equation}
Similarly, we have
\[
\mathbb{E}_{(\tr,\tba)}[u((\tr,\tba),\bmu,d)^2]\leq \mathbb{E}_{\tba}[\|\tba\|_2^2\cdot 2\bar{\alpha}\bar{d}d]\leq 2\bar{\alpha}\bar{d}^3\cdot d.
\]
Thus, we verify conditions (N-3) (ii) and (iii) in \cite{huber1967under} for any $d_0\geq0$. Clearly, $\mathbb{E}_{(\tr,\tba)}[\psi((\tr,\tba),\bmu^*)]$ is finite, which verifies condition (N-4) in \cite{huber1967under}. Thus, denote by
\[
Z_{T-t}(\bmu,\bmu^*)=\frac{\|\frac{1}{T-t}\cdot\sum_{\tau=t+1}^T(\psi((\tr_\tau, \tba_\tau), \bmu)-\psi((\tr_\tau,\tba_\tau), \bmu^*))-\lambda(\bmu)+\lambda(\bmu^*)\| }{\frac{1}{\sqrt{T-t}}+\|\lambda(\bmu)\|_2}.
\]
By \Cref{lem:huber}, for any $\epsilon>0$ and any $\gamma'\in(\frac{1}{2},1)$, we have
\begin{align}
&P\left( \sup_{\|\bmu-\bmu^*\|\leq d_0} Z_{T-t}(\bmu,\bmu^*)\geq2\epsilon \right)\leq 2\bar{\alpha}\bar{d}^3\epsilon^{-2} (T-t)^{-\gamma'} \label{eqn:52101}\\
&+2\exp\left(-\frac{\min\{4\bar{\alpha}^2\bar{d}^4q, \eps^2\bar{\alpha}^2\bar{d}^4\}\cdot(1-q)^{2}n^{1-\gamma'}}{2\bar{\alpha}\bar{d}^2(3-q) +2\bar{d}(2\bar{\alpha}\bar{d}^2+\eps \bar{\alpha}\bar{d}^3)(1-q)/3 }\right) \left(\frac{\gamma'(\log (T-t)+\log d_0)}{|\log(1-q)|}+1\right) (2M)^m \nonumber
\end{align}
as long as $T-t\geq n_0$, where $n_0$ satisfying $n_0^{\gamma'-\frac{1}{2}}=\frac{4\bar{\alpha}\bar{d}^2}{\epsilon}$ and $\gamma'\in(\frac{1}{2},1)$ is an arbitrary number. Moreover, we set $M\geq(6\bar{\alpha}^2\bar{d}^2)/(\epsilon \bar{\alpha}\bar{d}^2)=6\bar{\alpha}/\epsilon$ and $q=1/M$. 

We first specify $\epsilon=\frac{\underline{\alpha}\underline{\beta}}{24\bar{\alpha}\bar{d}^2}$ and $\gamma'=13/25$. Then, \eqref{eqn:52101} will become 
\begin{align}
&P\left( \sup_{\|\bmu-\bmu^*\|\leq d_0} Z_{T-t}(\bmu,\bmu^*)\geq\frac{\underline{\alpha}\underline{\beta}}{12\bar{\alpha}\bar{d}^2} \right)\leq 2\bar{\alpha}\bar{d}^3\left(\frac{12\bar{\alpha}\bar{d}^2}{\underline{\alpha}\underline{\beta}}\right)^2 (T-t)^{-\frac{13}{25}} \label{eqn:52102}\\
&+2\exp\left(-\delta\cdot (T-t)^{12/25}\right) \left(\frac{13(\log (T-t)+\log d_0)}{25|\log(1-q)|}+1\right) (2M)^m \nonumber
\end{align}
as long as $T-t\geq (96\bar{\alpha}^2\bar{d}^4/(\underline{\alpha}\underline{\beta}))^{50}$, $M=144\bar{\alpha}^2\bar{d}^2/(\underline{\alpha}\underline{\beta})$, $q=1/M=\underline{\alpha}\underline{\beta}/(144\bar{\alpha}^2\bar{d}^2)$, and
\begin{equation}\label{eqn:52103}
\delta=\frac{\min\{4\bar{\alpha}^2\bar{d}^4q, \eps^2\bar{\alpha}^2\bar{d}^4\}\cdot(1-q)^{2}}{2\bar{\alpha}\bar{d}^2(3-q) +2\bar{d}(2\bar{\alpha}\bar{d}^2+\eps \bar{\alpha}\bar{d}^3)(1-q)/3 }
\end{equation}
with $M=144\bar{\alpha}^2\bar{d}^2/(\underline{\alpha}\underline{\beta})$ and $q=1/M=\underline{\alpha}\underline{\beta}/(144\bar{\alpha}^2\bar{d}^2)$. 
Note that 
\[
\nabla\delta(\bmu)-\nabla\delta(\bmu^*)=\frac{1}{T-t}\cdot\sum_{\tau=t+1}^T(\psi((\tr_\tau, \tba_\tau), \bmu)-\psi((\tr_\tau,\tba_\tau), \bmu^*))-\lambda(\bmu)+\lambda(\bmu^*).
\]
We know that \eqref{eqn:52102} implies that
\begin{equation}\label{eqn:52901}
\begin{aligned}
   P\left( \sup_{\|\bmu-\bmu^*\|\leq d_0} \frac{\|\nabla\delta(\bmu)-\nabla\delta(\bmu^*)\|_2}{\frac{1}{\sqrt{T-t}}+\|\lambda(\bmu)\|_2}\geq\frac{\underline{\alpha}\underline{\beta}}{12\bar{\alpha}\bar{d}^2} \right)\leq &2\bar{\alpha}\bar{d}^3\left(\frac{12\bar{\alpha}\bar{d}^2}{\underline{\alpha}\underline{\beta}}\right)^2 (T-t)^{-\frac{13}{25}} \\
&+2\exp\left(-\delta\cdot (T-t)^{12/25}\right) \left(\frac{13(\log (T-t)+\log d_0)}{25|\log(1-q)|}+1\right) (2M)^m.
\end{aligned}
\end{equation}
From \Cref{assump:finitesize}, we have $\|\tbmu\|_2\leq \gamma\sqrt{m}$ and $\|\bmu^*\|_2\leq\gamma\sqrt{m}$, which implies that $r_0=\|\tbmu-\bmu^*\|_2\leq2\gamma\sqrt{m}$. Therefore, we can simply set $d_0=2\gamma\sqrt{m}$ and we know that $r_0$ is always upper bounded by $d_0$, which implies that
\begin{equation}\label{eqn:52902}
\begin{aligned}
   P\left( \sup_{\|\bmu-\bmu^*\|\leq r_0} \frac{\|\nabla\delta(\bmu)-\nabla\delta(\bmu^*)\|_2}{\frac{1}{\sqrt{T-t}}+\|\lambda(\bmu)\|_2}\geq\frac{\underline{\alpha}\underline{\beta}}{12\bar{\alpha}\bar{d}^2} \right)&\leq P\left( \sup_{\|\bmu-\bmu^*\|\leq d_0} \frac{\|\nabla\delta(\bmu)-\nabla\delta(\bmu^*)\|_2}{\frac{1}{\sqrt{T-t}}+\|\lambda(\bmu)\|_2}\geq\frac{\underline{\alpha}\underline{\beta}}{12\bar{\alpha}\bar{d}^2} \right)\\
   &\leq 2\bar{\alpha}\bar{d}^3\left(\frac{12\bar{\alpha}\bar{d}^2}{\underline{\alpha}\underline{\beta}}\right)^2 (T-t)^{-\frac{13}{25}} \\
&+2\exp\left(-\delta\cdot (T-t)^{12/25}\right) \left(\frac{13(\log (T-t)+\log 2\gamma\sqrt{m})}{25|\log(1-q)|}+1\right) (2M)^m
\end{aligned}
\end{equation}
where the first inequality follows from $r_0\leq d_0=2\gamma\sqrt{m}$ almost surely and the second inequality follows from \eqref{eqn:52901} by setting $d_0=2\gamma\sqrt{m}$. Therefore, our proof of \eqref{eqn:newfinalterm} is completed by noting that
\[
\|\lambda(\bmu)\|_2=\|2\bar{\alpha}\bar{d}^2(\bmu-\bmu^*)+\mathbb{E}_{\tba}[\tba\cdot (F(\tba^\top\bmu|\tba)-F(\tba^\top\bmu^*|\tba))]\|_2\leq 3\bar{\alpha}\bar{d}^2\cdot\|\bmu-\bmu^*\|_2.
\]
\end{myproof}

\begin{myproof}[Proof of \Cref{lem:61501}]
Following \citet{li2021online}, we denote by
\[
\phi((r,\ba),\bmu)=-\ba^\top\cdot\bI(r>\ba^\top\bmu).
\]
Then, from Lemma 1 of \citet{li2021online}, we have
\begin{equation}\label{eqn:61003}
\mathbb{E}_{(\tr,\tba)}[\tr-\tba^\top\tbmu]^+-\mathbb{E}_{(\tr,\tba)}[\tr-\tba^\top\bmu^*]^+=\mathbb{E}_{(\tr,\tba)}[\phi((\tr,\tba),\bmu^*)]\cdot(\tbmu-\bmu^*)+\mathbb{E}_{(\tr,\tba)}\left[ \int_{\tba^\top\tbmu}^{\tba^\top\bmu^*}(\bI(\tr>v)-\bI(\tr>\tba^\top\bmu^*))dv \right].
\end{equation}
Also, from Lemma 2 of \cite{li2021online}, we have
\begin{equation}\label{eqn:61004}
\begin{aligned}
    \frac{1}{T-t}\cdot\sum_{\tau=t+1}^T([\tr_{\tau}-\tba^\top_{\tau}\tbmu]^+-[\tr_{\tau}-\tba^\top_{\tau}\bmu^*]^+)=&\frac{1}{T-t}\cdot\sum_{\tau=t+1}^T \phi((\tr_{\tau},\tba_{\tau}),\bmu^*)\cdot(\tbmu-\bmu^*)\\
    &+\frac{1}{T-t}\cdot\sum_{\tau=t+1}^T\int_{\tba_{\tau}^\top\tbmu}^{\tba_{\tau}^\top\bmu^*}(\bI(\tr_{\tau}>v)-\bI(\tr_{\tau}>\tba_{\tau}^\top\bmu^*))dv.
\end{aligned}
\end{equation}
Plugging \eqref{eqn:61003} and \eqref{eqn:61004} into \eqref{eqn:61005}, we have that 
\begin{equation}\label{eqn:61006}
\begin{aligned}
  &\frac{\underline{\alpha}\underline{\beta}}{2}\cdot\|\tbmu-\bmu^*\|_2^2\leq \underbrace{\frac{1}{T-t}\cdot\sum_{\tau=t+1}^T \phi((\tr_{\tau},\tba_{\tau}),\bmu^*)\cdot(\tbmu-\bmu^*)-\mathbb{E}_{(\tr,\tba)}[\phi((\tr,\tba),\bmu^*)]\cdot(\tbmu-\bmu^*)}_{I}\\
  &+\underbrace{\frac{1}{T-t}\cdot\sum_{\tau=t+1}^T\int_{\tba_{\tau}^\top\tbmu}^{\tba_{\tau}^\top\bmu^*}(\bI(\tr_{\tau}>v)-\bI(\tr_{\tau}>\tba_{\tau}^\top\bmu^*))dv-\mathbb{E}_{(\tr,\tba)}\left[ \int_{\tba^\top\tbmu}^{\tba^\top\bmu^*}(\bI(\tr>v)-\bI(\tr>\tba^\top\bmu^*))dv \right]}_{II}.
\end{aligned}
\end{equation}
We proceed by bounding the term I and term II separately. \\
\textbf{Bound I:} by Matrix Hoeffding's inequality, we have that
\[
P\left( \|\frac{1}{T-t}\cdot\sum_{\tau=t+1}^T \phi((\tr_{\tau},\tba_{\tau}),\bmu^*)-\mathbb{E}_{(\tr,\tba)}[\phi((\tr,\tba),\bmu^*)]\|_2\geq\epsilon \right)\leq m\cdot\exp\left(-\frac{\epsilon^2\cdot(T-t)}{\bar{d}^2}\right)
\]
which implies that
\begin{equation}\label{eqn:61007}
    P\left(\left|I\right|\geq\epsilon\cdot\|\tbmu-\bmu^*\|_2\right)\leq m\cdot\exp\left(-\frac{\epsilon^2\cdot(T-t)}{\bar{d}^2}\right).
\end{equation}
\textbf{Bound II:} 
We define a function 
\[
\eta((r,\ba),\bmu_1,\bmu_2)=\int_{\ba^\top\bmu_1}^{\ba^\top\bmu_2}(\bI(r>v)-\bI(r>\ba^\top\bmu^*))dv
\]
for any $\bmu_1, \bmu_2\in \hat{\mathcal{S}}$. We have
\begin{equation}\label{eqn:60801}
II= \frac{1}{T-t}\cdot\sum_{\tau=t+1}^T\eta((\tr_{\tau},\tba_{\tau}),\tbmu,\bmu^*)-\mathbb{E}_{(\tr,\tba)}[\eta((\tr,\tba),\tbmu,\bmu^*)].
\end{equation}
We utilize a splitting scheme to split the set $\{\bmu:\|\bmu-\bmu^*\|_2\leq \sqrt{m}\cdot\gamma\}$ into disjoint cubes, similar to \cite{huber1967under} and \cite{li2021online}. It is clear to see that $\|\tbmu-\bmu^*\|_2\leq\sqrt{m}\cdot\gamma$. The idea is to divide the region $\{\bmu:\|\bmu-\bmu^*\|_2\leq\sqrt{m}\cdot\gamma\}$ into a slowly increasing number of smaller cubes. We consider the concentric cubes:
\[
C_k=\{\bmu:\|\bmu-\bmu^*\|_2\leq(1-q)^k\cdot\sqrt{m}\gamma\},~~k=1,\dots,k_0
\]
where $q$ is a constant such that $\bar{\alpha}\cdot\bar{d}^2\cdot\frac{q(1+q)}{(1-q)^2}\leq \frac{\underline{\alpha}\underline{\beta}}{32}$, and $k_0$ is a constants such that $(1-q)^{k_0}\cdot\sqrt{m}\gamma=\epsilon$. We then further divide the region $C_{k-1}\setminus C_k$ into cubes $\{\Omega_{kl}\}_{l=1}^{l_k}$ with edges of length $2(1-q)^k\cdot q\cdot\sqrt{m}\gamma$ such that the centers of these cubes $\bm{\xi}_{kl}$ satisfies
\[
\|\bm{\xi}_{kl}\|_2=(1-q)^{k-1}(1-\frac{q}{2})\cdot\sqrt{m}\gamma,~~ l=1,\dots,k_l.
\]
In total, there are no more than $k_0\cdot(\frac{2}{q})^m$ number of cubes. We now denote by $\tilde{\Omega}$ the cube that contains $\tbmu$ and denote by $\tilde{\bm{\xi}}$ the center of the cube $\tilde{\Omega}$. Note that both $\tilde{\Omega}$ and $\tilde{\bm{\xi}}$ are random, where the randomness arises from $\tbmu$. 
\[\begin{aligned}
\frac{1}{T-t}\cdot\sum_{\tau=t+1}^T\eta((\tr_{\tau},\tba_{\tau}),\tbmu,\bmu^*)-\mathbb{E}_{(\tr,\tba)}[\eta((\tr,\tba),\tbmu,\bmu^*)]=&\frac{1}{T-t}\cdot\sum_{\tau=t+1}^T\eta((\tr_{\tau},\tba_{\tau}),\tbmu,\tbxi)-\mathbb{E}_{(\tr,\tba)}[\eta((\tr,\tba),\tbmu,\tbxi)]\\
&+\frac{1}{T-t}\cdot\sum_{\tau=t+1}^T\eta((\tr_{\tau},\tba_{\tau}),\tbxi,\bmu^*)-\mathbb{E}_{(\tr,\tba)}[\eta((\tr,\tba),\tbxi,\bmu^*)]
\end{aligned}\]
which implies that 
\begin{equation}\label{eqn:61001}
\begin{aligned}
    &P\left(\frac{1}{T-t}\cdot\sum_{\tau=t+1}^T\eta((\tr_{\tau},\tba_{\tau}),\tbmu,\bmu^*)-\mathbb{E}_{(\tr,\tba)}[\eta((\tr,\tba),\tbmu,\bmu^*)]\geq\epsilon^2+\epsilon\|\tbmu-\bmu^*\|_2+\frac{\underline{\alpha}\underline{\beta}\|\tbmu-\bmu^*\|_2^2}{8}\right)\\
    \leq&P\left(\frac{1}{T-t}\cdot\sum_{\tau=t+1}^T\eta((\tr_{\tau},\tba_{\tau}),\tbmu,\tbxi)-\mathbb{E}_{(\tr,\tba)}[\eta((\tr,\tba),\tbmu,\tbxi)]\geq\frac{\epsilon^2}{2}+\frac{\epsilon\cdot\|\tbmu-\bmu^*\|_2}{2}+\frac{\underline{\alpha}\underline{\beta}\|\tbmu-\bmu^*\|_2^2}{16}\right)\\
    &+P\left(\frac{1}{T-t}\cdot\sum_{\tau=t+1}^T\eta((\tr_{\tau},\tba_{\tau}),\tbxi,\bmu^*)-\mathbb{E}_{(\tr,\tba)}[\eta((\tr,\tba),\tbxi,\bmu^*)]\geq\frac{\epsilon^2}{2}+\frac{\epsilon\cdot\|\tbmu-\bmu^*\|_2}{2}+\frac{\underline{\alpha}\underline{\beta}\|\tbmu-\bmu^*\|_2^2}{16}\right).
\end{aligned}
\end{equation}
We have the following result.
\begin{claim}\label{claim:61101}
It holds that
\[\begin{aligned}
&P\left(\frac{1}{T-t}\cdot\sum_{\tau=t+1}^T\eta((\tr_{\tau},\tba_{\tau}),\tbmu,\tbxi)-\mathbb{E}_{(\tr,\tba)}[\eta((\tr,\tba),\tbmu,\tbxi)]\geq\frac{\epsilon^2}{2}+\frac{\epsilon\cdot\|\tbmu-\bmu^*\|_2}{2}+\frac{\underline{\alpha}\underline{\beta}\|\tbmu-\bmu\|_2^2}{16}\right)\\
&\leq k_0\cdot\left(\frac{2}{q}\right)^m\cdot\exp\left(-\frac{\epsilon^2(T-t)}{2q^2\bar{d}^2}\right)
\end{aligned}\]
where $q$ is a constant such that $\bar{\alpha}\cdot\bar{d}^2\cdot\frac{q(1+q)}{(1-q)^2}\leq \frac{\underline{\alpha}\underline{\beta}}{32}$, and $k_0$ is a constants such that $(1-q)^{k_0}\cdot\sqrt{m}\gamma=\epsilon$.
\end{claim}
We also have the following result.
\begin{claim}\label{claim:61102}
It holds that
\[\begin{aligned}
&P\left(\frac{1}{T-t}\cdot\sum_{\tau=t+1}^T\eta((\tr_{\tau},\tba_{\tau}),\tbxi,\bmu^*)-\mathbb{E}_{(\tr,\tba)}[\eta((\tr,\tba),\tbxi,\bmu^*)]\geq\frac{\epsilon^2}{2}+\frac{\epsilon\cdot\|\tbmu-\bmu^*\|_2}{2}+\frac{\underline{\alpha}\underline{\beta}\|\tbmu-\bmu^*\|_2^2}{16}\right)\\
&\leq k_0\cdot\left(\frac{2}{q}\right)^m\cdot\exp\left(-\frac{\epsilon^2\cdot(T-t)q^2}{2\bar{d}^2}\right)
\end{aligned}\]
where $q$ is a constant such that $\bar{\alpha}\cdot\bar{d}^2\cdot\frac{q(1+q)}{(1-q)^2}\leq \frac{\underline{\alpha}\underline{\beta}}{32}$, and $k_0$ is a constants such that $(1-q)^{k_0}\cdot\sqrt{m}\gamma=\epsilon$.
\end{claim}
Combining \Cref{claim:61101}, \Cref{claim:61102} and \eqref{eqn:61001}, we have that
\begin{equation}\label{eqn:61105}
\begin{aligned}
 &P\left( \text{II}\geq \epsilon^2+\epsilon\|\tbmu-\bmu^*\|_2+\frac{\underline{\alpha}\underline{\beta}\|\tbmu-\bmu^*\|_2^2}{8} \right)\\   =&P\left(\frac{1}{T-t}\cdot\sum_{\tau=t+1}^T\eta((\tr_{\tau},\tba_{\tau}),\tbmu,\bmu^*)-\mathbb{E}_{(\tr,\tba)}[\eta((\tr,\tba),\tbmu,\bmu^*)]\geq\epsilon^2+\epsilon\|\tbmu-\bmu^*\|_2+\frac{\underline{\alpha}\underline{\beta}\|\tbmu-\bmu^*\|_2^2}{8}\right)\\
    \leq& k_0\cdot\left(\frac{2}{q}\right)^m\cdot\exp\left(-\frac{\epsilon^2(T-t)}{2q^2\bar{d}^2}\right)+k_0\cdot\left(\frac{2}{q}\right)^m\cdot\exp\left(-\frac{\epsilon^2\cdot(T-t)q^2}{2\bar{d}^2}\right)
\end{aligned}
\end{equation}
where $q$ is a constant such that $\bar{\alpha}\cdot\bar{d}^2\cdot\frac{q(1+q)}{(1-q)^2}\leq \frac{\underline{\alpha}\underline{\beta}}{32}$, and $k_0$ is a constants such that $(1-q)^{k_0}\cdot\sqrt{m}\gamma=\epsilon$. Our bound of the term II is now completed.

We now use the bound on I and II to prove the probability bound on the event $\|\tbmu-\bmu^*\|_2^2\geq\epsilon^2$. Combining \eqref{eqn:61007} and  \eqref{eqn:61105}, we have
\[\begin{aligned}
&P\left(\text{I}+\text{II}\geq 2\epsilon^2+2\epsilon\|\tbmu-\bmu^*\|_2+\frac{\underline{\alpha}\underline{\beta}\|\tbmu-\bmu^*\|_2^2}{4}\right)\\
\leq& P\left(\text{I}\geq\epsilon^2+\epsilon\|\tbmu-\bmu^*\|_2+\frac{\underline{\alpha}\underline{\beta}\|\tbmu-\bmu^*\|_2^2}{8} \right)+P\left(\text{II}\geq\epsilon^2+\epsilon\|\tbmu-\bmu^*\|_2+\frac{\underline{\alpha}\underline{\beta}\|\tbmu-\bmu^*\|_2^2}{8} \right)\\
\leq & m\cdot\exp\left(-\frac{\epsilon^2\cdot(T-t)}{\bar{d}^2}\right)+
k_0\cdot\left(\frac{2}{q}\right)^m\cdot\left(\exp\left(-\frac{\epsilon^2(T-t)}{2q^2\bar{d}^2}\right)+ \exp\left(-\frac{\epsilon^2\cdot(T-t)q^2}{2\bar{d}^2}\right) \right)
\end{aligned}\]
where $c_2$ is a constant. Further note that 
\[
\frac{\underline{\alpha}\underline{\beta}}{2}\cdot\|\tbmu-\bmu^*\|_2^2\leq \text{I}+\text{II},
\]
and
\[
\frac{\underline{\alpha}\underline{\beta}}{2}\cdot\|\tbmu-\bmu^*\|_2^2\geq 2\epsilon^2+2\epsilon\|\tbmu-\bmu^*\|_2+\frac{\underline{\alpha}\underline{\beta}\|\tbmu-\bmu^*\|_2^2}{4}
\]
implies that
\[
\|\tbmu-\bmu^*\|_2^2\geq \epsilon^2\cdot \frac{4(2+\sqrt{2\underline{\alpha}\underline{\beta}+4})^2}{\underline{\alpha}^2\underline{\beta}^2}.
\]
Therefore, we have that
\[\begin{aligned}
P\left( \|\tbmu-\bmu^*\|_2^2\geq \epsilon^2\cdot \frac{4(2+\sqrt{2\underline{\alpha}\underline{\beta}+4})^2}{\underline{\alpha}^2\underline{\beta}^2} \right)\leq& m\cdot\exp\left(-\frac{\epsilon^2\cdot(T-t)}{\bar{d}^2}\right)\\
&+
k_0\cdot\left(\frac{2}{q}\right)^m\cdot\left(\exp\left(-\frac{\epsilon^2(T-t)}{2q^2\bar{d}^2}\right)+ \exp\left(-\frac{\epsilon^2\cdot(T-t)q^2}{2\bar{d}^2}\right) \right).
\end{aligned}\]
By noting that $k_0=O(\log(\frac{1}{\eps})$, we have
\[
P\left( \|\tbmu-\bmu^*\|_2^2\geq k_1/(\underline{\alpha}^2\underline{\beta}^2)\cdot\eps^2 \right)\leq k_2\cdot \left(\frac{2}{q}\right)^m\cdot \log\left(\frac{1}{\eps}\right)\cdot \exp\left(-\frac{k_3\eps^2(T-t)}{\bar{d}^2} \right)
\]
where $k_1$, $k_2$, $k_3$ are constant numbers and $q$ is a constant such that $\bar{\alpha}\cdot\bar{d}^2\cdot\frac{q(1+q)}{(1-q)^2}\leq \frac{\underline{\alpha}\underline{\beta}}{32}$. Our proof is thus completed.
\end{myproof}
\begin{myproof}[Proof of \Cref{claim:61101}]
We denote by 
\[
g((r,\ba),\bmu_1,d)=\sup_{\bmu_2:\|\bmu_2-\bmu_1\|_2\leq d}\eta((r,\ba),\bmu_1,\bmu_2).
\]
Then we have
\[
\frac{1}{T-t}\cdot\sum_{\tau=t+1}^T\eta((\tr_{\tau},\tba_{\tau}),\tbmu,\tbxi)-\mathbb{E}_{(\tr,\tba)}[\eta((\tr,\tba),\tbmu,\tbxi)]\leq \frac{1}{T-t}\cdot\sum_{\tau=t+1}^Tg((\tr_{\tau},\tba_{\tau}),\tbxi,\tilde{d})+\mathbb{E}_{(\tr,\tba)}[g((\tr,\tba),\tbxi,\tilde{d})]
\]
where $\tilde{d}$ denotes the edge length of the cube $\tilde{\Omega}$. We now consider the cubes $\{\Omega_{kl}\}$ and the center concentric cubes $C_{k_0}$ separately.\\
(i). If the cube $\tilde{\Omega}\in\{\Omega_{kl}\}$, then, we note that
\[
\mathbb{E}_{(\tr,\tba)}[g((\tr,\tba),\tbxi,\tilde{d})]\leq \mathbb{E}_{\tba}\left[\sup_{\tbmu_2:\|\tbmu_2-\tbxi\|_2\leq \tilde{d}}\int^{\tba^\top\bmu^*}_{\tba^\top\tbmu_2}\int_{\tba^\top\tbmu_2}^{\tba^\top\tbxi}(\bI(r>v)-\bI(r>\tba^\top\bmu^*))dvdF(r|\tba)\right]\leq\bar{\alpha}\bar{d}^2\cdot \tilde{d}\cdot(\|\bmu^*-\tbxi\|_2+\tilde{d}).
\]
By definition, we have $\tilde{d}=q\cdot \|\bmu^*-\tbxi\|_2$ and $\|\bmu^*-\tbmu\|_2\geq\|\bmu^*-\tbxi\|_2-\tilde{d}$. Then, we have
\[
\mathbb{E}_{(\tr,\tba)}[g((\tr,\tba),\tbxi,\tilde{d})]\leq\bar{\alpha}\cdot\bar{d}^2\cdot\frac{q(1+q)}{(1-q)^2}\cdot\|\bmu^*-\tbmu\|_2^2\leq \frac{\underline{\alpha}\underline{\beta}\|\tbmu-\bmu^*\|_2^2}{32}
\]
by specifying $q$ such that $\bar{\alpha}\cdot\bar{d}^2\cdot\frac{q(1+q)}{(1-q)^2}\leq \frac{\underline{\alpha}\underline{\beta}}{32}$.\\
(ii). If the cube $\tilde{\Omega}=C_{k_0}$, then by noting $\tbxi=\bmu^*$ and $(1-q)^{k_0}\cdot\sqrt{m}\gamma=\epsilon$, we have that
\[
\mathbb{E}_{(\tr,\tba)}[g((\tr,\tba),\tbxi,\tilde{d})]\leq \mathbb{E}_{\tba}\left[\sup_{\bmu_2:\|\bmu_2-\tbxi\|_2\leq \tilde{d}}\int^{\tba^\top\bmu^*}_{\tba^\top\bmu_2}\int_{\tba^\top\bmu_2}^{\tba^\top\tbxi}(\bI(r>v)-\bI(r>\tba^\top\bmu^*))dvdF(r|\tba)\right]\leq\bar{\alpha}\cdot\bar{d}^2\cdot \epsilon^2.
\]
Therefore, for both (i) and (ii), it holds that
\[\begin{aligned}
&P\left(\frac{1}{T-t}\cdot\sum_{\tau=t+1}^T\eta((\tr_{\tau},\tba_{\tau}),\tbmu,\tbxi)-\mathbb{E}_{(\tr,\tba)}[\eta((\tr,\tba),\tbmu,\tbxi)]\geq\bar{\alpha}\bar{d}^2\epsilon^2+\frac{\epsilon\cdot\|\tbmu-\bmu^*\|_2}{2}+\frac{\underline{\alpha}\underline{\beta}\|\tbmu-\bmu^*\|_2^2}{16}\right)\\
\leq&P\left(\frac{1}{T-t}\cdot\sum_{\tau=t+1}^Tg((\tr_{\tau},\tba_{\tau}),\tbxi,\tilde{d})+\mathbb{E}_{(\tr,\tba)}[g((\tr,\tba),\tbxi,\tilde{d})]\geq\bar{\alpha}\bar{d}^2\epsilon^2+\frac{\epsilon\cdot\|\tbmu-\bmu^*\|_2}{2}+\frac{\underline{\alpha}\underline{\beta}\|\tbmu-\bmu^*\|_2^2}{16}\right)\\
\leq&P\left(\frac{1}{T-t}\cdot\sum_{\tau=t+1}^Tg((\tr_{\tau},\tba_{\tau}),\tbxi,\tilde{d})-\mathbb{E}_{(\tr,\tba)}[g((\tr,\tba),\tbxi,\tilde{d})]\geq\frac{\epsilon\cdot\|\tbmu-\bmu^*\|_2}{2}\right)\\
\leq& P\left(\frac{1}{T-t}\cdot\sum_{\tau=t+1}^Tg((\tr_{\tau},\tba_{\tau}),\tbxi,\tilde{d})-\mathbb{E}_{(\tr,\tba)}[g((\tr,\tba),\tbxi,\tilde{d})]\geq\frac{\epsilon\cdot \tilde{d}}{2q}\right).
\end{aligned}\]
We apply the Hoeffding inequality to bound the above probability for each cube $\Omega_{kl}$, and $C_{k_0}$. It holds that 
\[
g((r,\ba),\bxi_{kl}, d_{kl})\leq \bar{d}\cdot d_{kl},~\text{and~} g((r,\ba),\bxi_{k_0}, d_{k_0})\leq \bar{d}\cdot d_{k_0} ~~\forall (r,\ba)
\]
where $d_{kl}$ denotes the length of the edge of the cube $\Omega_{kl}$ and $d_{k_0}$ denotes the length of the edge of the cube $C_{k_0}$.
By Hoeffding's inequality, we have
\[
P\left(\frac{1}{T-t}\cdot\sum_{\tau=t+1}^Tg((\tr_{\tau},\tba_{\tau}),\bxi_{kl},d_{kl})-\mathbb{E}_{(\tr,\tba)}[g((\tr,\tba),\bxi_{kl},d_{kl})]\geq\frac{\epsilon\cdot d_{kl}}{2q}\right)\leq\exp\left(-\frac{\epsilon^2(T-t)}{2q^2\bar{d}^2}\right)
\]
for each cube $\Omega_{kl}$, and 
\[
P\left(\frac{1}{T-t}\cdot\sum_{\tau=t+1}^Tg((\tr_{\tau},\tba_{\tau}),\bxi_{k_0},d_{k_0})-\mathbb{E}_{(\tr,\tba)}[g((\tr,\tba),\bxi_{k_0},d_{k_0})]\geq\frac{\epsilon\cdot d_{k_0}}{2q}\right)\leq\exp\left(-\frac{\epsilon^2(T-t)}{2q^2\bar{d}^2}\right)
\]
for the cube $C_{k_0}$.
Note that
\[
\frac{1}{T-t}\cdot\sum_{\tau=t+1}^Tg((\tr_{\tau},\tba_{\tau}),\tbxi,\tilde{d})-\mathbb{E}_{(\tr,\tba)}[g((\tr,\tba),\tbxi,\tilde{d})]\geq\frac{\epsilon\cdot \tilde{d}}{2q}
\]
implies that
\[
\frac{1}{T-t}\cdot\sum_{\tau=t}^Tg((\tr_{\tau},\tba_{\tau}),\bxi_{k_0},d_{k_0})-\mathbb{E}_{(\tr,\tba)}[g((\tr,\tba),\bxi_{k_0},d_{k_0})]\geq\frac{\epsilon\cdot d_{k_0}}{2q},
\]
or there exists at least one $\Omega_{kl}$ such that
\[
\frac{1}{T-t}\cdot\sum_{\tau=t}^Tg((\tr_{\tau},\tba_{\tau}),\bxi_{kl},d_{kl})-\mathbb{E}_{(\tr,\tba)}[g((\tr,\tba),\bxi_{kl},d_{kl})]\geq\frac{\epsilon\cdot d_{kl}}{2q}.
\]
Applying the union bound, our proof is completed.
\end{myproof}
\begin{myproof}[Proof of \Cref{claim:61102}]
Note that 
\[
\eta((r,\ba),\tbxi,\bmu^*)\leq\bar{d}\cdot\|\tbxi-\bmu^*\|_2.
\]
Then, if $\tilde{\Omega}\in\{\Omega_{kl}\}$, we have that
\[
\eta((r,\ba),\tbxi,\bmu^*)\leq\bar{d}\cdot\|\tbxi-\bmu^*\|_2\leq \frac{\bar{d}\cdot\|\tbmu-\bmu^*\|_2}{q}
\]
and if $\tilde{\Omega}=C_{k_0}$, we have that
\[
\eta((r,\ba),\tbxi,\bmu^*)\leq\bar{d}\cdot\|\tbxi-\bmu^*\|_2=0
\]
by noting $\tbxi=\bxi_{k_0}=\bmu^*$.
Now, for each $\Omega_{kl}$, by Hoeffding's inequality, we have
\[
P\left(\frac{1}{T-t}\cdot\sum_{\tau=t+1}^T\eta((\tr_{\tau},\tba_{\tau}),\bxi_{kl},\bmu^*)-\mathbb{E}_{(\tr,\tba)}[\eta((\tr,\tba),\bxi_{kl},\bmu^*)]\geq\frac{\epsilon\cdot\|\tbmu-\bmu^*\|_2}{2}\right)\leq \exp\left(-\frac{\epsilon^2\cdot(T-t)q^2}{2\bar{d}^2}\right).
\]
Our proof is completed from the union bound over all $\Omega_{kl}$.
\end{myproof}

\end{APPENDICES}

\end{document}